\documentclass{article}

\usepackage[preprint]{corl_2026} % Uncomment for pre-prints (e.g., arxiv).

\usepackage{multicol}
\usepackage[normalem]{ulem}
\usepackage[table,xcdraw]{xcolor}
\usepackage{algorithm}
\usepackage{algpseudocode}
\usepackage{amsfonts} 
\usepackage[skins,theorems]{tcolorbox}
\usepackage{graphicx}
\usepackage{multirow}
\usepackage{array,booktabs}
\usepackage{bm}
\usepackage[capitalise, noabbrev]{cleveref}
\usepackage{tabularx}
\usepackage{subfig}
\usepackage{xspace}
\usepackage{wrapfig}
\usepackage{needspace}
\usepackage{placeins}

\makeatletter
\renewcommand{\section}{%
  \@startsection{section}{1}{\z@}%
                {-1.5ex \@plus -0.4ex \@minus -0.2ex}%
                { 0.8ex \@plus  0.2ex \@minus  0.2ex}%
                {\large\bf\raggedright}%
}
\renewcommand{\subsection}{%
  \@startsection{subsection}{2}{\z@}%
                {-1.2ex \@plus -0.4ex \@minus -0.2ex}%
                { 0.4ex \@plus  0.2ex}%
                {\normalsize\bf\raggedright}%
}
\renewcommand{\paragraph}{%
  \@startsection{paragraph}{4}{\z@}%
                {0.2ex \@plus 0.1ex \@minus 0.1ex}%
                {-1em}%
                {\normalsize\bf}%
}
\makeatother

\title{Rotation-Aware Point-Cloud Embeddings for Vision-Based In-Hand Reorientation}

\author{
  Yashom Dighe\\
  Dept. of Computer Science and Engineering\\
  University at Buffalo\\
  \texttt{yashomna@buffalo.edu}
  \And
  Karthik Dantu\\
  Dept. of Computer Science and Engineering\\
  University at Buffalo\\
  \texttt{kdantu@buffalo.edu}
}

\begin{document}
\maketitle
\vspace{-1.4em}

\begin{nolinenumbers}
\noindent\includegraphics[width=\textwidth,trim=0 65pt 0 0,clip]{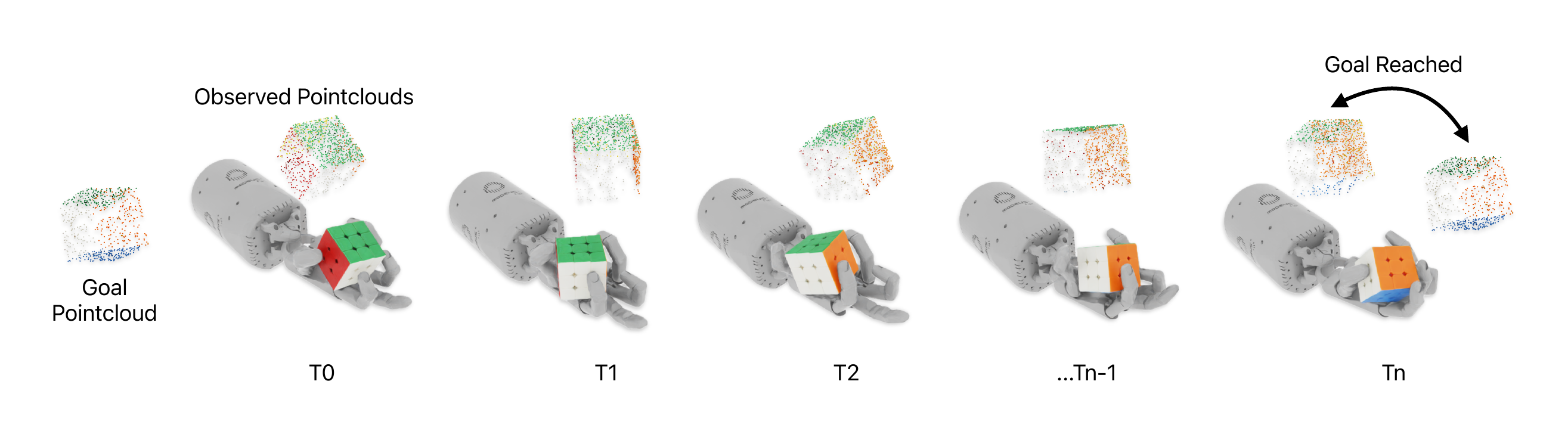}
\captionof{figure}{\emph{Policy rollout}. We train a rotation-aware point-cloud encoder that enables direct RL policy learning from observed and goal point clouds, rotating the object toward the target geometry without object pose, relative pose, dense flow, or teacher-action supervision.}
\label{fig:policy-rollout}
\end{nolinenumbers}
\vspace{-0.2em}

\begin{abstract}
Point-cloud goals provide a direct way to specify dexterous in-hand reorientation: instead of defining an object-specific pose frame or estimating 6D pose at test time, the policy is given the desired 3D geometry of the object. Yet raw point-cloud goal conditioning is poorly conditioned for policy learning. Current and goal clouds are unordered, independently sampled, and often visibility-dependent, so their discrepancy entangles object rotation with permutation, resampling, and unstable correspondence structure. For this reason, prior point-cloud manipulation methods typically add structure outside the representation itself, such as explicit pose or relative-pose inputs, dense flow features, or distillation from privileged teachers. We close this gap by learning a rotation-aware point-cloud embedding whose Euclidean latent distance is calibrated to the SO(3) geodesic error between object orientations. The resulting representation turns current–goal comparison into a smooth control signal, allowing a model-free RL policy to act from current/goal point-cloud embeddings, proprioception, and centroid metadata, without object pose, relative pose, dense flow, or teacher-action supervision. In in-hand reorientation experiments, this interface matches privileged-state and distillation-based baselines while avoiding brittle test-time computation of structured pose or flow inputs. These results suggest that point-cloud goals become practical for this task when the representation, rather than an external module, encodes the task-relevant geometry of rotation. We also show evidence that generic visual point-cloud pretraining is insufficient for such a current-goal comparison because it discards the task relevant state and preserves only shape features. 
\end{abstract}
\vspace{-0.9em}

\vspace{-1.0em}

\section{Introduction}
\label{sec:intro}
Point clouds are a natural goal representation for geometric manipulation: they explicitly specify what the robot should make the scene look like in 3D. For a task like in-hand reorientation, this means the goal can be given by showing the object in its desired orientation, rather than defining an object-specific pose frame or requiring a 6D pose estimate at policy inference. Our conjecture is that since point clouds encode shape, pose, contact geometry, and spatial arrangement, a policy should in principle be able to solve such tasks by comparing the current point cloud to a goal point cloud without any additional inputs. In practice, point-cloud goal conditioning is difficult because the current-goal comparison is hard to make control-friendly in raw point-set space. In this paper, we show that metric learning can embed point clouds in a representation where this comparison becomes feasible for dexterous control, without providing the policy object pose, relative pose, dense flow, or distillation from a privileged teacher.

The complexity of using point clouds comes from the representation itself: they are unordered surface samples with no persistent identity across observations. As a result, changes in the observation are entangled with permutation, resampling, visibility, and discontinuous nearest-neighbor assignments making it. Prior work has made point-cloud goals usable by adding structure outside the raw point-cloud representation. Geometry-Dex~\citep{huang2021generalization} learn RL policies that use point-cloud observations for object geometry, but supplies the policy with explicit pose information for the reorientation objective. Flow-based methods such as HACMan also learn RL policies by providing structured current-goal features by specifying how points should move from the current geometry toward the target geometry~\citep{zhou2023hacman}. These approaches are effective but also presuppose that these structured quantities can be computed and specified robustly. While recent methods such as FoundationPose~\cite{wen2024foundationposeunified6dpose} have made substantial progress in 6D object pose estimation, estimating pose reliably from visual observations remains a challenging problem, as reflected by benchmarks such as BOP~\cite{hodan2018bop}. Similarly, robust point-cloud registration under partial observations and occlusion is still difficult; for example, HACMan~\citep{zhou2023hacman} identifies inaccurate registration as one of its failure modes. Distillation provides another form of scaffolding: \citep{chen2022visualdexterity,chen2021system} first train a privileged teacher and then train a point-cloud visual student to imitate its actions. This approach works around the brittle pose estimation/registration operation by directly learning on the raw pointclouds by providing stable action supervision. But, distillation can result in a learned representation that is coupled to a particular embodiment and to the exploration or data collection policy that generated the dataset and fundamentally depends on the availability of quality expert supervision or trajectories labeled with robot actions. 

Our goal is to remove this scaffolding and ask whether the representation itself can make point-cloud goal conditioning usable for standard policy learning. A generic point-cloud encoder trained for reconstruction, shape, or identity may preserve what object is present while failing to organize the task-relevant state in a way that makes current-goal comparison useful for control. We propose that, given two point clouds of the same object under different SO(3) transforms, the difference in their latent representation should reflect the difference in the transforms. To achieve this, we train a point-cloud encoder with a metric-learning~\citep{kaya2019deep} objective that calibrates latent distance to the task's $SO(3)$ geodesic error. We then use embeddings of the current and goal point clouds as the observation interface to solve the down-stream task of in-hand reorientation using model-free RL \textbf{without using} object pose, relative pose, dense flow, or teacher-action supervision. We demonstrate that a model-free policy not only learns stably from our representation, but also matches the performance of a distillation baseline and out-performs methods using privileged information like pose, and flow in a deployment setting. This is a demanding test because model-free RL must learn from reward rather than per-step target actions. In contrast, a policy using a task-agnostic Point-MAE encoder fine-tuned on the same YCB objects fails to learn, suggesting that generic point-cloud features do not provide the task-relevant signal needed for reorientation.

% \section{Motivation}
% \input{Sections/motivation}

\section{Methodology}
% \input{Sections/problem}

% \section{Method}
\subsection{Pose-Aware Encoder}
\label{sec:encoder}
We learn a PointNet++~\cite{qi2017pointnet++} based point-cloud encoder, $\phi$, that maps each colored object cloud $\mathcal{P}\in\mathbb{R}^{N\times 6}$ to a compact embedding $z=\phi(\mathcal{P})$. The encoder is trained in a Siamese style~\cite{bromley1993signature} to independently encode point clouds as 128-dimensional embeddings whose Euclidean distance matches the geodesic distance between object orientations. Formally, for two clouds of the same object at orientations $q_1$ and $q_2$, we regress the Euclidean latent distance to the SO(3) geodesic rotation error,
\begin{align}
\label{eq:loss}
    \mathcal{L}_{\text{metric}}=\mathbb{E}[(\|\phi(\mathcal{P}_1)-\phi(\mathcal{P}_2)\|_2 - 2\arccos|\langle q_1,q_2\rangle|)^2]
\end{align}
Thus nearby orientations are encouraged to have nearby embeddings and distant orientations to have distant embeddings.

We train the encoder on colored point clouds $(x,y,z,r,g,b)$ sampled from the textured mesh models in the YCB dataset~\citep{calli2017yale}. A single encoder is trained across all objects rather than separately per instance. To improve robustness to partial observations that arise in the downstream in-hand reorientation task, training uses an occlusion curriculum that progressively removes points from the input clouds, exposing the model to increasingly incomplete geometry. We then fine-tune the encoder on a smaller IsaacLab dataset generated by rendering objects with RGB-D cameras and back-projecting the depth images into point clouds, producing hand-like occlusions that better match the observations seen during policy learning. Details of this curriculum and fine-tuning can be found in Appendix~\ref{app:train}.
 
For learning the reorientation policy, $\phi$ is frozen and applied separately to the current and goal clouds.
% \yom{Details on how its trained}
\subsection{In-hand reorientation}
The downstream task of in-hand reorientation can be formulated as a goal-augmented Markov decision process $\mathcal{M}=\langle \mathcal{S},\mathcal{A},\mathcal{G},p,r,\rho_0,\gamma\rangle$. At the start of each episode, a goal $g \in \mathcal{G}$ is sampled from a goal distribution $p(g)$. In our setting $g \equiv q_g \in SO(3)$. The environment is initialized by sampling $(s_0,g)\sim \rho_0(\cdot)$ and the agent interacts for $T$ steps by selecting actions $a_t \sim \pi(a_t\mid o_t,g)$, inducing transitions $s_{t+1}\sim p(s_{t+1}\mid s_t,a_t,g)$. The objective is to learn a policy that maximizes the expected discounted return over goals 
\vspace{-0.25em}
\begin{align}
\max_{\pi}\;\; \mathbb{E}_{g\sim p(g)}\Big[\mathbb{E}_{\tau\sim \pi,p}\big[\sum_{t=0}^{T-1}\gamma^t\, r(s_t,a_t,g)\big]\Big]. \nonumber
\end{align}
\vspace{0.05em}
We use the standard IsaacLab~\citep{mittal2025isaaclab} in-hand reorientation task definition, with no task-specific modifications.

\textbf{System:} The robot is a simulated Shadow Hand~\citep{shadowrobotShadowDexterous} with 20 actuated joints manipulating a single rigid object in-hand. The object is chosen from the YCB dataset~\cite{calli2017yale}, a widely used benchmark set of everyday household objects with standardized RGB-D/3D scans and ground-truth models for robotic manipulation. Control is executed in simulation with a fixed physics timestep and action decimation. The policy outputs normalized joint commands mapped to joint position targets.

\textbf{Goals and progress metric:} At the start of an episode (and optionally upon success), we sample a goal orientation $q_g \in SO(3)$. The task also defines a nominal in-hand reference position $p_{\mathrm{ref}}$ to which the object should remain close to encourage stable grasps and prevent drops. Progress toward the goal is measured on $SO(3)$ via the geodesic rotation distance $d_R(q, q_g)\in[0,\pi]$, implemented as the angle of the relative quaternion $q \otimes q_g^{*}$. An episode is considered successful when $d_R(q_t,q_g) < \beta$.

\textbf{Reward:} The reward encourages minimizing in-hand position drift and reducing rotational error, with action regularization and a success bonus:

\vspace{-0.95em}
\begin{align} 
r(s_t,a_t,g) \nonumber
= w_p \,\lVert &p_t - p_{\mathrm{ref}} \rVert_2 
 + w_R \,\frac{1}{d_R(q_t,q_g)+\epsilon}
 + w_a \,\lVert a_t \rVert_2^2 
 + 
\begin{cases}
b,  \quad \text{if } d_R(q_t,q_g) < \beta \\ \nonumber
0,  \quad \text{otherwise.}
\end{cases} \nonumber
\end{align} 
\vspace{-0.55em}

Episodes terminate early if the object moves too far from $p_{\mathrm{ref}}$ (drop/out-of-reach), and otherwise time out after 600 environment steps. This count is reset upon a successful reorientation. We set the maximum number of consecutive goal successes to 5.

% \subsection{Policy Implementation}
\label{sec:policy}
% \kar{Need a few sentences here about our basic policy and how SOTA policies are orthogonal to what we are trying to do. We might also have to nuance this where our approach might not provide as much of a benefit.}

We solve the reorientation MDP using PPO~\citep{schulman2017proximal} with an asymmetric actor-critic architecture. 

\noindent\paragraph{Observation Space}
The actor network receives an observation vector $o_t$ comprising:
\begin{itemize}
    \item \textbf{Learned embeddings}: The current object embedding $z_{\text{obj}} = \phi(\mathcal{P}_t)$ and the goal embedding $z_{\text{goal}} = \phi(\mathcal{P}_{\text{goal}})$ (128 dimensions each, 256 total).
    \item \textbf{Proprioception}: Joint positions, velocities, and hand-relative fingertip poses (133 dimensions).
    \item \textbf{Geometric metadata}: The centroids of the current and goal point clouds in the environment frame (6 dimensions total), providing explicit relative translation information to supplement the rotation-focused embeddings. 
\end{itemize}

\noindent\paragraph{Asymmetric Critic}
To facilitate faster convergence, we employ an asymmetric critic that has access to privileged state information $s_t$, including the ground-truth object state (pose, velocity), hand state, and physical contact forces. 
\begin{figure*}[!htbp]
    \centering
    \subfloat[Encoder Architecture\label{fig:enc_arch}]{
        \includegraphics[width=0.991\linewidth]{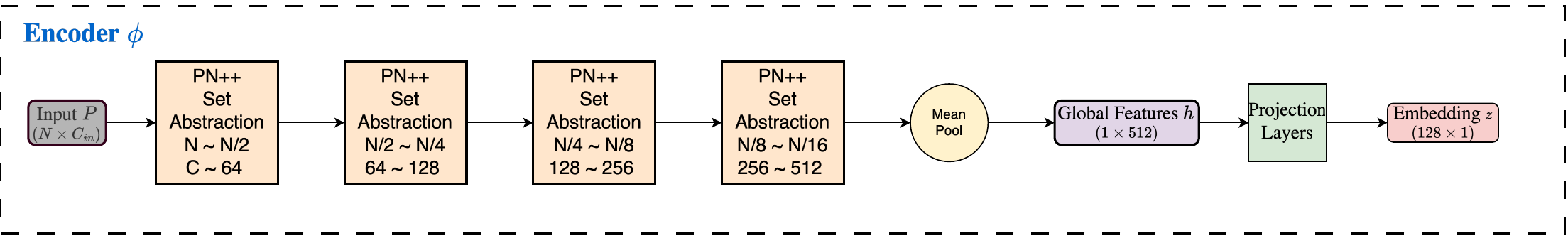}
    } \\
    \subfloat[Encoder Training \label{fig:enc_train}]{
        \includegraphics[width=0.270\linewidth]{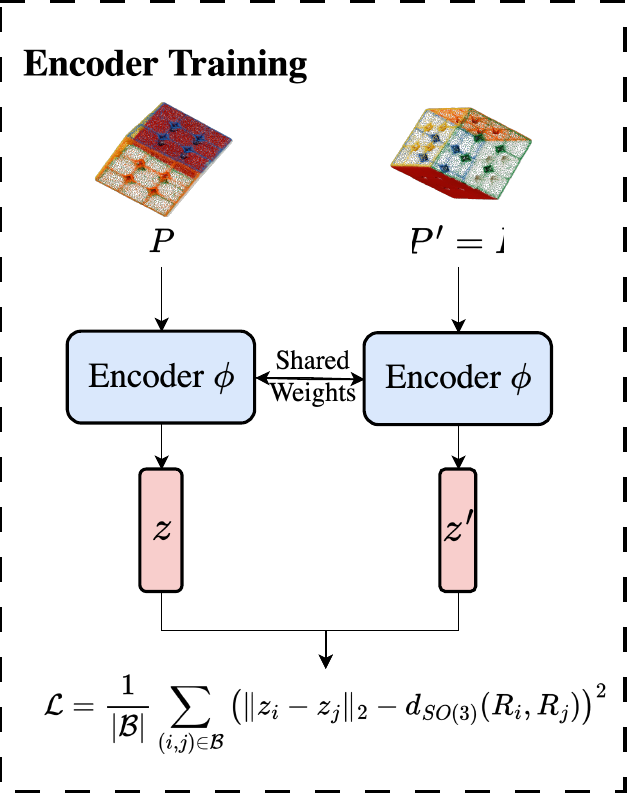}
    } 
    \subfloat[Policy Learning \label{fig:policy_train}]{
        \includegraphics[width=0.705\linewidth]{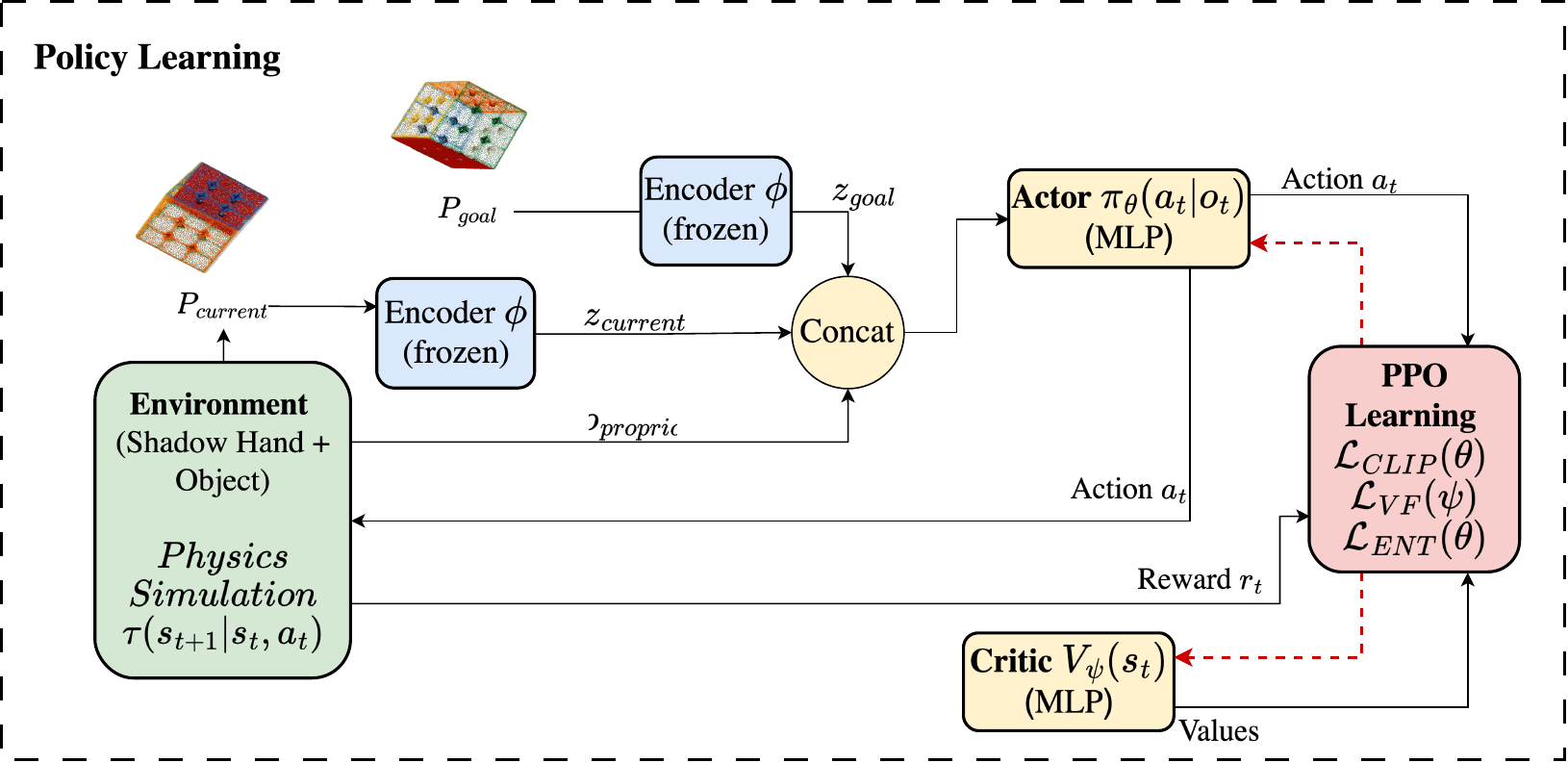}
    }
        % \centerin
        % \includegraphics[width=1\linewidth]{Media/latte_encoder.pdf}
        \caption{Overview of the method. (a) Encoder architecture. (b) Encoder training uses paired rotations of the same object and regresses $\|z_i-z_j\|_2$ to the SO(3) geodesic distance $d_{SO(3)}(R_i,R_j)$. (c) Policy learning freezes the encoder and feeds current/goal embeddings, centroids, and proprioception to an actor--critic trained with PPO $(\mathcal{L}_{CLIP}, \mathcal{L}_{VF}, \mathcal{L}_{ENT})$.}
        
        \label{fig:method}
\end{figure*}

\section{Results \& Evaluation}
\subsection{Policy Learning}

First, we evaluate whether a policy using our learned point-cloud representation can match policies that use more structured object representations. We compare against baselines inspired by teacher-student distillation~\cite{chen2022visualdexterity,chen2021system} and per-point dense flow~\cite{zhou2023hacman}. Both the baselines are implemented using the publicly available code. Further details are provided in Appendix~\ref{app:baselines}.

All methods are trained in the same simulated environment using IsaacLab with three $128 \times 128$ RGB-D cameras observing the hand from front-left, front-right, and top-rear views. For each camera, we convert the rendered depth image into 3D points using the pinhole camera model~\citep{szeliski2022computer} and use IsaacLab segmentation utilities to mask object points, producing three partial object point clouds. These are fused into the current object point cloud used by vision-based policies. The goal point cloud is obtained by sampling the canonical object mesh and rotating it to the target orientation. We also apply point jittering and random shuffling to the observed point cloud to better approximate real sensor point clouds.

Table~\ref{tab:policy_interfaces} summarizes the actor inputs. All policies receive proprioception and differ only in how object and goal information are represented: our method uses learned point-cloud embeddings, teacher baselines use pose information from simulator state or FoundationPose~\citep{wen2024foundationposeunified6dpose}, the distilled student uses raw current/goal point clouds, and flow baselines use dense displacement fields computed from simulator state or registration using standard open3d registration pipeline~\citep{open3dColoredPoint}.

We train separate per-object policies for four YCB objects: peach, pear, Rubik's cube, and tuna can, using RSL-RL~\citep{schwarke2025rslrl}. To isolate the representation question from object graspability, we choose objects for which the privileged teacher can learn reliable reorientation within approximately \(10^8\) environment interaction steps. In preliminary experiments, we observed that fingers often get stuck in looped objects such as mugs or scissors, small objects such as spoons or knives tend to fall out of the hand, and large objects such as the power drill are difficult to grasp and rotate.

We report the number of successful reorientations and the average number of timesteps required in 100 independent trials. Table~\ref{tab:success} presents the benchmark results. The policy trained with our representation consistently succeeds, keeping up with the distilled student, teacher, and flow baselines. However, it takes the longer to achieve success as indicated by the avg. timesteps column. During training, the first half of each run is dominated by maximizing the number of successes, after which the policy begins optimizing for speed. Our training runs are capped at \(10^8\) environment interaction steps. To save space we report the learning curves in the Appendix~\ref{app:learning_curves} which show that training for longer lets the policy learn faster motions even under our observations.
Furthermore, the significant gap between the privileged and non-privileged variants of the estimator- and registration-based baselines highlights the difficulty of recovering structured intermediate quantities in a deployment setting. As stated earlier, our approach is motivated by this observation: rather than imposing pose, registration, or dense correspondence as intermediate structure, we learn a representation in which current and goal point clouds can be compared directly at the embedding level. Our policy and the distilled operate using visual input directly letting them bypass the the brittle upstream modules like pose estimator and pointcloud registration.
\begin{table}[]
\centering
\caption{Policy interfaces evaluated for in-hand reorientation. Columns differ only in the actor's test-time observation.}
\label{tab:policy_interfaces}
\resizebox{\textwidth}{!}{%
\begin{tabular}{@{}c|l@{}}
\toprule
Method                                                        & \multicolumn{1}{c}{\begin{tabular}[c]{@{}c@{}}Actor input\\ Robot proprioception +\end{tabular}}                                                             \\ \midrule
Ours                                                          & Embeddings of current and goal point clouds                                                                                                                   \\ \midrule
\begin{tabular}[c]{@{}c@{}}Teacher\\ (sim state)\end{tabular} & Ground-truth current pose, goal pose, and relative orientation                                                                                                          \\ \midrule
\begin{tabular}[c]{@{}c@{}}Teacher\\ (estimator~\cite{wen2024foundationposeunified6dpose})\end{tabular} & Estimator-derived object pose, ground-truth goal pose, and estimator-derived relative orientation                                                                                 \\ \midrule
\begin{tabular}[c]{@{}c@{}}Distilled \\ Student~\citep{chen2022visualdexterity}\end{tabular}  & Current + goal point clouds (XYZ, RGB)                                                                                                                       \\ \midrule
\begin{tabular}[c]{@{}c@{}}Flow~\cite{zhou2023hacman}\\ (sim state)\end{tabular}    & \begin{tabular}[c]{@{}l@{}}Current XYZ + $\Delta x$, $\Delta y$, $\Delta z$ features derived by rotating the current point cloud using \\ sim state\end{tabular} \\ \midrule
\begin{tabular}[c]{@{}c@{}}Flow~\cite{zhou2023hacman}\\ (registration~\citep{open3dColoredPoint})\end{tabular} & Current XYZ + $\Delta x$, $\Delta y$, $\Delta z$ features derived using an Open3D registration pipeline                                                       \\ \bottomrule
\end{tabular}
}
\end{table}
\begin{table}[]
\centering
\caption{Benchmark results. Number of successful reorientations over 100 evaluation trials and average timesteps to reorient. Each trial consists of one randomly sampled target orientation.}
\label{tab:success}
\resizebox{\textwidth}{!}{%
\begin{tabular}{c|clclclclclcl}
\hline
\multirow{2}{*}{Object} & \multicolumn{12}{c}{Method}                                                                                                                                                                                                                                                                                                                                                                                                                                                                                                                                                                                                                                            \\ \cline{2-13} 
                        & \multicolumn{2}{c|}{Ours}                                                                                     & \multicolumn{2}{c|}{\begin{tabular}[c]{@{}c@{}}Teacher\\ (sim state)\end{tabular}}                            & \multicolumn{2}{c|}{\begin{tabular}[c]{@{}c@{}}Teacher\\ (estimator)\end{tabular}}                            & \multicolumn{2}{c|}{\begin{tabular}[c]{@{}c@{}}Distilled \\ Student\end{tabular}}                             & \multicolumn{2}{c|}{\begin{tabular}[c]{@{}c@{}}Flow\\ (sim state)\end{tabular}}                              & \multicolumn{2}{c}{\begin{tabular}[c]{@{}c@{}}Flow\\ (registration)\end{tabular}}       \\ \hline
\multicolumn{1}{l|}{}   & \multicolumn{1}{l|}{successes} & \multicolumn{1}{l|}{\begin{tabular}[c]{@{}l@{}}Avg.\\ timesteps\end{tabular}} & \multicolumn{1}{l|}{successes} & \multicolumn{1}{l|}{\begin{tabular}[c]{@{}l@{}}Avg.\\ timesteps\end{tabular}} & \multicolumn{1}{l|}{successes} & \multicolumn{1}{l|}{\begin{tabular}[c]{@{}l@{}}Avg.\\ timesteps\end{tabular}} & \multicolumn{1}{l|}{successes} & \multicolumn{1}{l|}{\begin{tabular}[c]{@{}l@{}}Avg.\\ timesteps\end{tabular}} & \multicolumn{1}{l}{successes} & \multicolumn{1}{l|}{\begin{tabular}[c]{@{}l@{}}Avg.\\ timesteps\end{tabular}} & \multicolumn{1}{l}{successes} & \begin{tabular}[c]{@{}l@{}}Avg.\\ timesteps\end{tabular} \\ \hline
Peach                   & 100                           & \multicolumn{1}{l|}{85.29}                                                    & 100                           & \multicolumn{1}{l|}{25.40}                                                    & 53                            & \multicolumn{1}{l|}{598.8}                                                    & 100                           & \multicolumn{1}{l|}{50.32}                                                    & 100                          & \multicolumn{1}{l|}{69.65}                                                    & 37                           & 930.11                                                   \\
Rubik's Cube            & 100                           & \multicolumn{1}{l|}{71.39}                                                    & 100                           & \multicolumn{1}{l|}{40.21}                                                    & 33                            & \multicolumn{1}{l|}{1556.0}                                                   & 100                           & \multicolumn{1}{l|}{41.58}                                                    & 100                          & \multicolumn{1}{l|}{47.53}                                                    & 33                           & 756.3                                                    \\
Pear                    & 100                           & \multicolumn{1}{l|}{116.76}                                                   & 100                           & \multicolumn{1}{l|}{48.27}                                                    & 58                            & \multicolumn{1}{l|}{942.1}                                                    & 100                           & \multicolumn{1}{l|}{52.32}                                                    & 100                          & \multicolumn{1}{l|}{137.05}                                                   & 22                           & 633.5                                                    \\
Tuna Can                & 100                           & \multicolumn{1}{l|}{154.05}                                                   & 100                           & \multicolumn{1}{l|}{92.36}                                                    & 48                            & \multicolumn{1}{l|}{1037.8}                                                   & 100                           & \multicolumn{1}{l|}{129.30}                                                   & 100                          & \multicolumn{1}{l|}{313.99}                                                   & 28                           & 835.2                                                    \\ \hline
\end{tabular}%
}
\end{table}

\subsection{Representation Diagnostics}
\vspace{-0.8em}
Next, we aim to measure how well our encoder actually distinguishes between rotated point clouds, in order to understand what signal is made available to the policy. As a comparison, we also trained a policy using an off-the-shelf self-supervised point-cloud encoder, Point-MAE~\citep{pang2022masked}, fine-tuned on the YCB objects. This policy completely failed to learn (as evidenced by the blue run in the learning curve in Appendix~\ref{app:learning_curves}.
To investigate this, we perform controlled rotation sweeps for each object. For every two-axis combination from \{x,y,z\}, we rotate the object over a grid of orientations and compute the distance between the corresponding point-cloud embeddings. We visualize these distances as surface plots, producing a landscape that reveals how the encoder organizes rotations in its latent space. We also generate the same distance landscapes for the point-cloud encoder used by the distilled student (using the L2 norm of embeddings since it jointly encodes current and goal), allowing us to compare our representation against both a generic self-supervised encoder and a task-trained distillation encoder.

 All figures in this section are for pear with additional examples for the remaining evaluation objects and special cases such as highly symmetric objects are provided in Appendix~\ref{app:angle_tsne}, Appendix~\ref{app:axis_delta_cosine}, Appendix~\ref{app:rotation_surface_plots}, and Appendix~\ref{app:symmetry_cases}

\begin{wrapfigure}{}{0.45\linewidth}
  \vspace{-0.6em}
  \centering
  \footnotesize\textbf{Ours}\\[0.2em]
  \includegraphics[width=\linewidth]{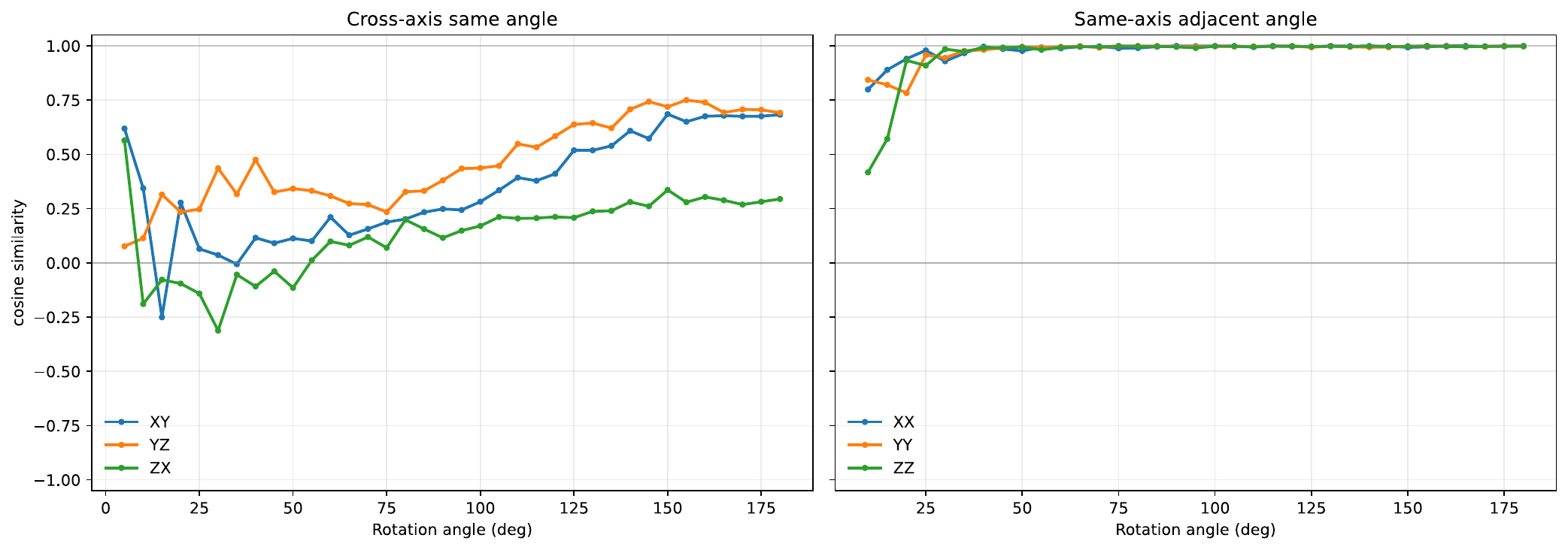}\\[0.25em]
  \textbf{PointMAE}\\[0.2em]
  \includegraphics[width=\linewidth]{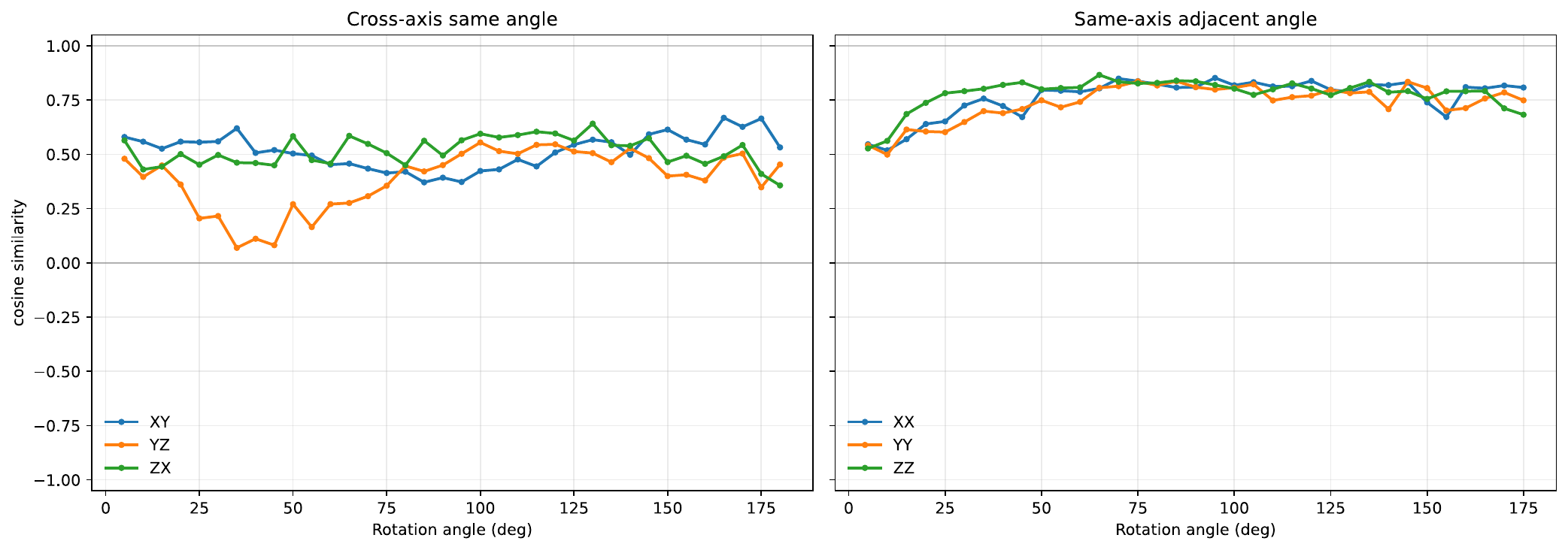}\\[0.25em]
  \textbf{Distilled Student}\\[0.2em]
  \includegraphics[width=\linewidth]{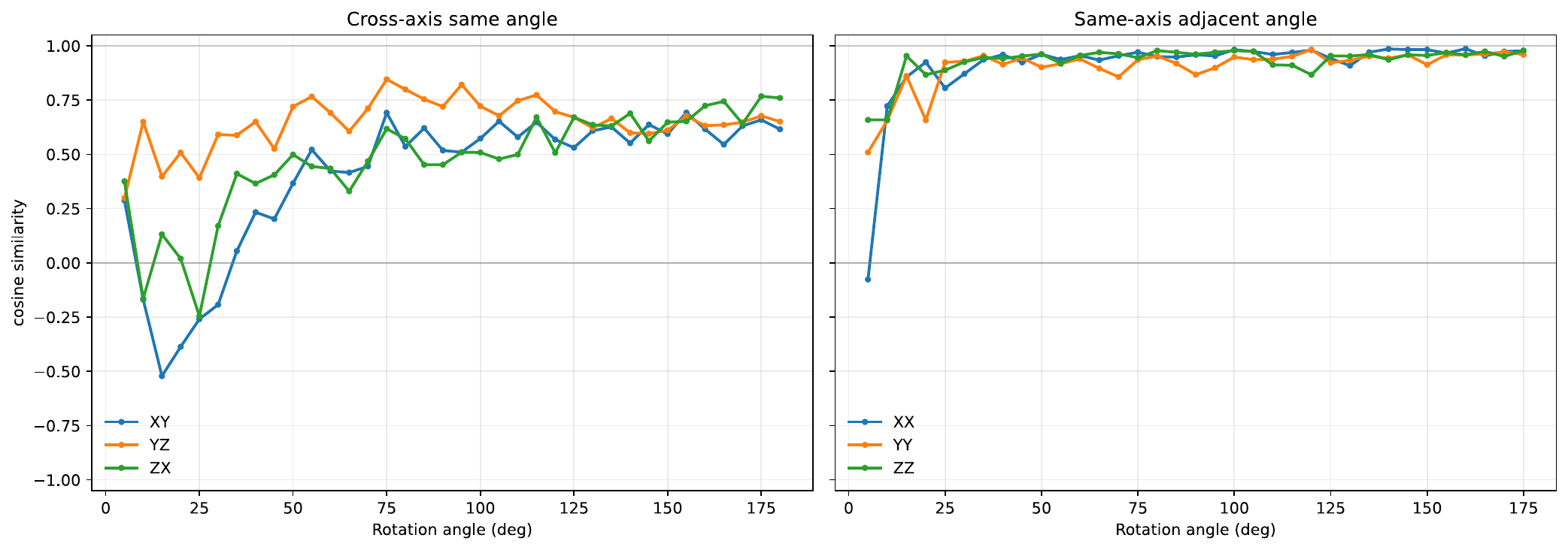}
  \caption{Cosine similarity between embedding deltas induced by rotations around different axes for pear.}
  \label{fig:axis_delta_cosine}
  \vspace{-1.0em}
\end{wrapfigure}
\begin{figure*}[!htbp]
  \centering
  \subfloat[Ours\label{fig:rotation_sweep_ours_pear}]{
    \includegraphics[width=0.9\linewidth]{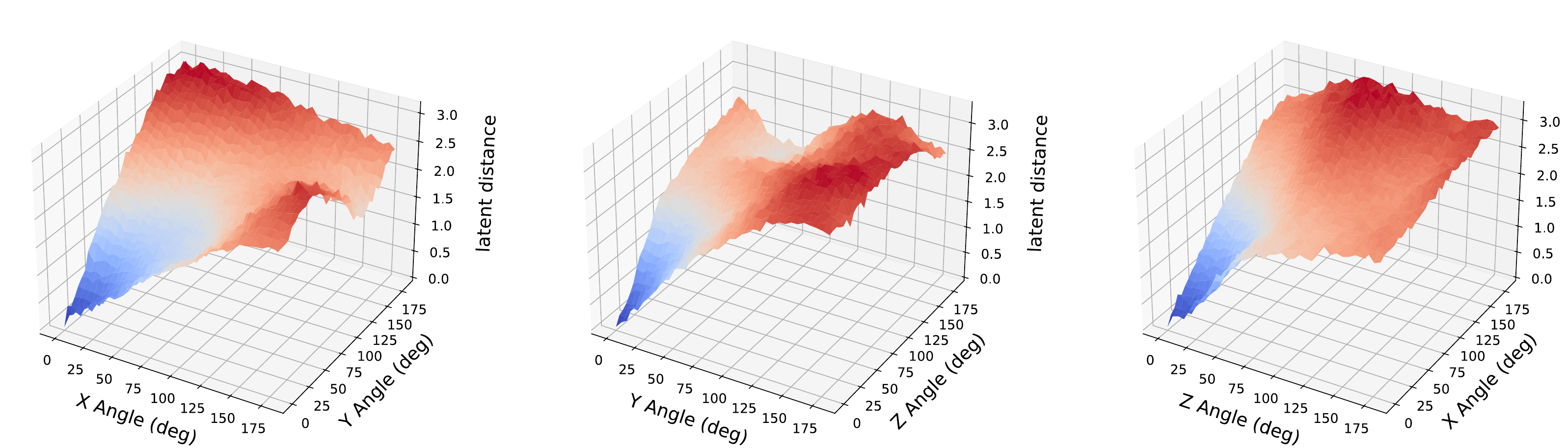}
  } \\
  \subfloat[PointMAE\label{fig:rotation_sweep_pointmae_pear}]{
    \includegraphics[width=0.9\linewidth]{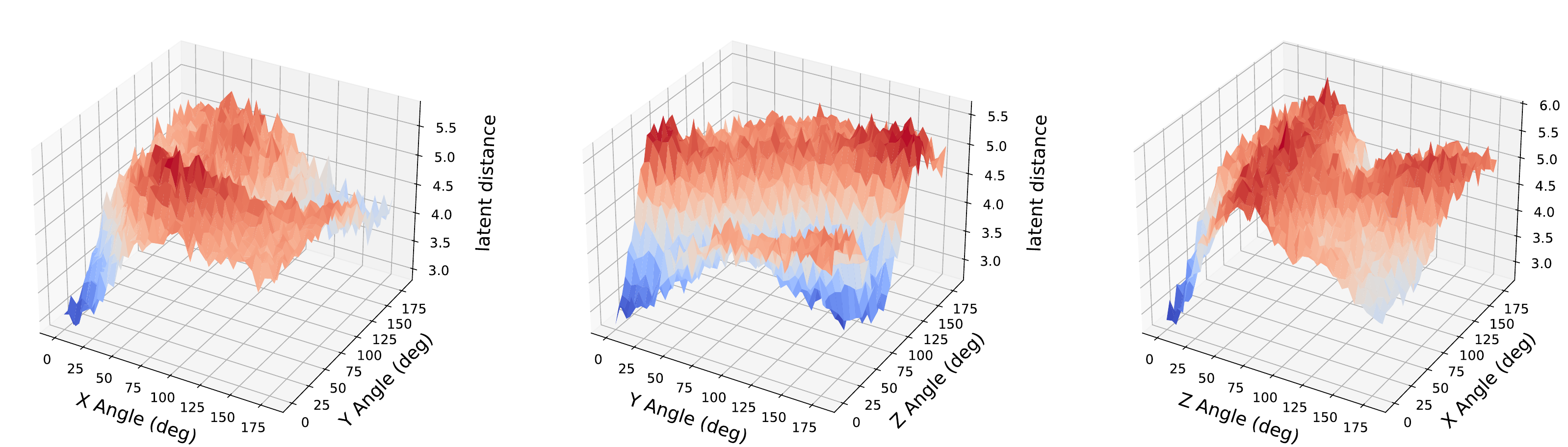}
  }\\
  \subfloat[Distilled Student\label{fig:rotation_sweep_student_pear}]{
    \includegraphics[width=0.9\linewidth]{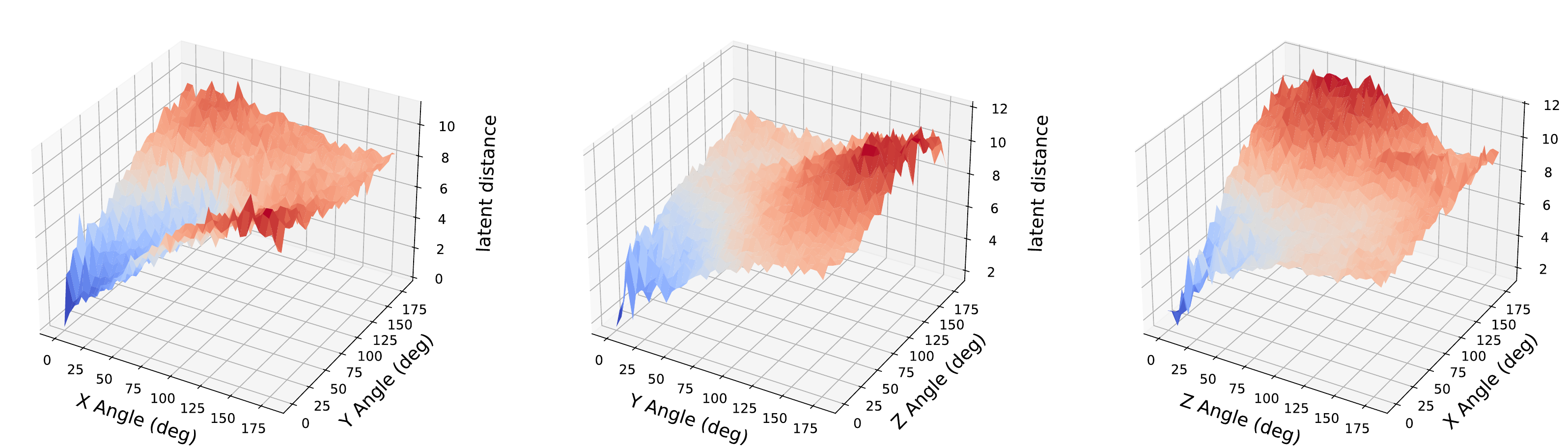}
  } \\
 \caption{Latent embedding distance as a function of simultaneous rotations around pairs of axes (X-Y, Y-Z, Z-X). Surfaces show how the encoders respond to 3D rotations of pear point clouds from 0° to 180°. Color scale: blue (low distance) to red (high distance).}
 \label{fig:rotation_sweep}
 \vspace{0.35em}
\end{figure*}

Figure~\ref{fig:rotation_sweep} shows these surface plots for pear; corresponding plots for peach, Rubik's cube, and tuna can are included in Appendix~\ref{app:rotation_surface_plots}. They clarify why Point-MAE fails: different relative orientations can induce similar distances, giving a policy little structured information about how far the object remains from the goal. In contrast, our metric-trained encoder produces a smoother and more structured delta-norm landscape. Surprisingly, this landscape is qualitatively similar to the one learned by the student encoder trained with action supervision through distillation. While the scales do not match, they both follow a similar trend of the norm increasing with angle. This suggests that the SO(3)-calibrated metric objective recovers much of the rotation-aware structure that an action-supervised student discovers indirectly from a privileged teacher, but without requiring teacher actions or robot interaction data for representation learning.

We also evaluate whether the embedding geometry captures information about the axis of rotation, since the same rotation angle can correspond to different physical rotations depending on the axis. A useful representation should therefore place embeddings of rotations about the same axis in a more consistent direction in latent space, while separating embeddings produced by rotations about different axes, even when the rotation angle is fixed.

\begin{wrapfigure}[10]{r}{0.28\linewidth}
  \vspace{-0.8em}
  \centering
  \includegraphics[width=\linewidth]{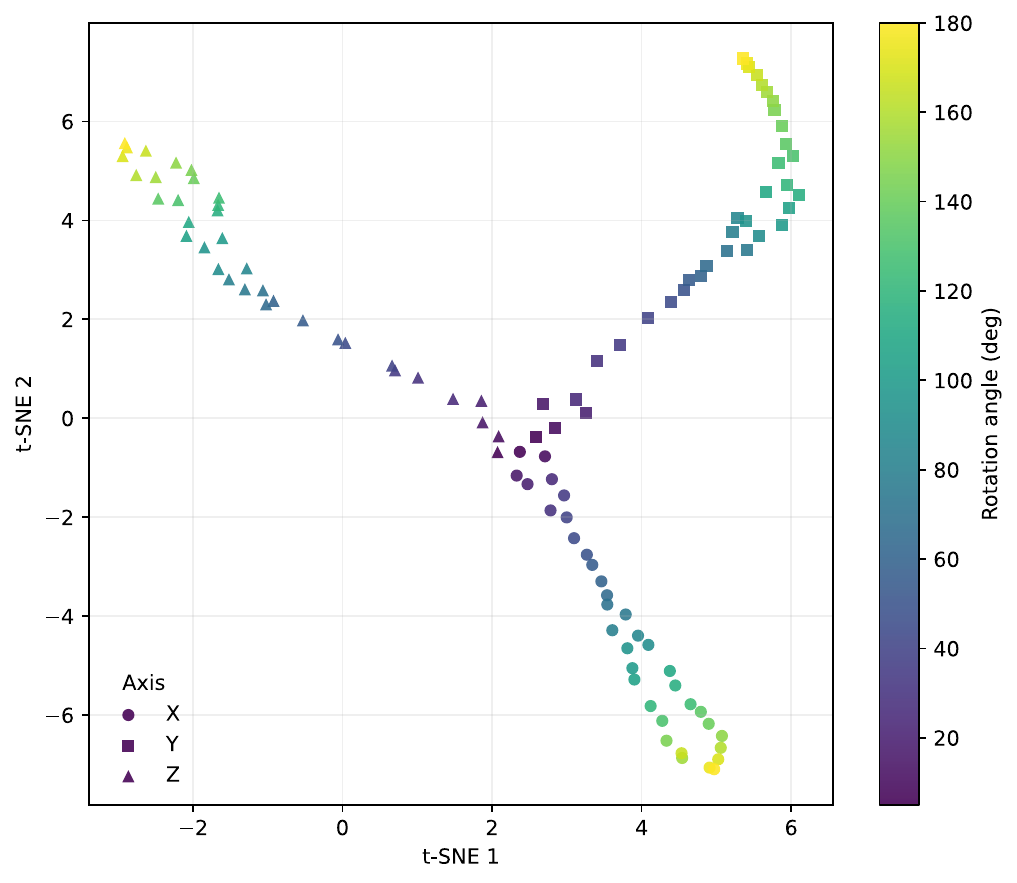}
  \caption{Pear t-SNE.}
  \label{fig:angle_tsne_pear}
  \vspace{-0.8em}
\end{wrapfigure}
To test this, we perform two controlled diagnostics. In the first, we rotate the object by the same angle about different axes and compute the cosine similarity between the resulting embeddings. In the second, we rotate the object by different angles about the same axis and again measure cosine similarity between the embeddings. This allows us to test whether embeddings are organized by rotation axis, rotation magnitude, or both. Figure~\ref{fig:axis_delta_cosine} shows the result of this test. All three encoders have low cosine similarity between rotations along different axes, but only our encoder and the student encoder correctly organize same-axis rotations together as shown by the cosine similarity being $\approx1$. At smaller rotations axes become degenerate this the low early similarity
These results clearly indicate that a generic reconstruction-based, self-supervised encoder may not be well suited for robotic tasks beyond recognizing shape. While PointMAE is sensitive to rotation perturbations, the signal does not contain structured information beyond the fact that a change exists.

Finally, we provide a qualitative analysis of the learned representation using t-SNE visualizations~\citep{van2008visualizing} of the embeddings. They show a clear structure in which rotations are meaningfully separated by angle and axis. At smaller rotations, the distinction disappears, as near-zero rotations around any axis are equivalent. This pattern persists across other objects in Appendix~\ref{app:angle_tsne} (Figure~\ref{fig:appendix_angle_tsne}).
\FloatBarrier

\section{Related Work}
\label{sec:related}
\textbf{\emph{Goal Representations in Manipulation}}
A large body of work exploits the fact that image observations and image goals share a grid, enabling goal-reaching via learned visual representations, latent-distance rewards, and goal relabeling. RIG \cite{nair2018rig} is a representative approach that learns a visual latent space and uses latent distance as reward, with relabeling to turn generic experience into goal-reaching data. Skew-Fit \cite{pong2019skew} adjusts goal sampling to emphasize harder outcomes. This line builds on hindsight relabeling as a general mechanism for goal-conditioned RL such as HER \cite{andrychowicz2017her} and its curriculum variants \cite{ren2019hgg}, and it connects to contrastive formulations of goal-reaching \cite{eysenbach2022contrastive_gcrl} and unsupervised exploration followed by solving arbitrary image goals \cite{mendonca2021lexa}. Complementarily, visual-foresight model-based methods plan toward goal images or designated pixels using learned dynamics, again leveraging the shared 2D coordinate system to make progress measurable \cite{finn2017visual_foresight,ebert2018retrying,xie2019tools_visual_foresight}. Image-goal methods have also been used for dexterous reorientation~\citep{openai2019learningdexterousinhandmanipulation}. Language enables flexible goal specification where systems either learn grounding and control jointly, or they use intermediate structures that keep execution grounded in perception. GRIF \cite{myers2023grif} leverages abundant image-goal data so language can steer goal-conditioned visuomotor behavior without requiring fully labeled language demonstrations, while RT-H \cite{belkhale2024rth} induces a hierarchy of reusable skills guided by language but grounded in perception during execution. More broadly, many language-to-perceptual-target designs fit the same pattern of restoring a comparison-friendly space before control.

% \textbf{\emph{3D representations for manipulation:}} Many recent systems incorporate explicit 3D structure—voxels, point clouds, or learned 3D feature fields to make spatial reasoning more direct and robust to viewpoint. PerAct \cite{shridhar2022peract} uses voxelized 3D inputs for action prediction. VoxPoser \cite{huang2023voxposer} uses 3D voxel value maps as a planning/grounding intermediate from language. PolarNet \cite{chen2023polarnet} and Act3D \cite{gervet2023act3d} use point clouds or 3D feature-field transformers for language-guided and multi-task manipulation and PerAct2 \cite{peract2} extends voxel-style action prediction and benchmarking to bimanual tasks. Generative action modeling has recently been combined with 3D scene representations to capture multimodal manipulation behaviors. 3D Diffuser Actor \cite{ke2024_3ddiffuseractor} conditions diffusion on 3D scene structure (with instruction/proprioception) to generate trajectories; Bi3D Diffuser Actor \cite{ke2024_bi3d} extends this to coordinated bimanual generation. These works support the value of explicit 3D, but in most cases the goal is language, sparse targets, or demonstrations rather than a \emph{3D goal object} directly compared to the current 3D observation.

\textbf{\emph{Point-cloud goals for manipulation:}} While there is considerable work using point clouds or 3D observations as policy inputs~\cite{shridhar2022peract,huang2023voxposer,chen2023polarnet,gervet2023act3d,ke2024_3ddiffuseractor,ke2024_bi3d,wan2025dexremoe,zhu2024point}, the set of works that specifically uses point clouds as goals is fairly limited. As mentioned previously, all of these methods add some external structure. HACMan \cite{zhou2023hacman} uses correspondence/flow-style signals to relate current and goal geometry, and SculptDiff \cite{bartsch2024sculptdiff} uses goal-conditioned policies over 3D shape targets for deformable manipulation. In dexterous in-hand settings, point clouds are used to specify object shapes for scaling and generalization, but explicit pose observations and targets are provided, as in DexReMoE \cite{wan2025dexremoe} and Geometry-Dex~\cite{huang2021generalization}. To the best of our knowledge, the only prior works at the time of writing that specify observations and goals only as point clouds to  train a pure vision only policy in an in-hand reorientation setting use privileged-teacher distillation~\citep{chen2022visualdexterity,chen2021system}. They achieve success by first training a teacher with privileged knowledge and then distilling it into a vision-only student. Our work differs by learning a representation that makes it possible to learn in a vision-only setting without using model-free RL explicit correspondences, pose inputs, or privileged supervision.

\section{Limitations and Future Work}
 In this work, we achieve our goal of learning policies that operate directly from point-cloud goals, without explicit pose estimation, or registration. However, our learned embedding primarily preserves rotation-relevant state information and is not explicitly constrained to preserve object shape. This is sufficient for the per-object policies evaluated here, but encoder should retain both object geometry and control-relevant state information to enable learning a general policy. Future work should therefore explore embeddings that jointly preserve shape and enable control-friendly current-goal comparison. Additionally, our method is currently validated only for SO(3) reorientation; extending it to general manipulation would require adapting the metric objective to capture transformations over the full SE(3) group.

While our evaluation is conducted in simulation, this choice is intentional: the goal of this work is to isolate the representation question without confounding it with camera calibration, segmentation errors, controller latency, or contact-model mismatch in a real deployment stack. To make the setting less idealized, the policy operates only on point clouds generated from rendered RGB-D observations under hand occlusion, with point jittering and random shuffling to approximate real sensor noise. Since the policy relies on 3D spatial structure rather than raw RGB appearance, it is less dependent on photorealistic texture and lighting cues, consistent with prior work on point-cloud robustness to visual and geometric variation~\cite{zhu2024point}. We further test robustness using domain randomization over object scale, mass, contact properties, and action noise, with the corresponding training curves provided in Appendix~\ref{app:learning_curves}. Although this does not constitute a real-robot transfer demonstration, the benchmark results and similar learning trend under randomized dynamics provide encouraging evidence that the representation remains effective under sim-to-real-relevant perturbations and motivates future real-world validation.

\section{Conclusion} 
We presented a rotation-aware point-cloud embedding for vision-based in-hand reorientation. By calibrating latent distances to SO(3) geodesic error, the representation allows a PPO policy to compare current and goal point clouds without explicit pose, registration, dense flow, or teacher-action supervision. Experiments show that this interface achieves successful reorientation across the evaluated objects and remains competitive with privileged and distillation-based baselines, while diagnostics show that generic point-cloud pretraining does not provide the same rotation-structured signal. These results suggest that point-cloud goals become useful for dexterous control when the representation itself encodes task-relevant rotational geometry.

% In this paper, we argued that point-cloud goals for dexterous reorientation become practical only if the goal-comparison signal itself is smooth and aligned with rigid-body rotation, rather than an implicit registration problem. To that end, we learned a rotation-aware point-cloud encoder whose Euclidean embedding distance is calibrated to the SO(3) geodesic misalignment between orientations, then froze that encoder and trained a standard PPO policy that conditions on current/goal embeddings plus proprioception (and simple centroid metadata). 

% Empirically, this task-aligned representation substantially stabilizes and accelerates model-free RL compared to a strong task-agnostic point-cloud baseline (Point-MAE), and reaches the same final performance as a pose-conditioned reference despite using no pose inputs. Offline diagnostics explain the improvement: latent distances track SO(3) error with a strong linear relationship and yield smooth, globally structured distance landscapes, giving PPO a controllable notion of progress where generic encoders can be jagged or flat. 

% Finally, our approach occupies a complementary point to explicit pose estimation: it preserves rotation-relevant structure without requiring pose at inference time, which can be attractive when robust pose is difficult to obtain. Key next steps are handling partial views and hand occlusions (e.g., completion, augmentation, history-aware policies) and extending beyond object-specific policies by adding an object-identity signal alongside the rotation-aware embedding

\clearpage

% \acknowledgments{}

\bibliography{references}

\clearpage

\appendix
\section*{Appendix}
\renewcommand{\thesection}{A\arabic{section}}
\setcounter{section}{0}
\section{Evidence for Sim-to-Real Transfer}
Our experiments are implemented in IsaacLab with high-fidelity physics and photorealistic rendering. However, the policy does not receive privileged object geometry or simulator state; it operates on point clouds generated from rendered RGB-D observations under hand occlusion. This reduces dependence on photorealistic texture, lighting, and background cues, since the policy primarily uses 3D spatial structure, consistent with prior work on point-cloud robustness to visual and geometric variation~\cite{zhu2024point}. The meshes in the YCB dataset are generated from real scanned objects rather than perfect synthetic CAD models, and we apply point jittering and random shuffling to better approximate real sensor point clouds.

This work focuses on isolating the representation question: whether point-cloud goal embeddings provide a control-useful state and goal representation without explicit pose estimation, registration, or dense correspondence. To test robustness under sim-to-real-relevant perturbations, we additionally train with domain randomization over object scale, mass, contact properties, and action noise. The randomized training runs show a learning trend similar to the non-randomized setting. Although some of the randomized runs have not fully converged, this trend, together with the benchmark results in Table~\ref{tab:success}, suggests that the representation remains effective under substantially perturbed simulation conditions and motivates future real-robot validation.

\section{Encoder Training Details}\label{app:train}
The PointNet++ pose encoder used by our policy is trained offline before RL.
Each training example contains two colored YCB point clouds of the same object,
sampled at known orientations $q_1$ and $q_2$. The model receives $N=1024$
points with $(x,y,z,r,g,b)$ channels, centers each cloud, normalizes XYZ to a unit
sphere, and predicts one embedding per cloud. The training target is to match the Euclidean distance between the embeddings to the geodesic distance as per Equation~\ref{eq:loss} which is restated below for completenesss. 
relative rotation:
\begin{align}
    \mathcal{L}_{\text{metric}}=\mathbb{E}[(\|\phi(\mathcal{P}_1)-\phi(\mathcal{P}_2)\|_2 - 2\arccos|\langle q_1,q_2\rangle|)^2]
\end{align}

The offline PointNet++ encoder uses four set-abstraction stages. In the main
configuration all stages keep a sampling ratio of $0.5$, use radius grouping,
and use up to 64 neighbors per query. The radii are
$(0.15,0.35,0.50,1.0)$ for stages 1--4. The corresponding MLP channel sizes
are $(6,64,64)$, $(67,128,128)$, $(131,192,512)$, and $(515,512,512)$, where
the extra three channels after the first stage are relative XYZ inputs used by
PointNet++ set abstraction. After mean pooling, 512-dimensional global descriptor is mapped bya two-layer head $512\rightarrow256\rightarrow128$, producing the 128-dimensional
embedding used by the policy. 

We first train on YCB mesh point clouds using controlled relative-rotation
pair sampling. The main configuration uses 16 relative-angle bins and 600 pairs
per object per epoch. To improve robustness to partial observations, we use a
four-stage occlusion curriculum: the first 10 epochs are unoccluded, followed
by 10-epoch stages with occlusion probabilities $0.3$, $0.6$, and $0.8$.
Occlusion is applied to one randomly selected side of the pair. After this mesh-based stage, we fine-tune on a rendered-cloud dataset recorded from IsaacLab.

\begin{figure*}[t]
  \centering
  \small
  \begin{tabular}{@{}c@{\hspace{0.01\linewidth}}c@{\hspace{0.01\linewidth}}c@{\hspace{0.01\linewidth}}c@{}}
    \textbf{No occlusion} & \textbf{Stage 1} & \textbf{Stage 2} & \textbf{Stage 3} \\
    \includegraphics[width=0.23\linewidth]{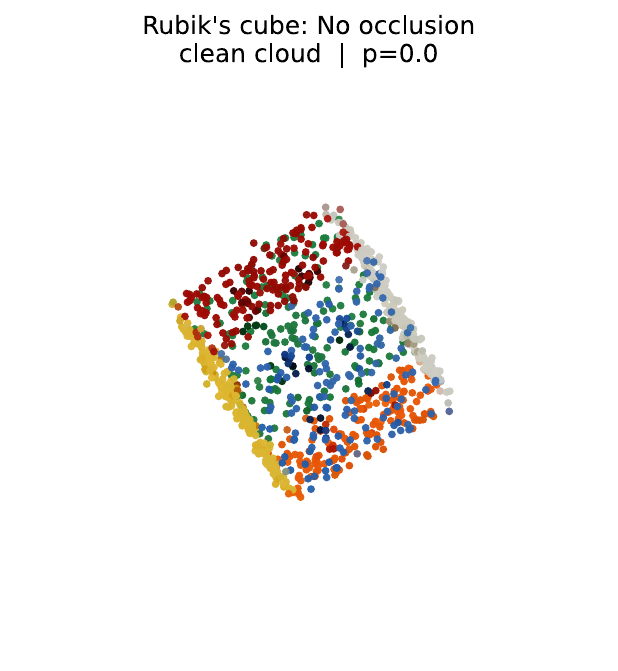} &
    \includegraphics[width=0.23\linewidth]{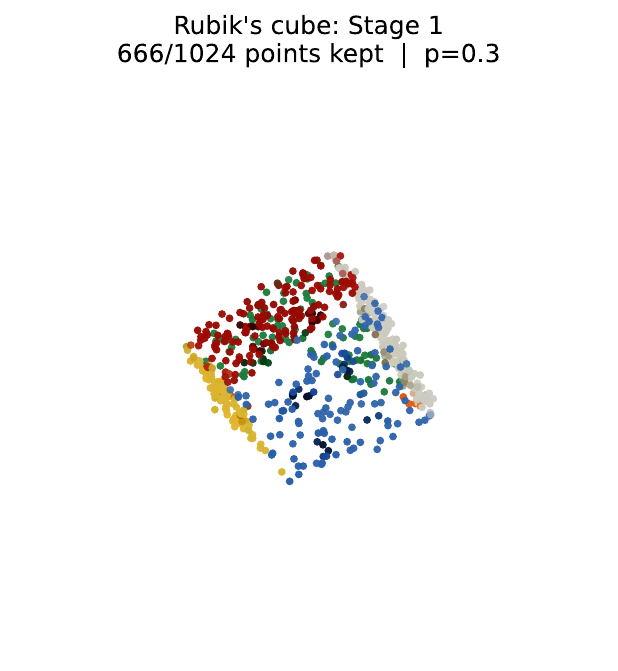} &
    \includegraphics[width=0.23\linewidth]{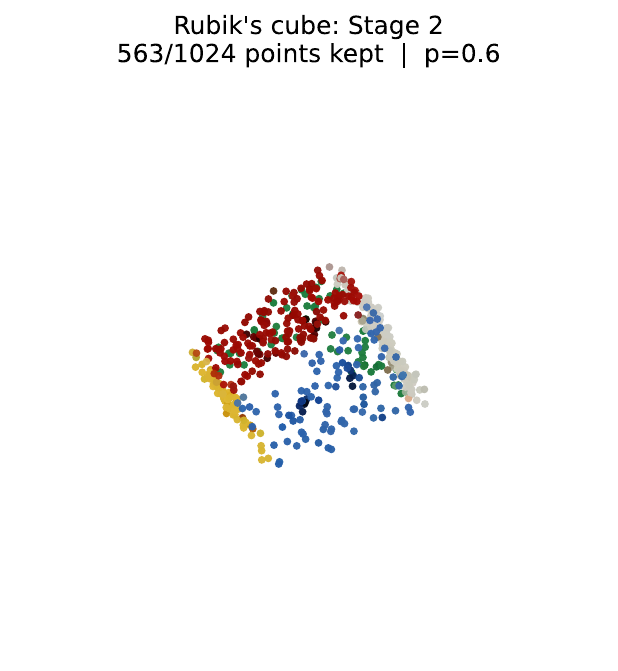} &
    \includegraphics[width=0.23\linewidth]{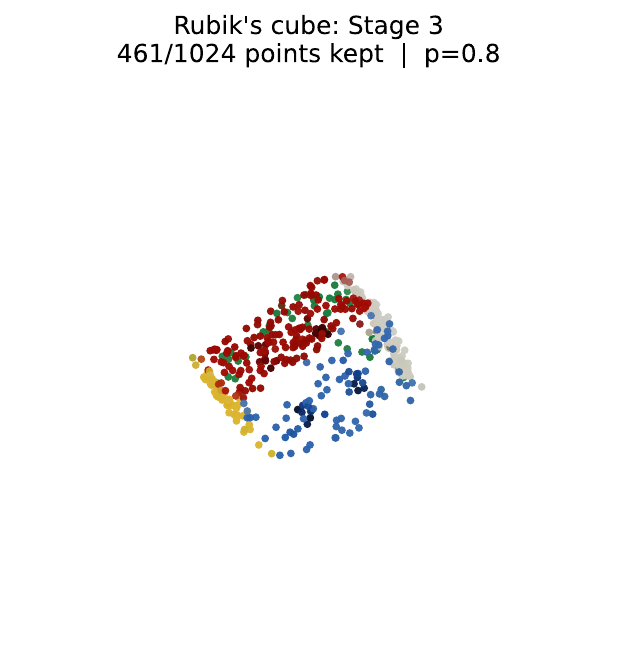} \\
    \includegraphics[width=0.23\linewidth]{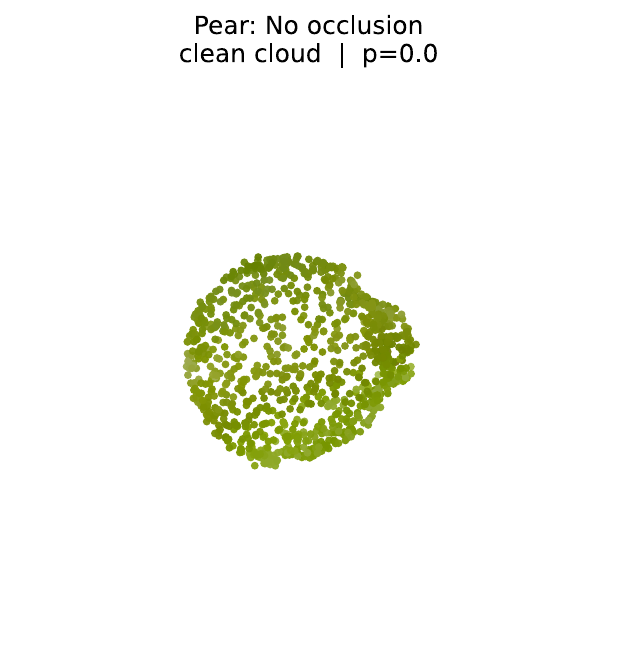} &
    \includegraphics[width=0.23\linewidth]{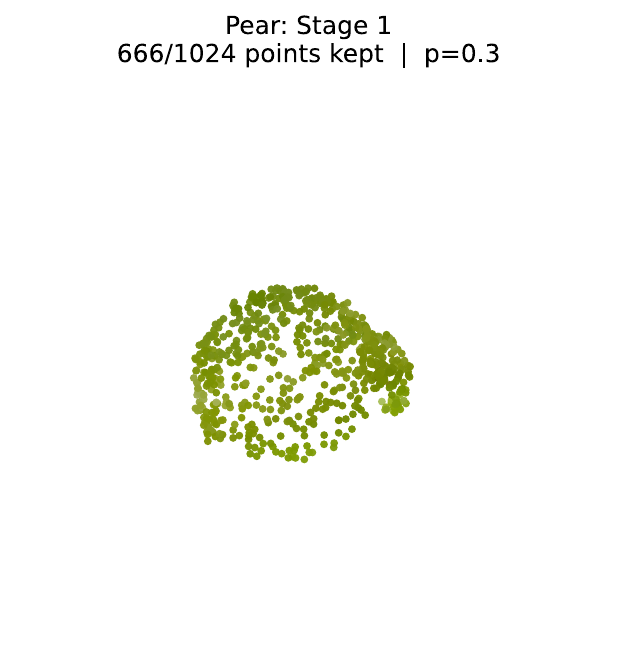} &
    \includegraphics[width=0.23\linewidth]{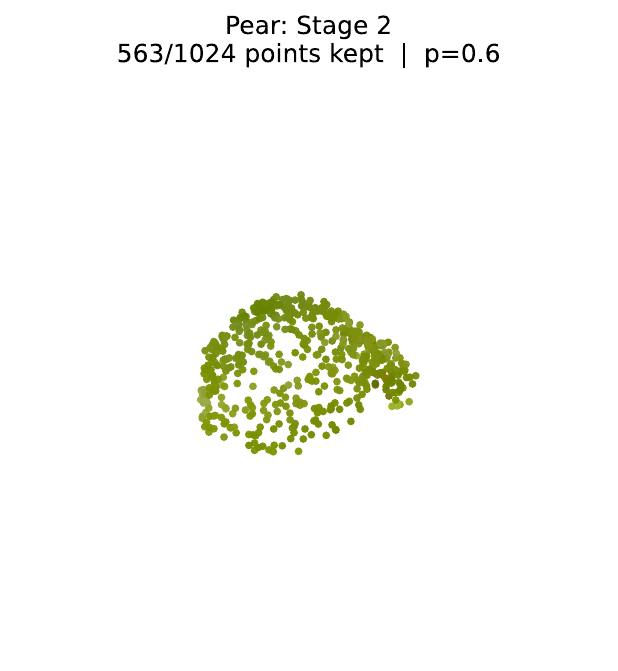} &
    \includegraphics[width=0.23\linewidth]{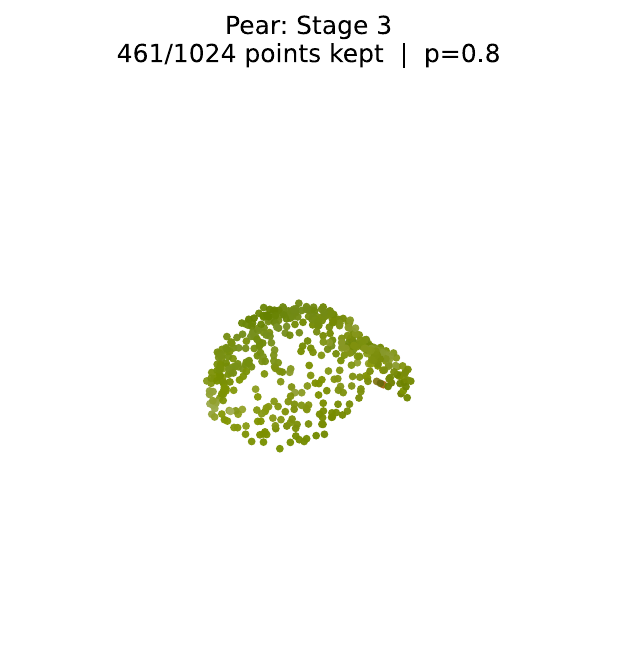}
  \end{tabular}
  \caption{Occlusion curriculum visualizations for Rubik's cube (top row) and pear (bottom row). Columns show the unoccluded mesh-cloud input followed by the three occlusion stages}
  \label{fig:occlusion_curriculum}
\end{figure*}

\begin{figure*}[t]
  \centering
  \small
  \begin{tabular}{@{}c@{\hspace{0.01\linewidth}}c@{\hspace{0.01\linewidth}}c@{\hspace{0.01\linewidth}}c@{}}
    \textbf{Sample 1} & \textbf{Sample 2} & \textbf{Sample 3} & \textbf{Sample 4} \\
    \includegraphics[width=0.23\linewidth]{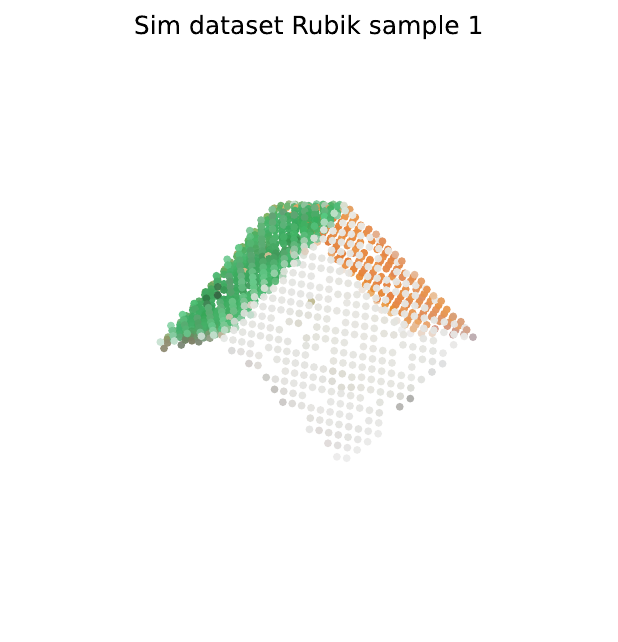} &
    \includegraphics[width=0.23\linewidth]{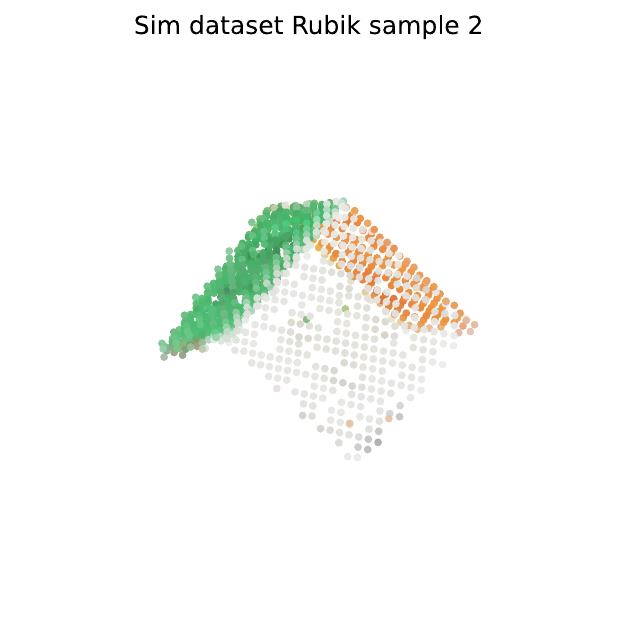} &
    \includegraphics[width=0.23\linewidth]{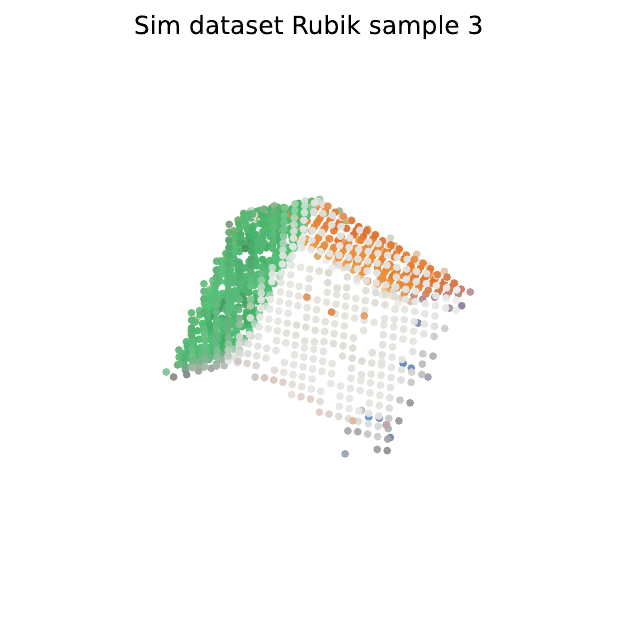} &
    \includegraphics[width=0.23\linewidth]{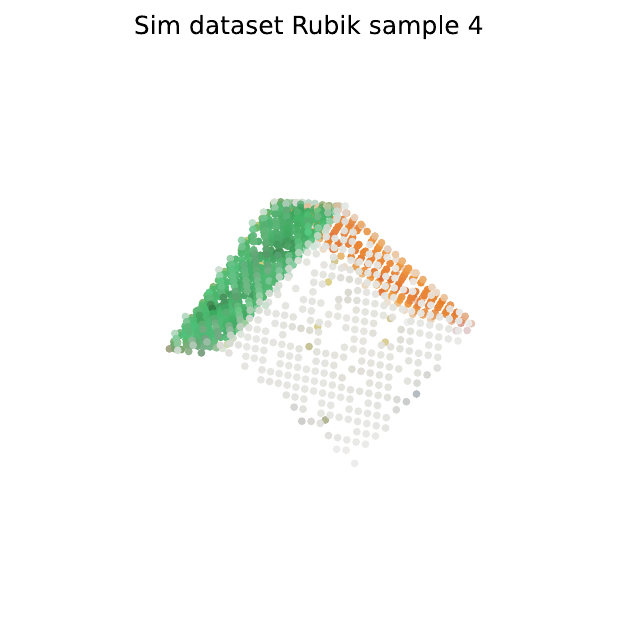} \\
    \includegraphics[width=0.23\linewidth]{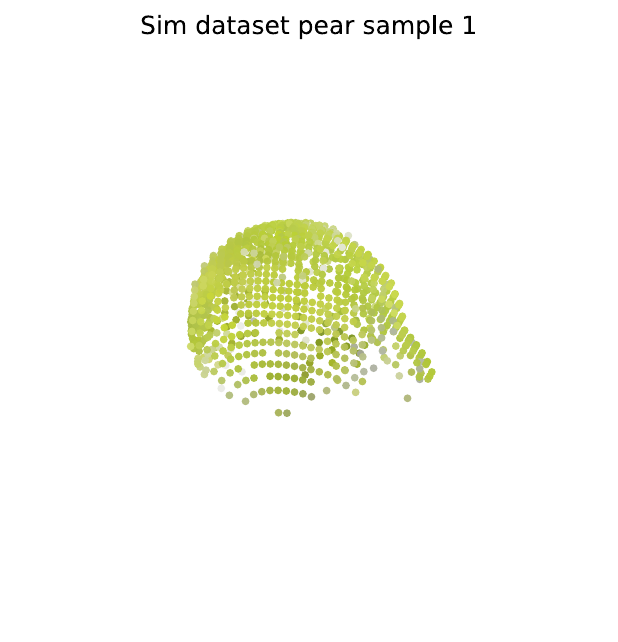} &
    \includegraphics[width=0.23\linewidth]{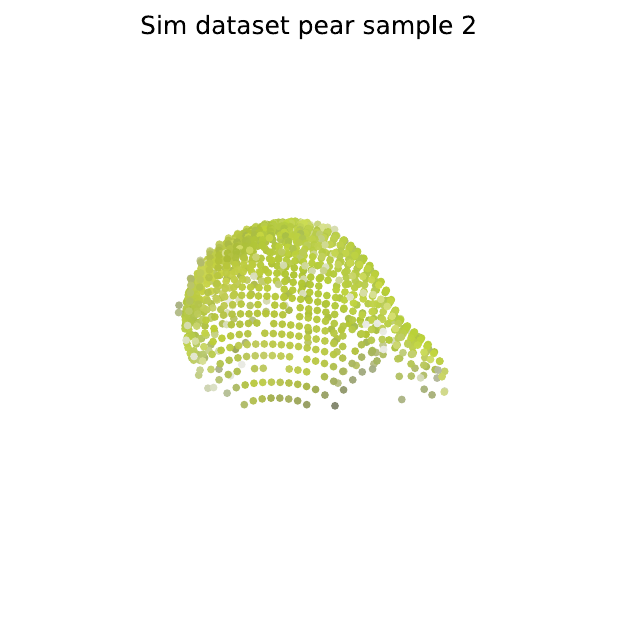} &
    \includegraphics[width=0.23\linewidth]{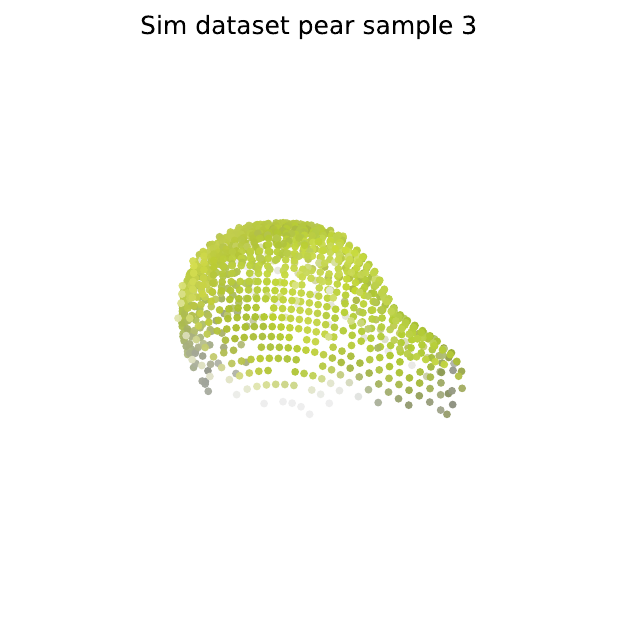} &
    \includegraphics[width=0.23\linewidth]{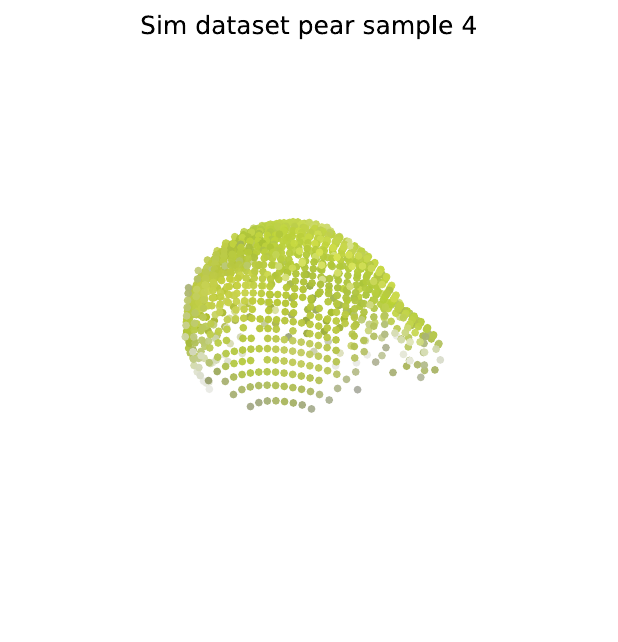}
  \end{tabular}
  \caption{Rendered RGB-D fine-tuning examples for Rubik's cube (top row) and pear (bottom row)}
  \label{fig:rendered_finetune_occlusion}
\end{figure*}

Both stages use Adam with initial learning rate $10^{-3}$ and no weight decay and cosine annealing with a linear warmup.
The curriculum progression and rendered fine-tuning observations are visualized
in Figures~\ref{fig:occlusion_curriculum} and~\ref{fig:rendered_finetune_occlusion}.

\clearpage

\section{Policy and Baseline Architecture Details}\label{app:baselines}
All direct PPO policies use the same Shadow Hand task, reward, action space,
and RSL-RL optimization settings unless noted otherwise. For the MLP-only
policies, both actor and critic are feedforward networks with hidden widths
$(1024,512,256,128)$, ELU activations, empirical observation normalization,
and a Gaussian action distribution initialized with unit standard deviation.
The PPO runner uses 64 steps per environment, 3500 iterations, five learning
epochs per update, four minibatches, learning rate $5\times10^{-4}$, clipped
value loss, entropy coefficient $0.005$, $\gamma=0.99$, $\lambda=0.95$,
desired KL $0.016$, and gradient clipping at 1.0. 
For asymmetric actor-critic training, the critic receives a privileged state vector $\mathbf{s} \in \mathbb{R}^{190}$ composed of the full simulator state (187-dim) plus the object--goal relative translation in the goal frame (3-dim). The full state includes:
\begin{itemize}
    \item Joint positions and velocities ($24 + 24$),
    \item Object pose (position~3, orientation~4), linear/angular velocity ($3 + 3$),
    \item Goal position~(3) and orientation~(4),
    \item Relative rotation $\mathbf{q}_{\mathrm{obj}} \otimes \mathbf{q}_{\mathrm{goal}}^{*}$~(4),
    \item Fingertip poses (position~15, orientation~20, velocity~30),
    \item Fingertip force/torque sensor readings~(30),
    \item Previous actions~(20).
\end{itemize}
Fingertip poses are transformed to the hand-root frame and all quaternions are canonicalized ($w \geq 0$).
All methods share a \emph{proprioceptive} vector $\mathbf{o}_{\mathrm{proprio}} \in \mathbb{R}^{133}$ consisting of:
\begin{itemize}
    \item Scaled joint positions $\tilde{\mathbf{q}} \in \mathbb{R}^{24}$
          (linearly rescaled to $[-1,1]$ by joint limits),
    \item Scaled joint velocities $\alpha \dot{\mathbf{q}} \in \mathbb{R}^{24}$,
    \item Fingertip positions in hand frame $\mathbf{p}^H_f \in \mathbb{R}^{15}$,
    \item Fingertip orientations in hand frame $\mathbf{q}^H_f \in \mathbb{R}^{20}$
          (canonicalized $w \geq 0$),
    \item Fingertip linear and angular velocities $\mathbf{v}_f \in \mathbb{R}^{30}$,
    \item Previous actions $\mathbf{a}_{t-1} \in \mathbb{R}^{20}$.
\end{itemize}

\subsection{Our Frozen PointNet++ Pose Embedding Policy}
Our policy uses the frozen encoder described in Appendix~\ref{app:train}. The actor
observation is
\begin{equation}
    [o_{\mathrm{prop}}, z_{\mathrm{cur}}, z_{\mathrm{goal}},
      c_{\mathrm{cur}}, c_{\mathrm{goal}}],
\end{equation}
where $o_{\mathrm{prop}}\in\mathbb{R}^{133}$ is proprioception,
$z_{\mathrm{cur}},z_{\mathrm{goal}}\in\mathbb{R}^{128}$ are independently
computed point-cloud embeddings, and
$c_{\mathrm{cur}},c_{\mathrm{goal}}\in\mathbb{R}^{3}$ are pre-normalization
cloud centroids. The actor input is therefore 395-dimensional. For training and inference (benchmark), $z_{\mathrm{cur}}$ is computed from the
fused multicamera RGB-D current cloud and $z_{\mathrm{goal}}$ is cached from a
canonical textured YCB mesh sampled at the target orientation. The encoder is
never updated during PPO; only the policy MLP is trained.

\subsection{Privileged Teacher Policy}
The teacher is trained using privileged simulator state directly as an input to the actor. The actor receives 133-dimensional proprioception together with object pose, goal pose, and relative pose features, for a 154-dimensional input. The pose features contain object position and quaternion, goal position and quaternion, relative translation expressed in the goal frame, and relative quaternion. The estimator-based pose baseline uses the same actor architecture and pose-feature interface, replacing the ground-truth current pose fields with pose estimated from FoundationPose~\cite{wen2024foundationposeunified6dpose} outputs at test time.

\subsection{Flow PointNet++}
The flow baseline,implemented as per~\citep{zhou2023hacman} keeps PPO training but moves the point-cloud encoder inside
the actor. The environment returns two observation groups: 133-dimensional
proprioception, a fixed-size point-cloud tensor $\mathbb{R}^{1024\times6}$. The tensor contains up to 1024 valid rendered current points. The first three point
channels are the rendered current XYZ coordinates, centered and scaled per
environment over valid points. The last three channels are a metric displacement field. In the simulator-state flow
variant, each current point $x$ is transformed by the ground-truth rigid
current-to-goal transform,
\begin{equation}
    x_{\mathrm{goal}} =
    R_{\mathrm{goal}}R_{\mathrm{cur}}^\top(x-t_{\mathrm{cur}})
    + t_{\mathrm{goal}},
\end{equation}
and the flow feature is $x_{\mathrm{goal}}-x$. The registration variant uses
the same actor interface after replacing this displacement field with
registration-derived flow.

The flow actor encodes this padded $\mathbb{R}^{1024\times6}$ cloud with a PointNet++ tower whose set-abstraction stages have channel sizes
$(6,64,64)$, $(67,128,128)$, $(131,192,256)$, and
$(259,256,256)$. The implementation uses radius-query grouping with a
32-neighbor cap, stage sampling ratios $(0.5,0.5,0.5,1.0)$, and mean pooling.
The 256-dimensional cloud
embedding is concatenated with normalized proprioception and passed to a
Gaussian actor MLP with hidden widths $(512,256,128)$. The critic remains the
standard privileged 190-dimensional MLP critic.

\subsection{Distilled Point-Cloud Student}
The distilled student follows the teacher-student~\cite{chen2022visualdexterity} setup while
using a PointNet++ student instead of the original voxel/sparse-convolution
student. We had to make this adaptation because the ogirinal code uses Minkowski Engine which is not compatible with pytorch 2.7 that IsaacSim 5.1 needs. 
The student receives only 133-dimensional proprioception and a fused
current-goal XYZRGB cloud of shape $\mathbb{R}^{2048\times6}$. The first
1024 rows form a rendered-current block with up to 1024 visible points; a
validity mask identifies which of those rows are render-valid, and padding
fills the remaining rows when fewer points are visible. The final 1024 rows
hold textured goal points. 

The student treats the current and goal points as one fused cloud without
explicit tags, separate towers, pose-oracle features, or relative-pose inputs.
Its PointNet++ encoder uses three set-abstraction stages with radius grouping,
sampling ratios $(0.25,0.25,1.0)$, radii $(0.025,0.05,0.1)$, and at most 64
neighbors per stage. The stage MLPs are $(6,64,128)$, $(131,192,256)$, and
$(259,256,256)$, followed by a final per-point MLP
$259\rightarrow256\rightarrow256$.
Max pooling produces a 256-dimensional descriptor. This descriptor is fused
with normalized proprioception by a Gaussian actor MLP with hidden widths
$(512,256,128)$. The student is trained with a DAgger-style RSL-RL
distillation runner using negative log likelihood of the teacher deterministic
action under the student's Gaussian distribution, with five learning epochs,
16 rollout steps per environment, learning rate $5\times10^{-4}$, and gradient
clipping at 1.0.

\subsection{Point-MAE Reference}
The Point-MAE reference keeps the same MLP actor-critic PPO stack as our frozen
embedding policy but replaces the metric-trained encoder with a frozen
Point-MAE encoder. It uses XYZ-only point clouds, groups 4096 points into 128
groups of size 32, and outputs 384-dimensional current and goal embeddings.
The actor receives proprioception, both embeddings, both centroids, and the
embedding-space L2 distance. This baseline tests whether generic
self-supervised point-cloud features provide a sufficient current-goal
rotation signal for this task.

\Needspace{0.78\textheight}
\section{Policy Learning Curves}
\label{app:learning_curves}

\begin{figure*}[!htbp]
  \centering
  \small
  \setlength{\tabcolsep}{2pt}
  \renewcommand{\arraystretch}{0.45}
  \begin{tabular}{@{}c@{\hspace{0.004\linewidth}}c@{\hspace{0.004\linewidth}}c@{\hspace{0.004\linewidth}}c@{}}
    (a) & (b) & (c) & (d) \\
    \includegraphics[width=0.24\linewidth]{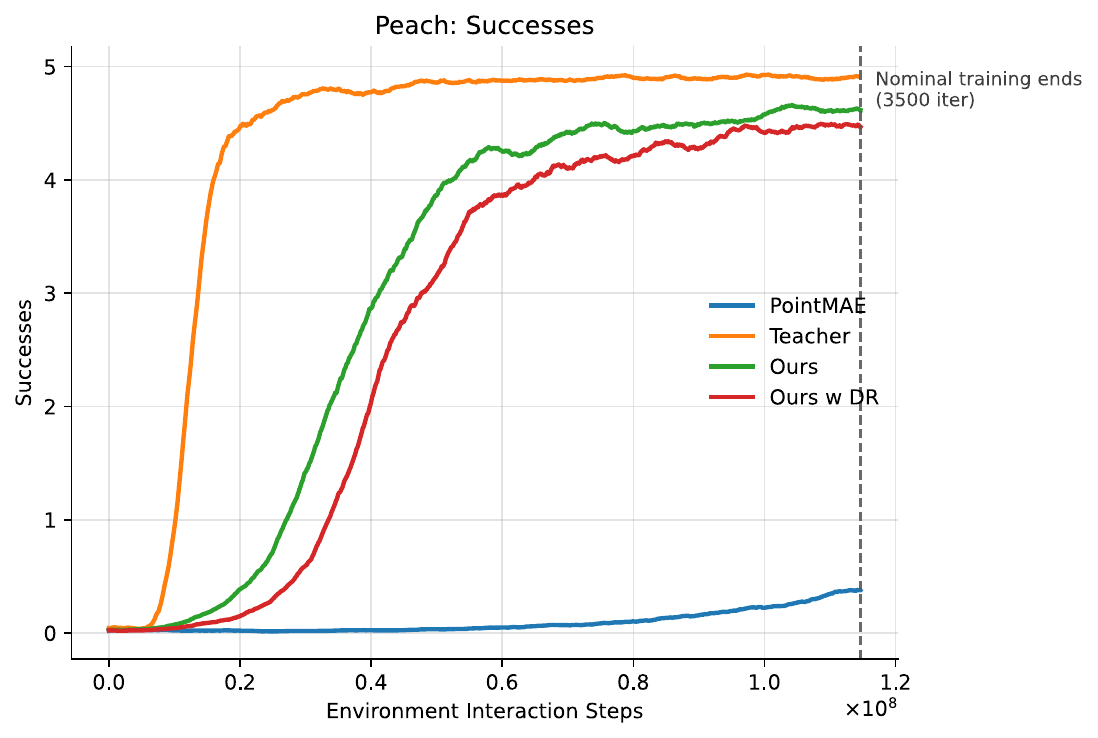} &
    \includegraphics[width=0.24\linewidth]{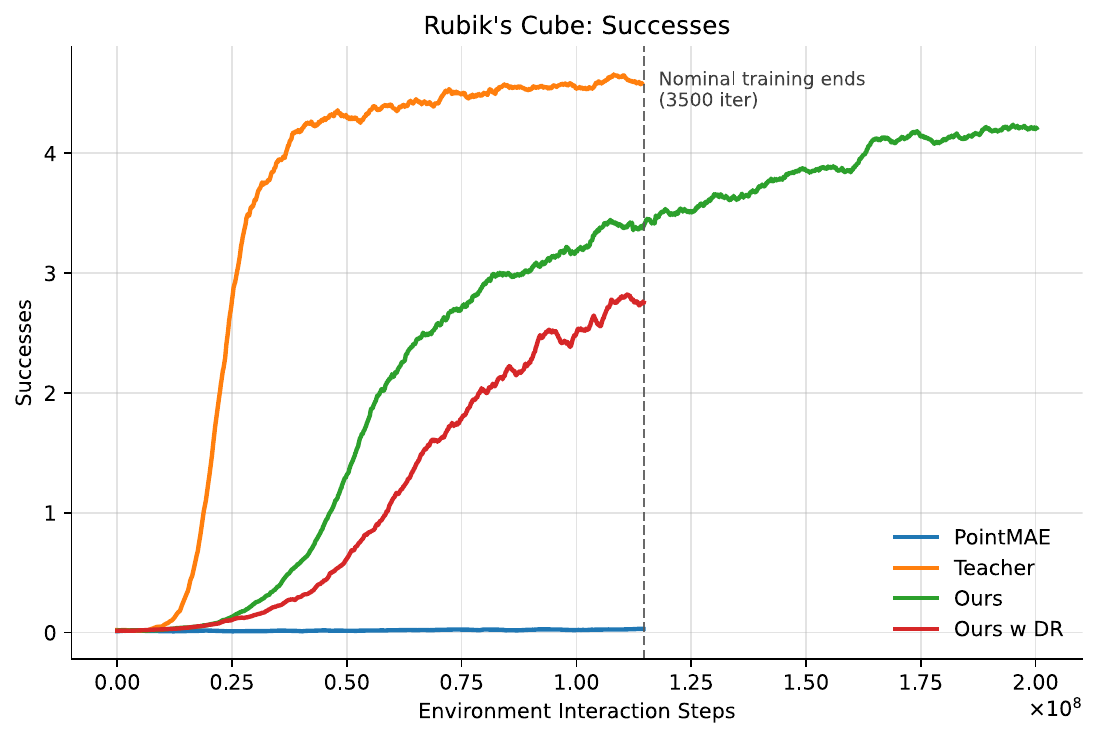} &
    \includegraphics[width=0.24\linewidth]{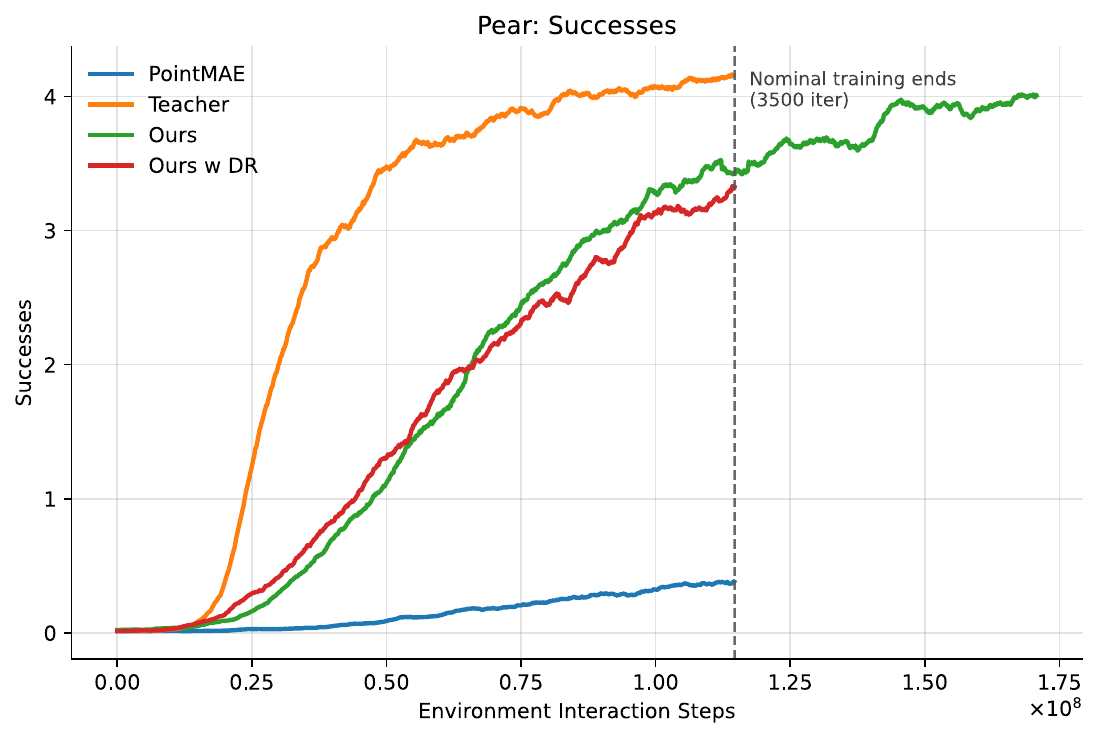} &
    \includegraphics[width=0.24\linewidth]{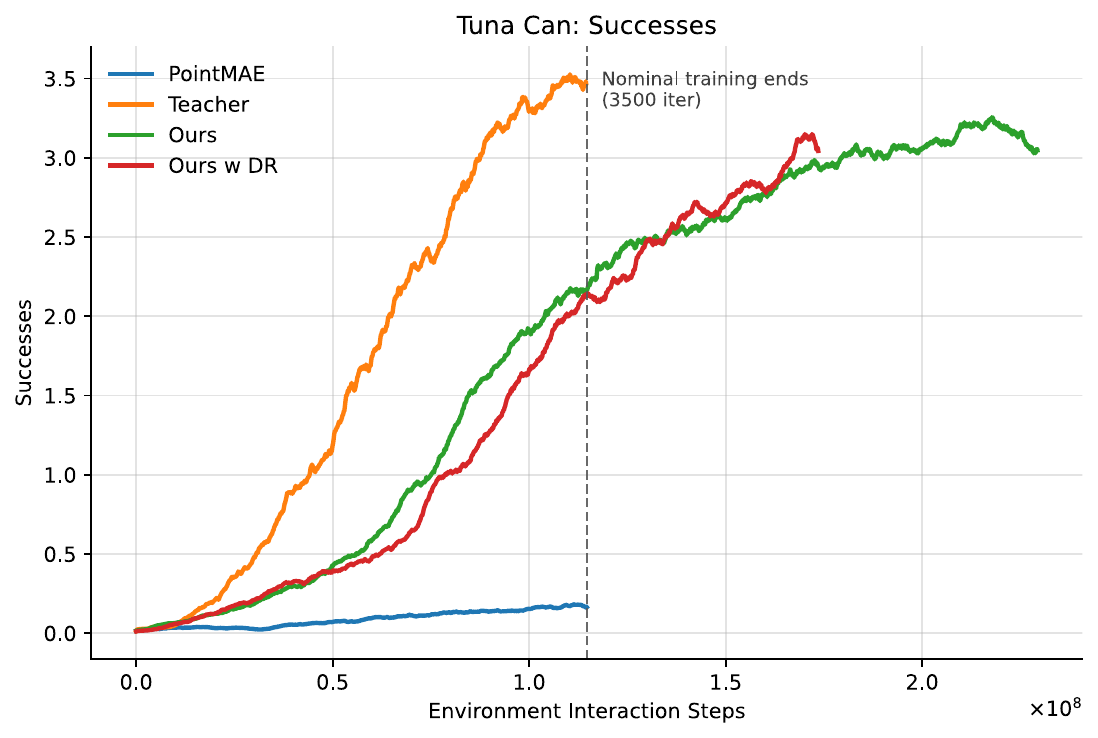} \\
    \includegraphics[width=0.24\linewidth]{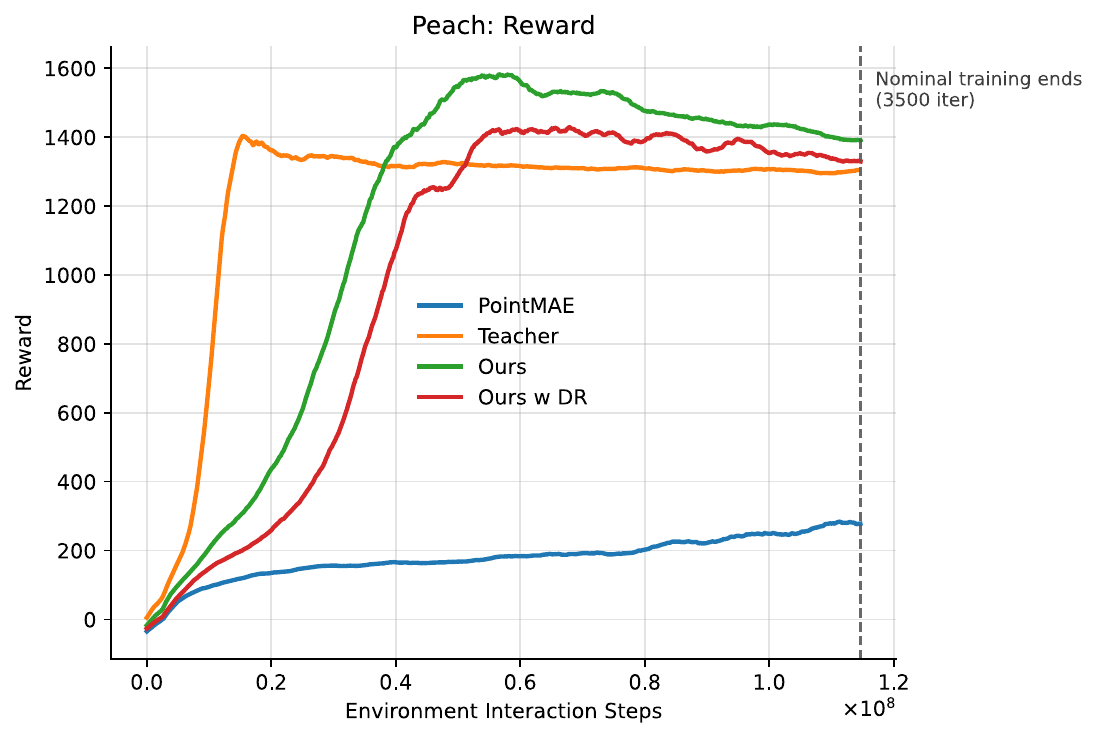} &
    \includegraphics[width=0.24\linewidth]{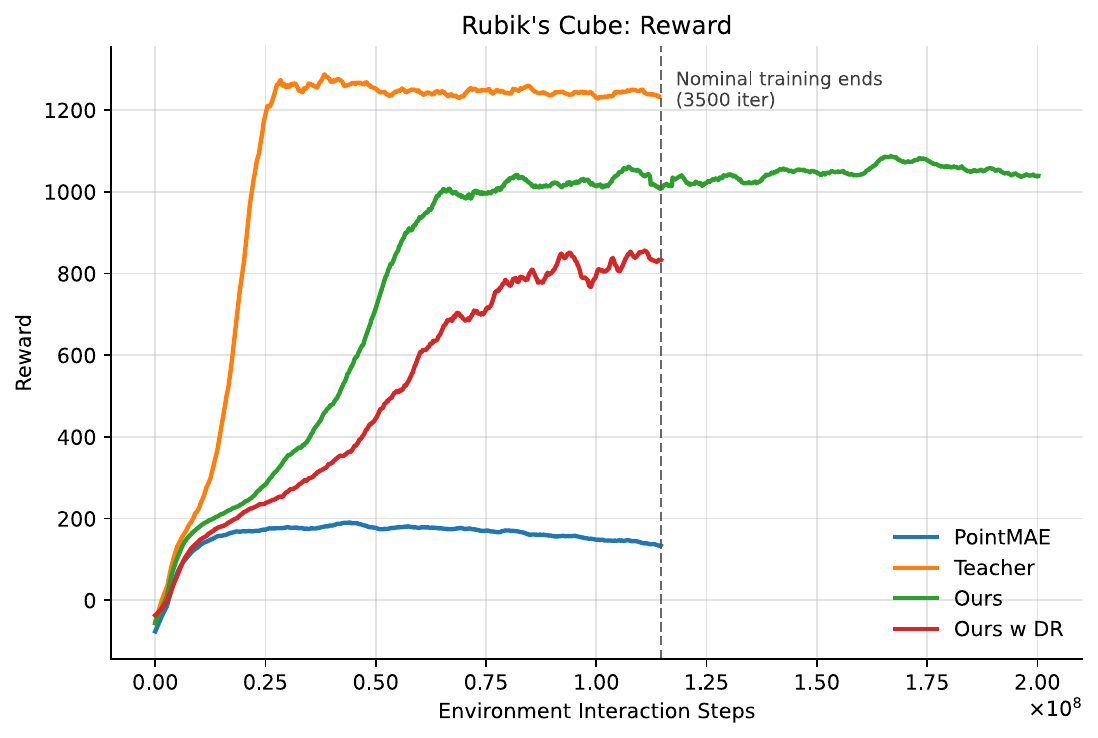} &
    \includegraphics[width=0.24\linewidth]{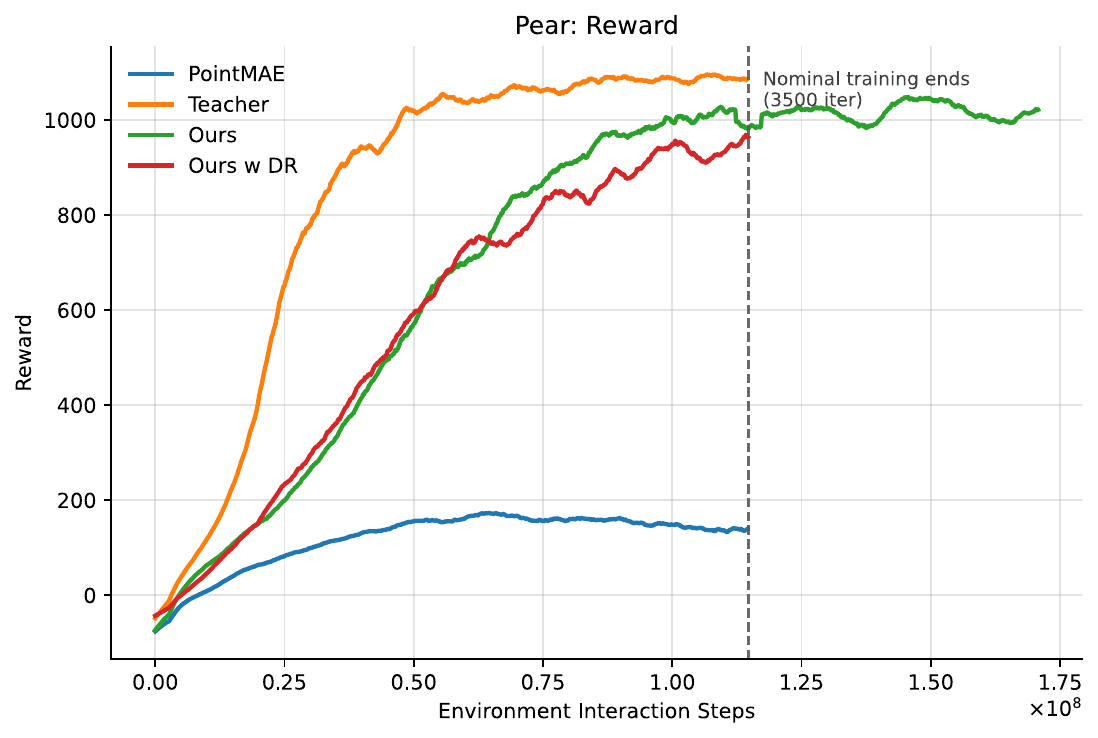} &
    \includegraphics[width=0.24\linewidth]{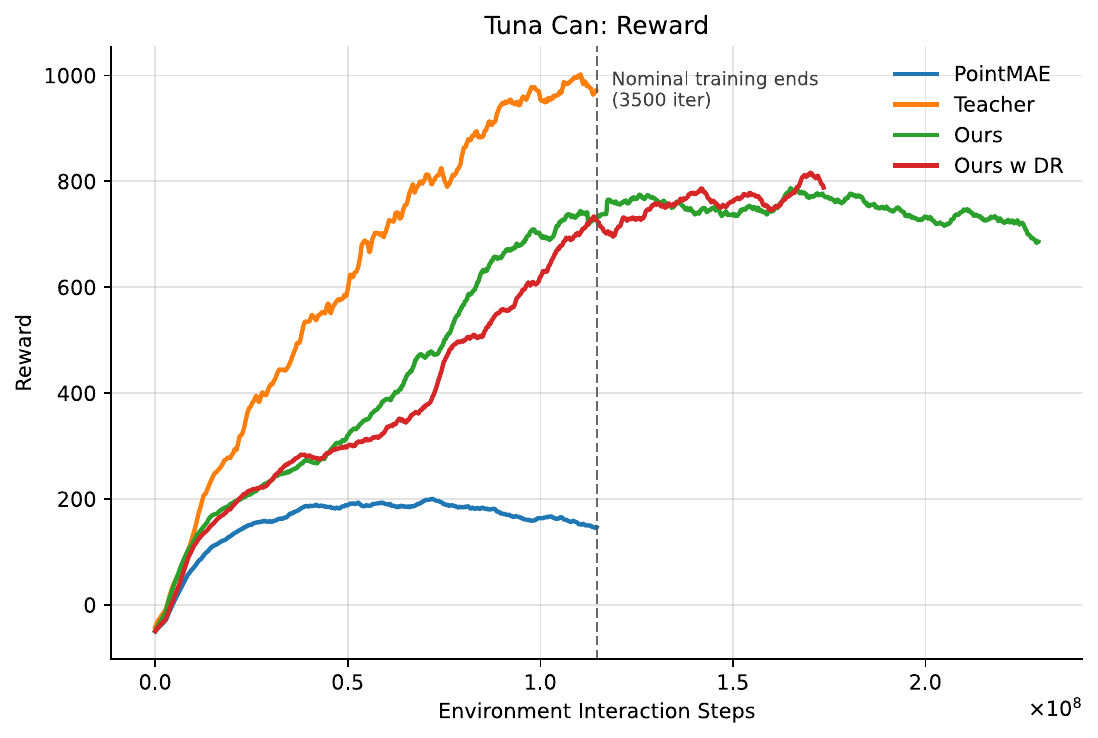} \\
    \includegraphics[width=0.24\linewidth]{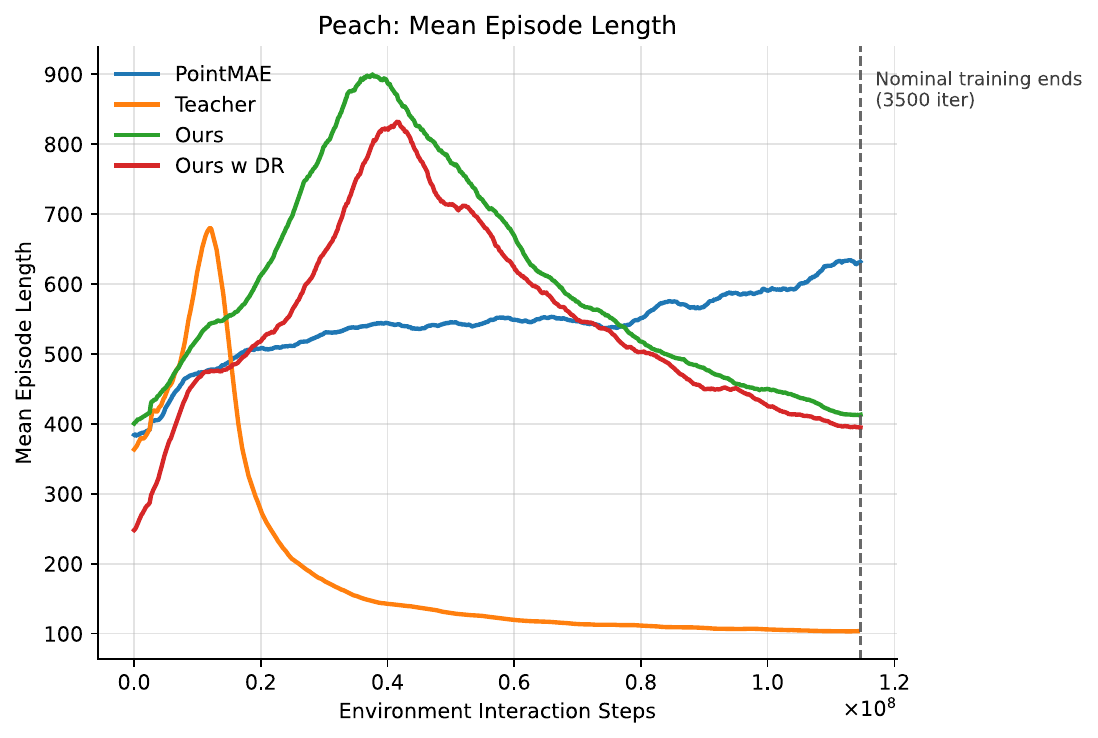} &
    \includegraphics[width=0.24\linewidth]{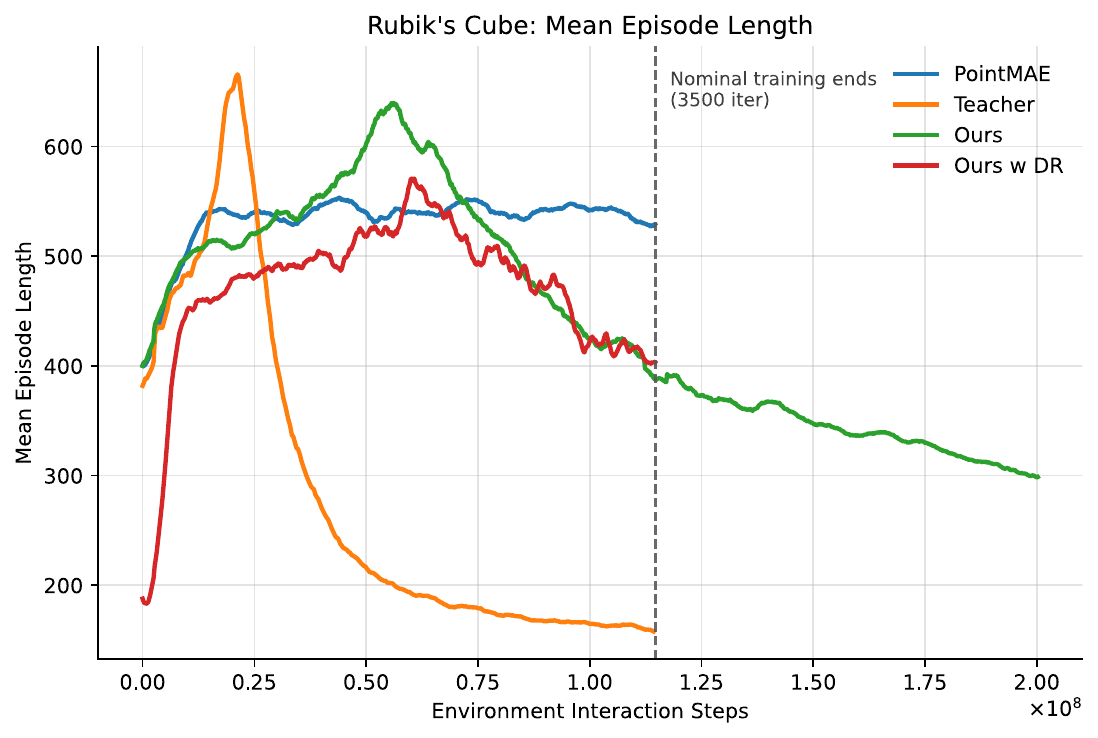} &
    \includegraphics[width=0.24\linewidth]{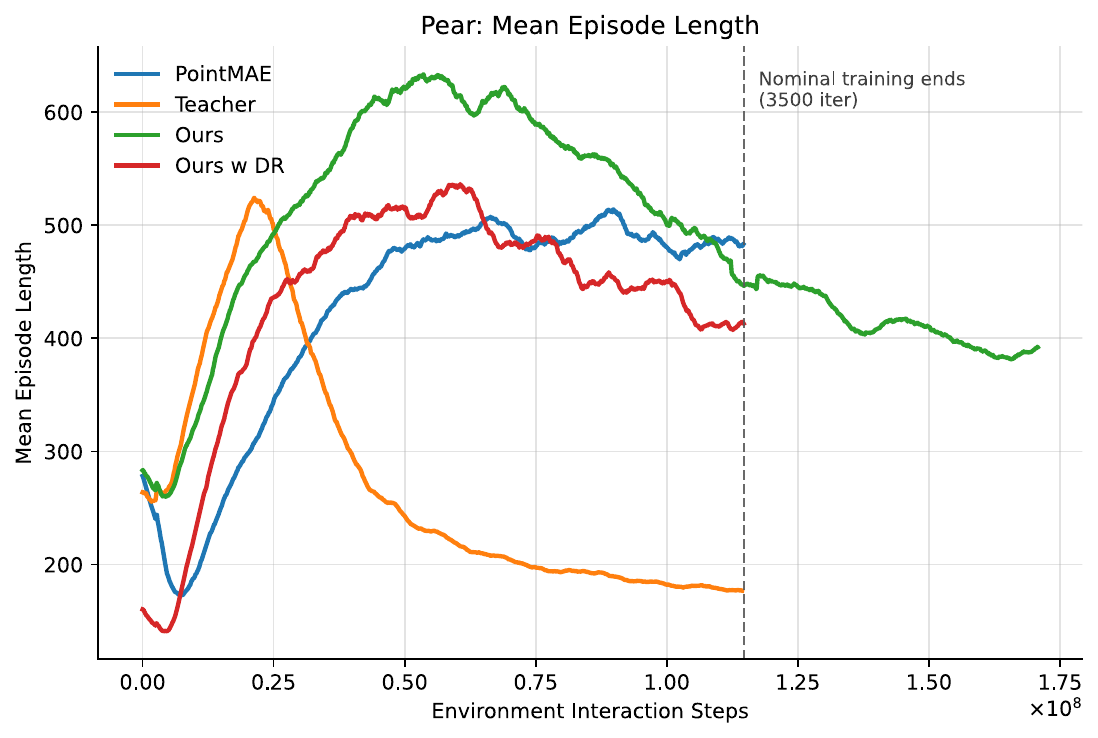} &
    \includegraphics[width=0.24\linewidth]{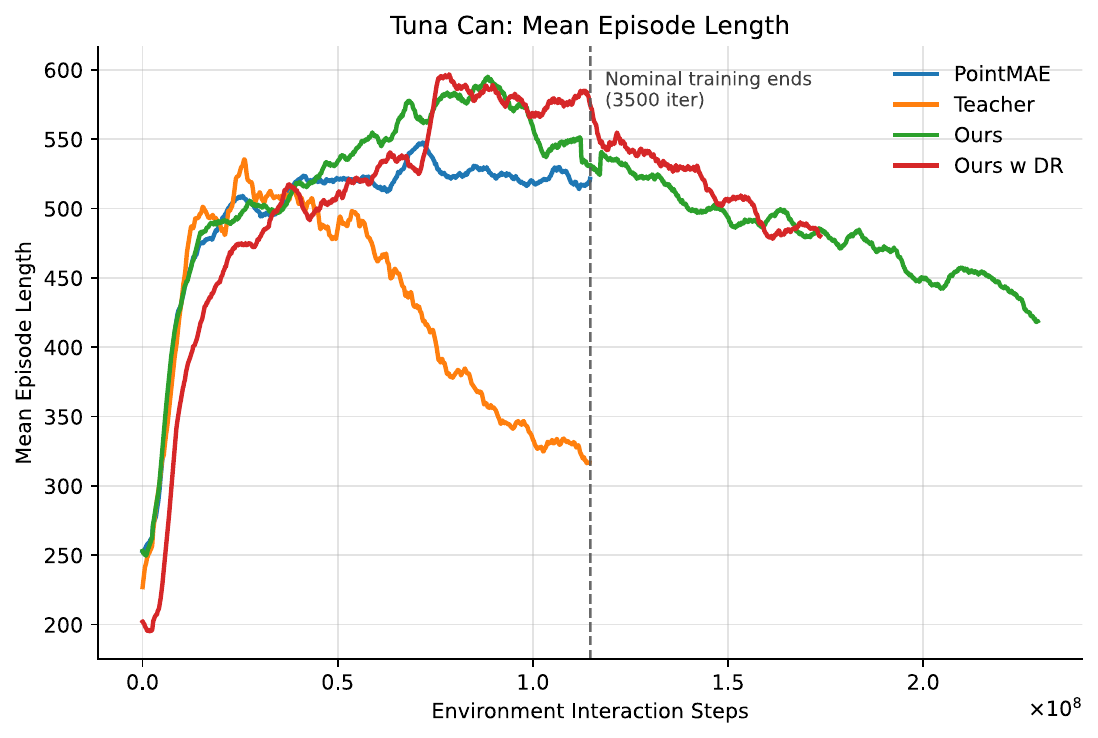}
  \end{tabular}
  \caption{Smoothed policy learning curves for all four evaluated YCB objects. Columns (a)--(d) correspond to peach, Rubik's cube, pear, and tuna can, respectively. Rows from top to bottom show consecutive successes, episode reward, and mean episode length. Orange is the teacher, green is ours, red is ours with domain randomization and blue is pointMAE. Ours (green) is trained longer to show that with more iterations we can converge to the optimal performance}
  \label{fig:appendix_learning_curves}
\end{figure*}
\FloatBarrier

\section{Encoder evaluation on all of YCB}
\label{app:encoder_evaluation_ycb}

\subsection{Qualitative Angle t-SNE}
\label{app:angle_tsne}

\begin{figure*}[!htbp]
  \centering
  \subfloat[Peach\label{fig:appendix_angle_tsne_peach}]{
    \includegraphics[width=0.32\linewidth]{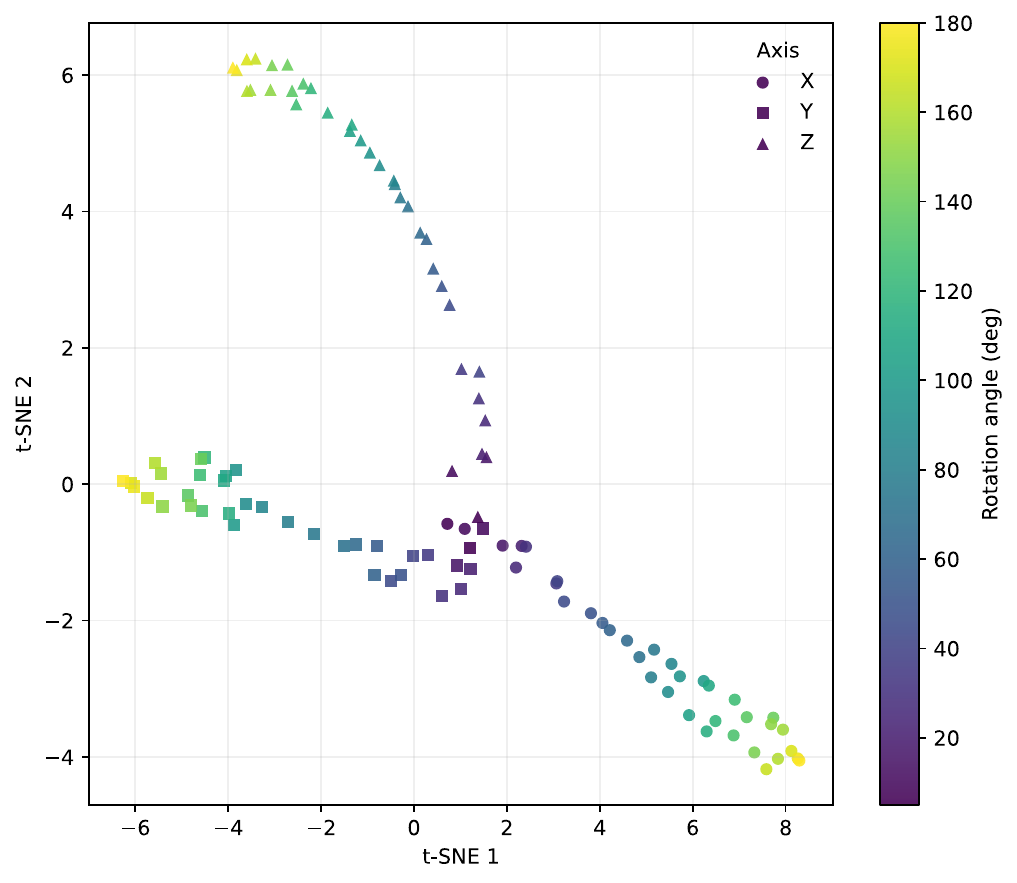}
  }
  \subfloat[Rubik's Cube\label{fig:appendix_angle_tsne_rubiks_cube}]{
    \includegraphics[width=0.32\linewidth]{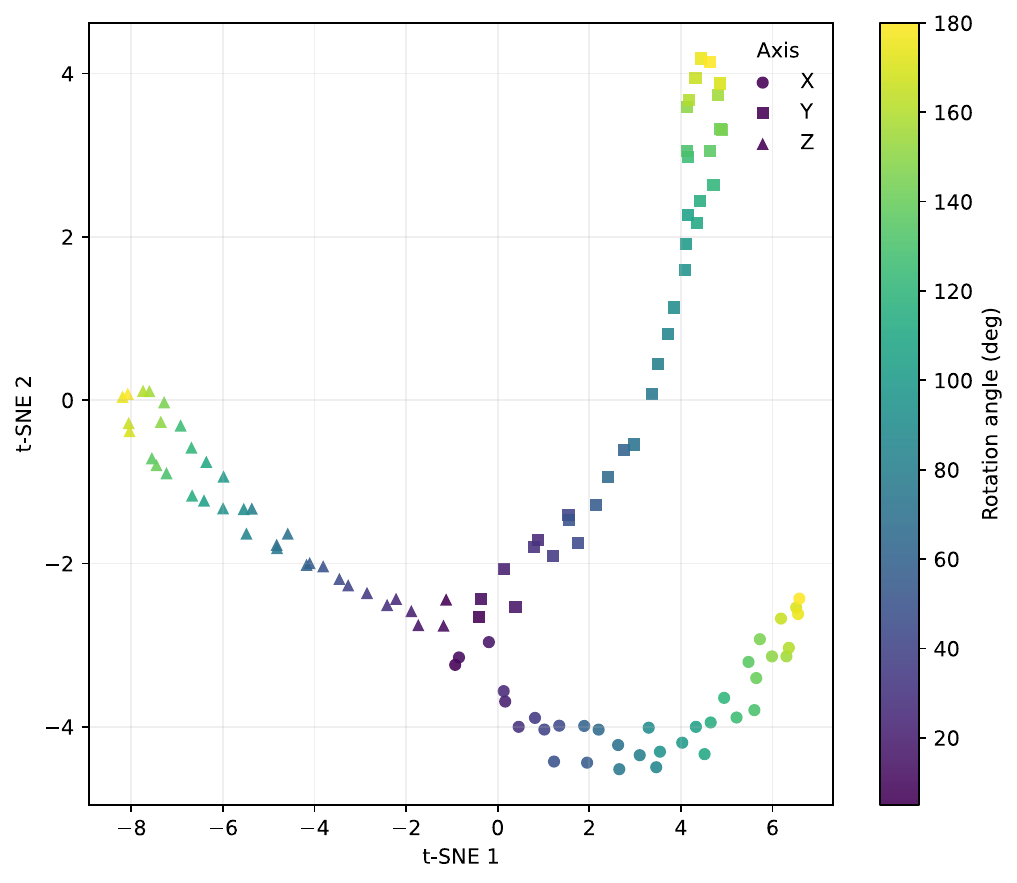}
  }
  \subfloat[Tuna Can\label{fig:appendix_angle_tsne_tuna_can}]{
    \includegraphics[width=0.32\linewidth]{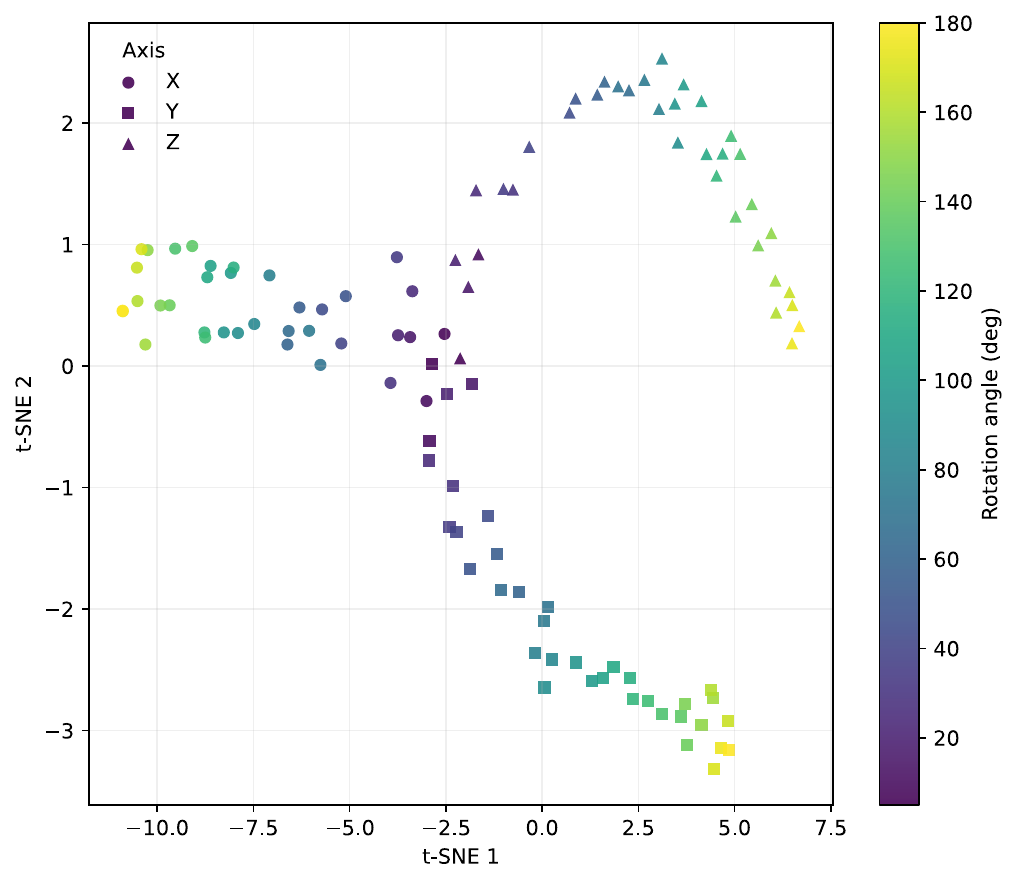}
  }
  \caption{Qualitative t-SNE visualizations of rotation samples from our encoder, colored by rotation angle.}
  \label{fig:appendix_angle_tsne}
\end{figure*}

\clearpage

\subsection{Rotation axis Diagnostics}
\label{app:axis_delta_cosine}

\begin{figure*}[!htbp]
  \centering
  \subfloat[Ours\label{fig:appendix_axis_delta_ours_peach}]{
    \includegraphics[width=0.82\linewidth]{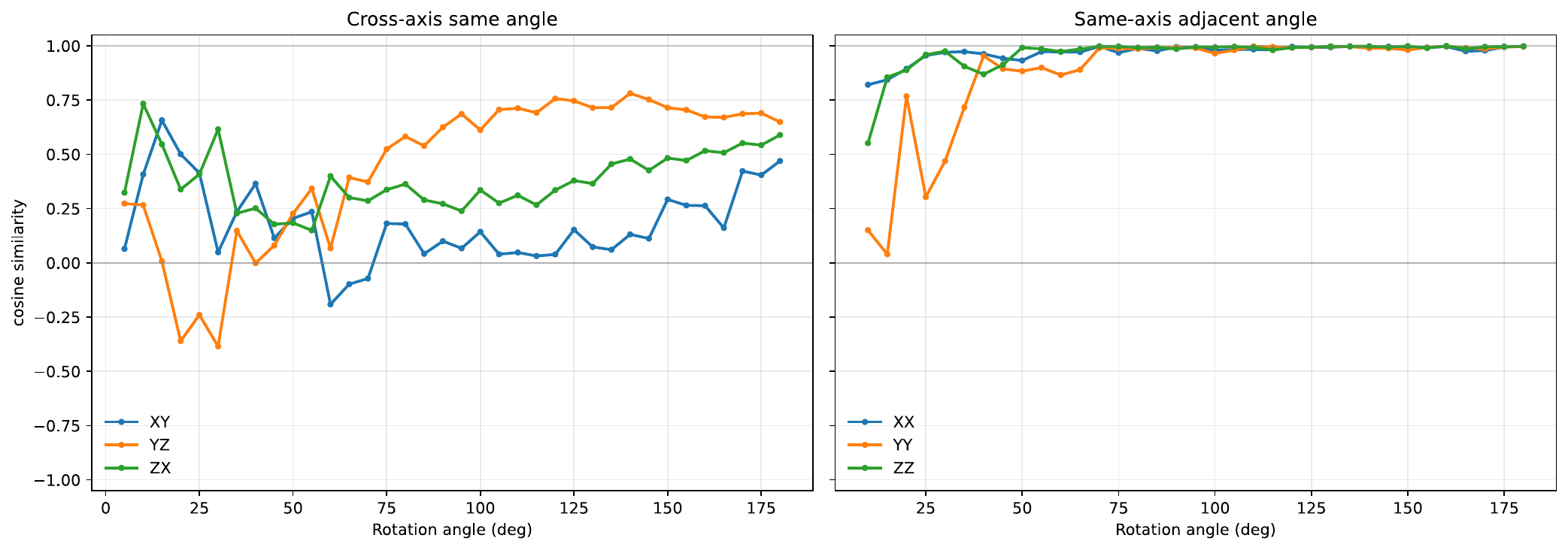}
  } \\
  \subfloat[PointMAE\label{fig:appendix_axis_delta_pointmae_peach}]{
    \includegraphics[width=0.82\linewidth]{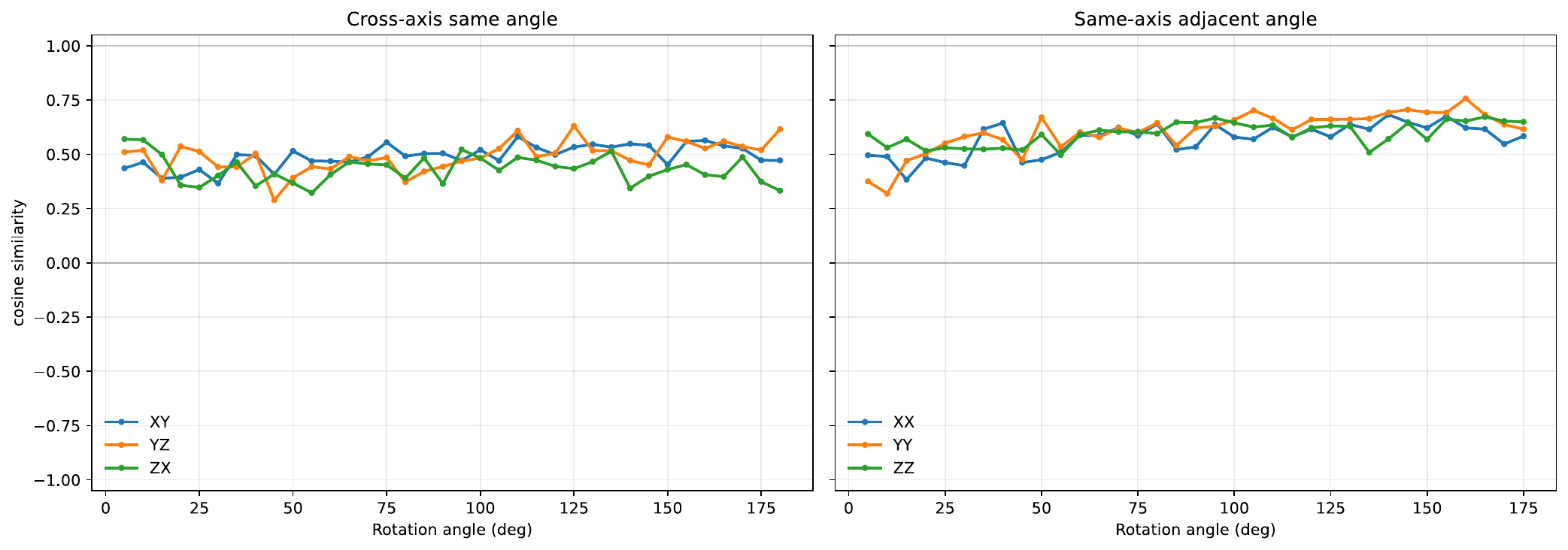}
  } \\
  \subfloat[Distilled Student\label{fig:appendix_axis_delta_student_peach}]{
    \includegraphics[width=0.82\linewidth]{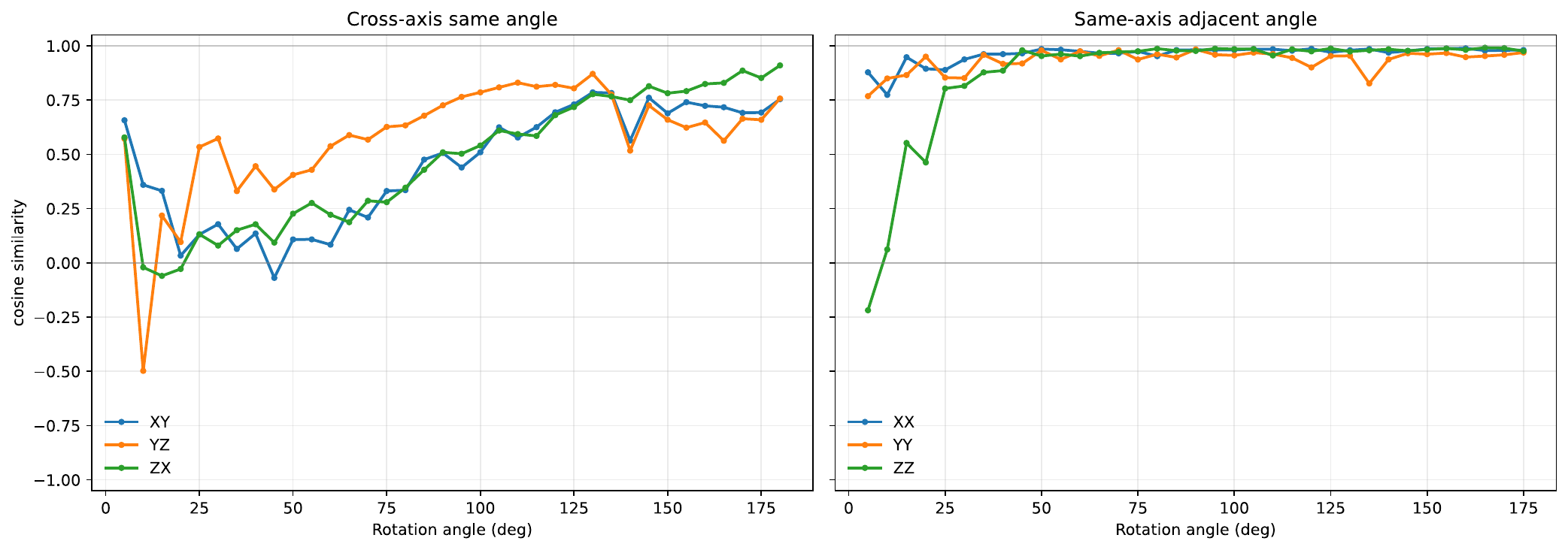}
  }
  \caption{Cosine similarity of embedding displacement directions under controlled single-axis rotations for peach point clouds. Cross-axis same-angle curves test whether rotations with equal magnitude but different axes are separated, while same-axis adjacent-angle curves provide a smoothness control.}
  \label{fig:appendix_axis_delta_peach}
\end{figure*}

\begin{figure*}[!htbp]
  \centering
  \subfloat[Ours\label{fig:appendix_axis_delta_ours_rubiks_cube}]{
    \includegraphics[width=0.82\linewidth]{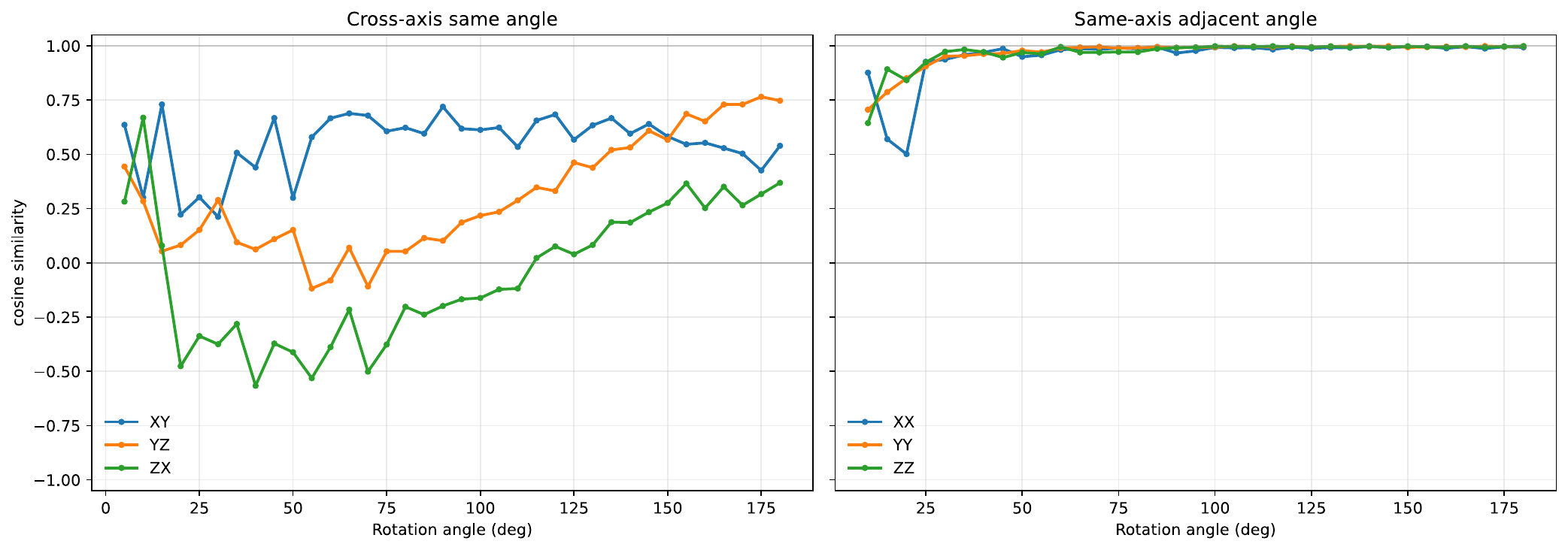}
  } \\
  \subfloat[PointMAE\label{fig:appendix_axis_delta_pointmae_rubiks_cube}]{
    \includegraphics[width=0.82\linewidth]{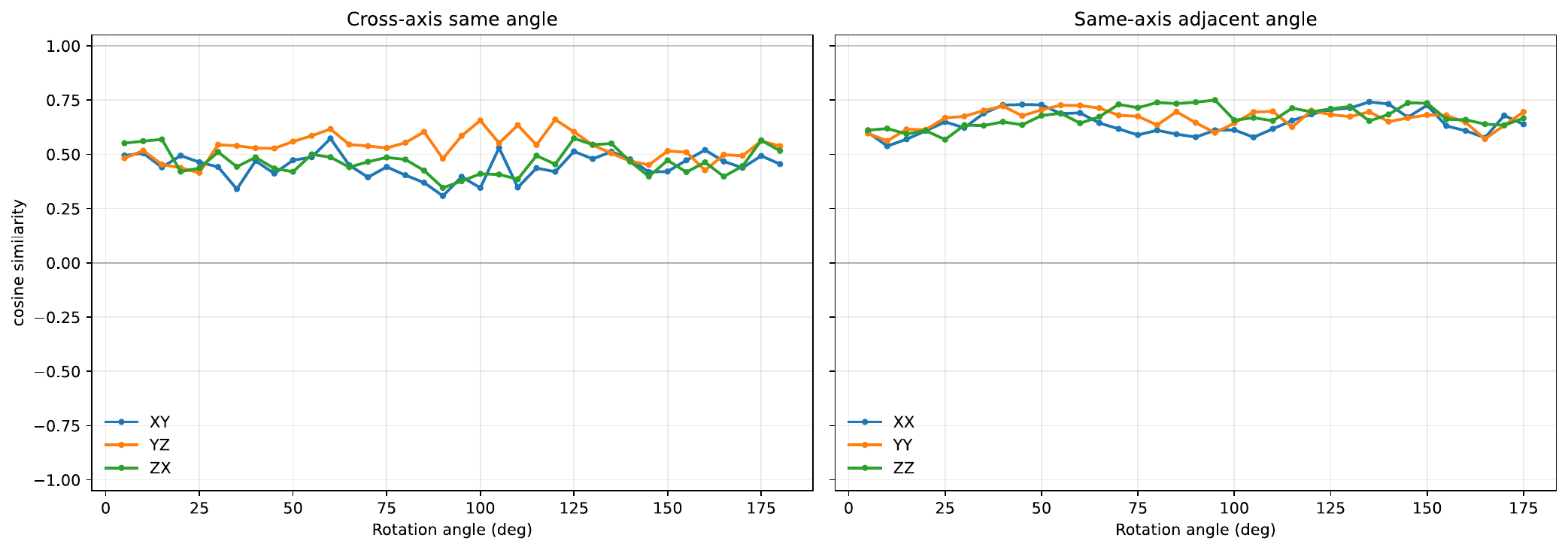}
  } \\
  \subfloat[Distilled Student\label{fig:appendix_axis_delta_student_rubiks_cube}]{
    \includegraphics[width=0.82\linewidth]{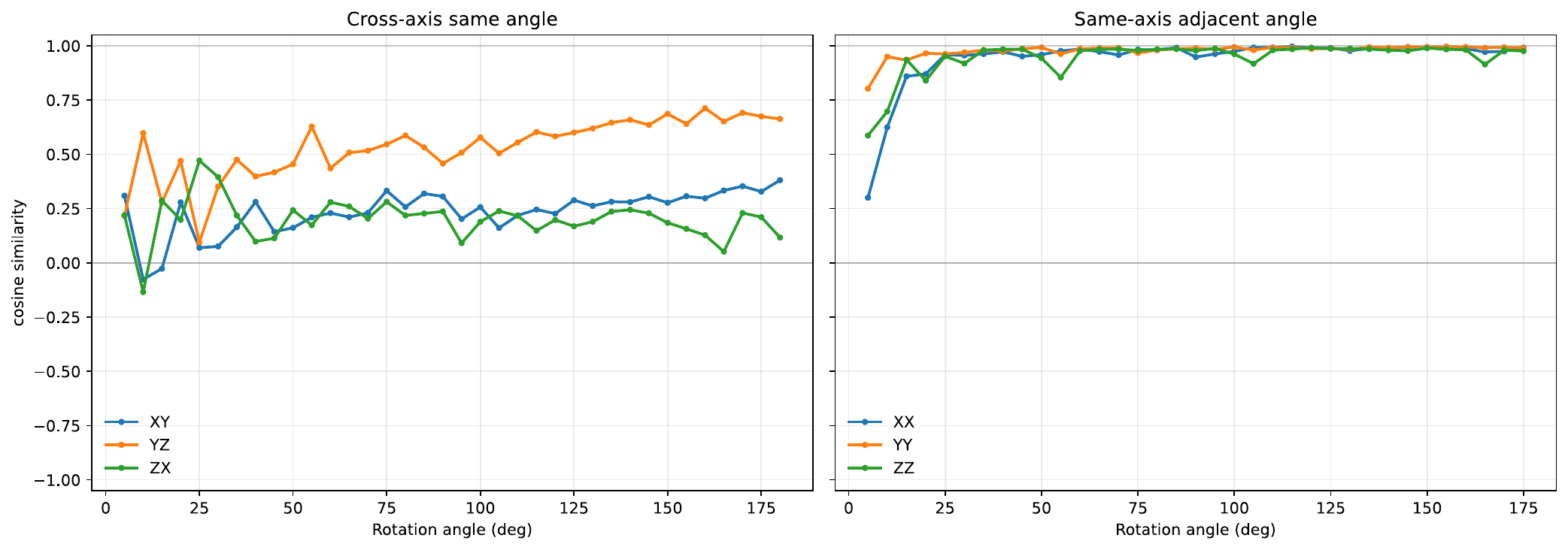}
  }
  \caption{Cosine similarity of embedding displacement directions under controlled single-axis rotations for Rubik's cube point clouds. Cross-axis same-angle curves test whether rotations with equal magnitude but different axes are separated, while same-axis adjacent-angle curves provide a smoothness control.}
  \label{fig:appendix_axis_delta_rubiks_cube}
\end{figure*}

\begin{figure*}[!htbp]
  \centering
  \subfloat[Ours\label{fig:appendix_axis_delta_ours_tuna_can}]{
    \includegraphics[width=0.82\linewidth]{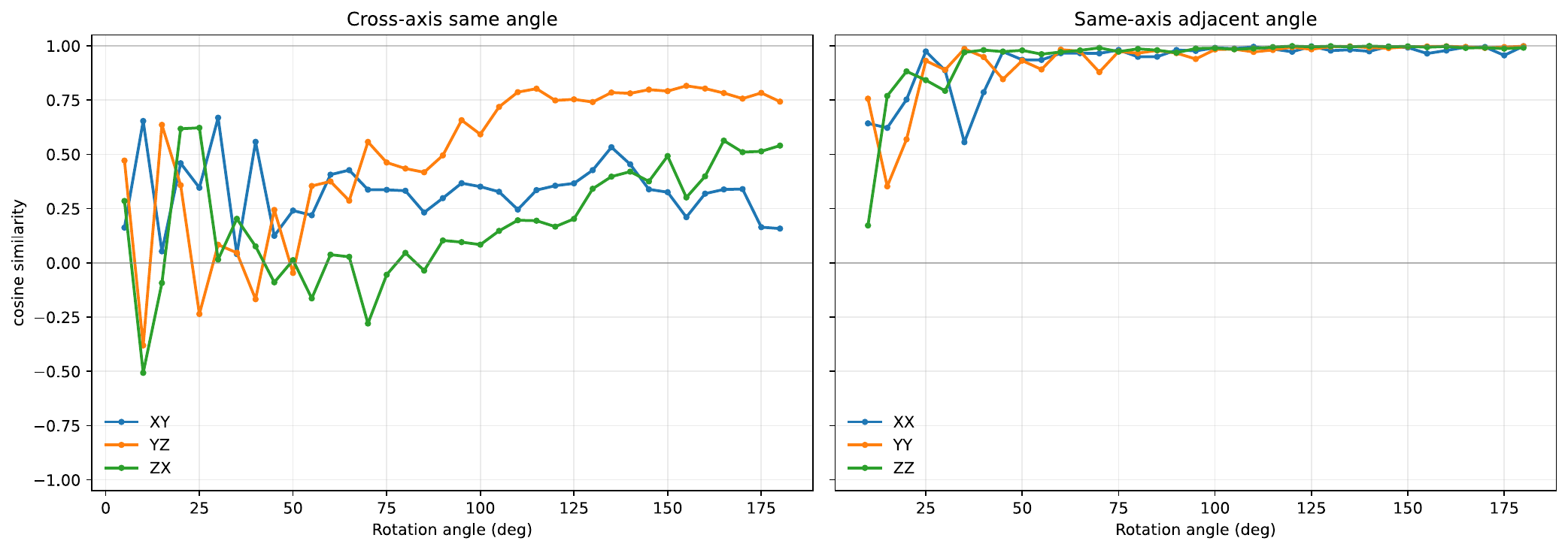}
  } \\
  \subfloat[PointMAE\label{fig:appendix_axis_delta_pointmae_tuna_can}]{
    \includegraphics[width=0.82\linewidth]{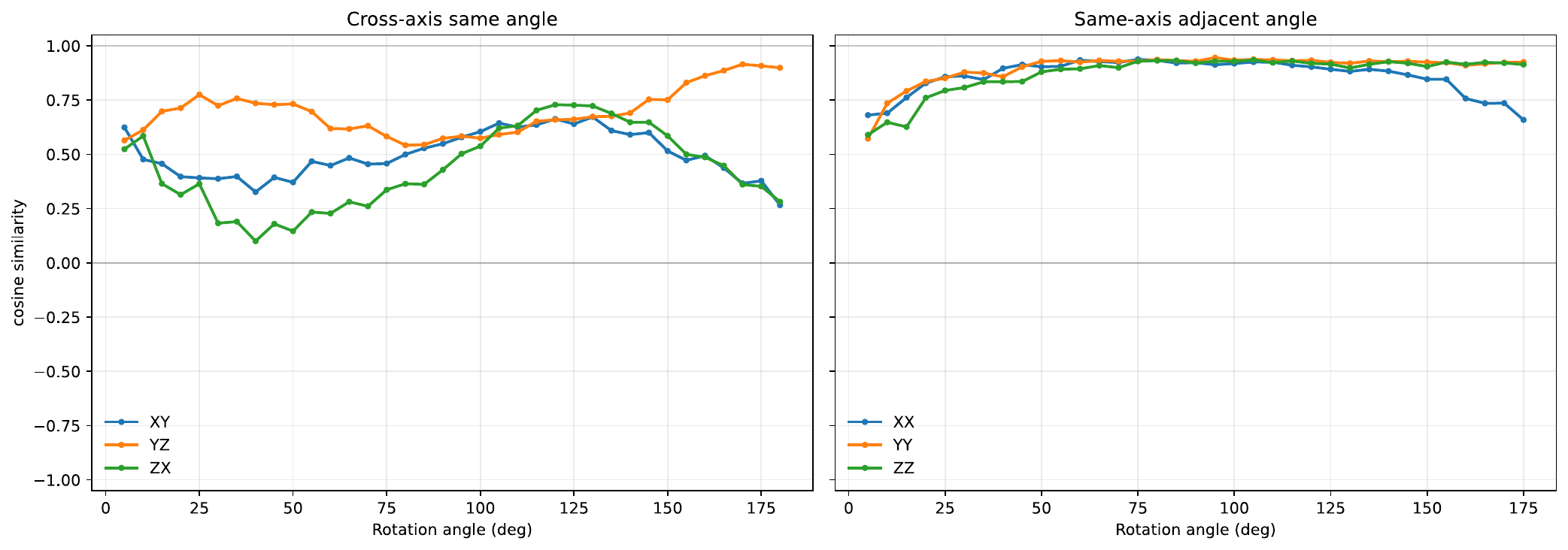}
  } \\
  \subfloat[Distilled Student\label{fig:appendix_axis_delta_student_tuna_can}]{
    \includegraphics[width=0.82\linewidth]{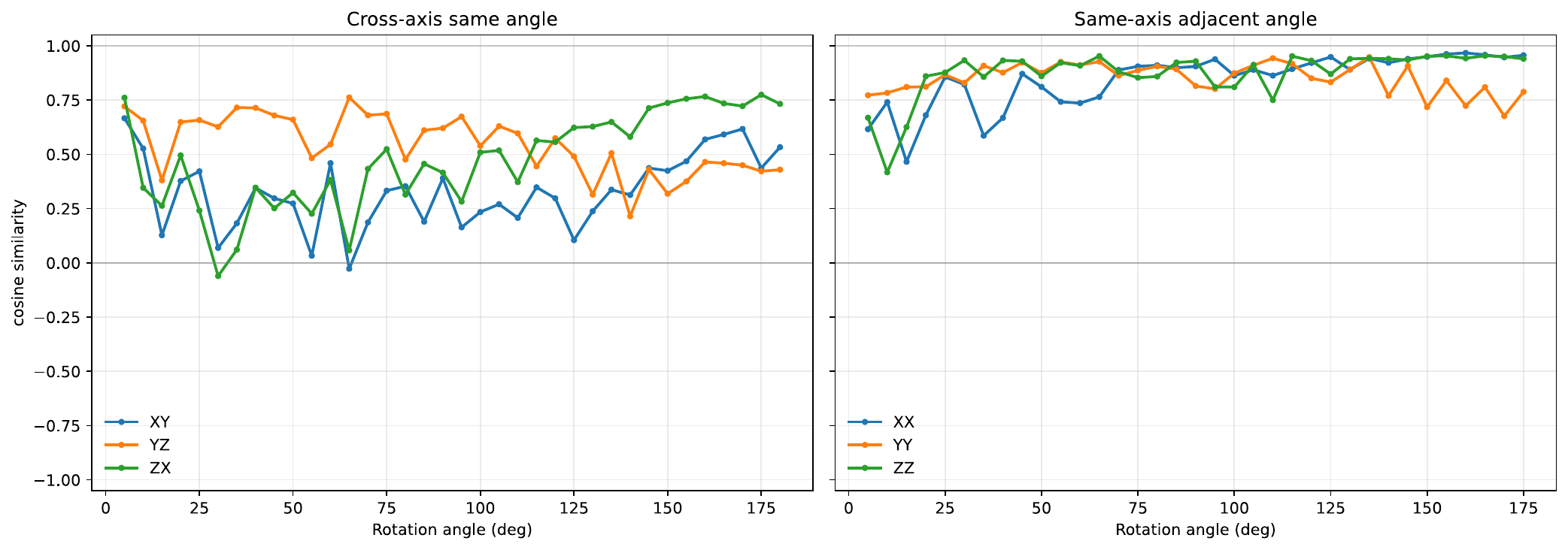}
  }
  \caption{Cosine similarity of embedding displacement directions under controlled single-axis rotations for tuna can point clouds. Cross-axis same-angle curves test whether rotations with equal magnitude but different axes are separated, while same-axis adjacent-angle curves provide a smoothness control.}
  \label{fig:appendix_axis_delta_tuna_can}
\end{figure*}

\FloatBarrier

\subsection{Symmetry Stress Cases}
\label{app:symmetry_cases}

\begin{figure*}[!htbp]
  \centering
  \subfloat[Medium clamp, angle\label{fig:appendix_symmetry_tsne_medium_clamp}]{
    \includegraphics[width=0.31\linewidth]{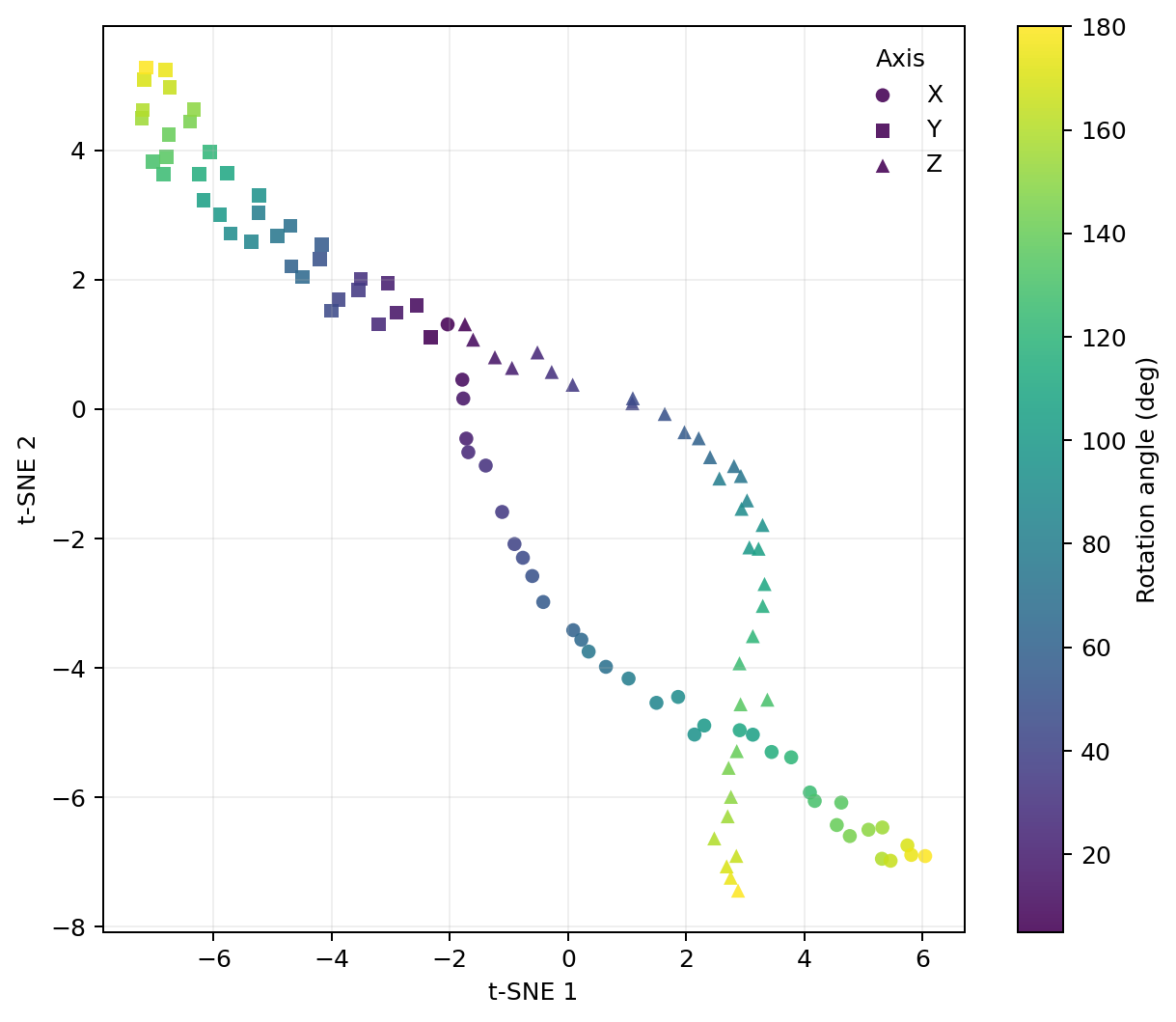}
  }
  \subfloat[Scissors, angle\label{fig:appendix_symmetry_tsne_scissors}]{
    \includegraphics[width=0.31\linewidth]{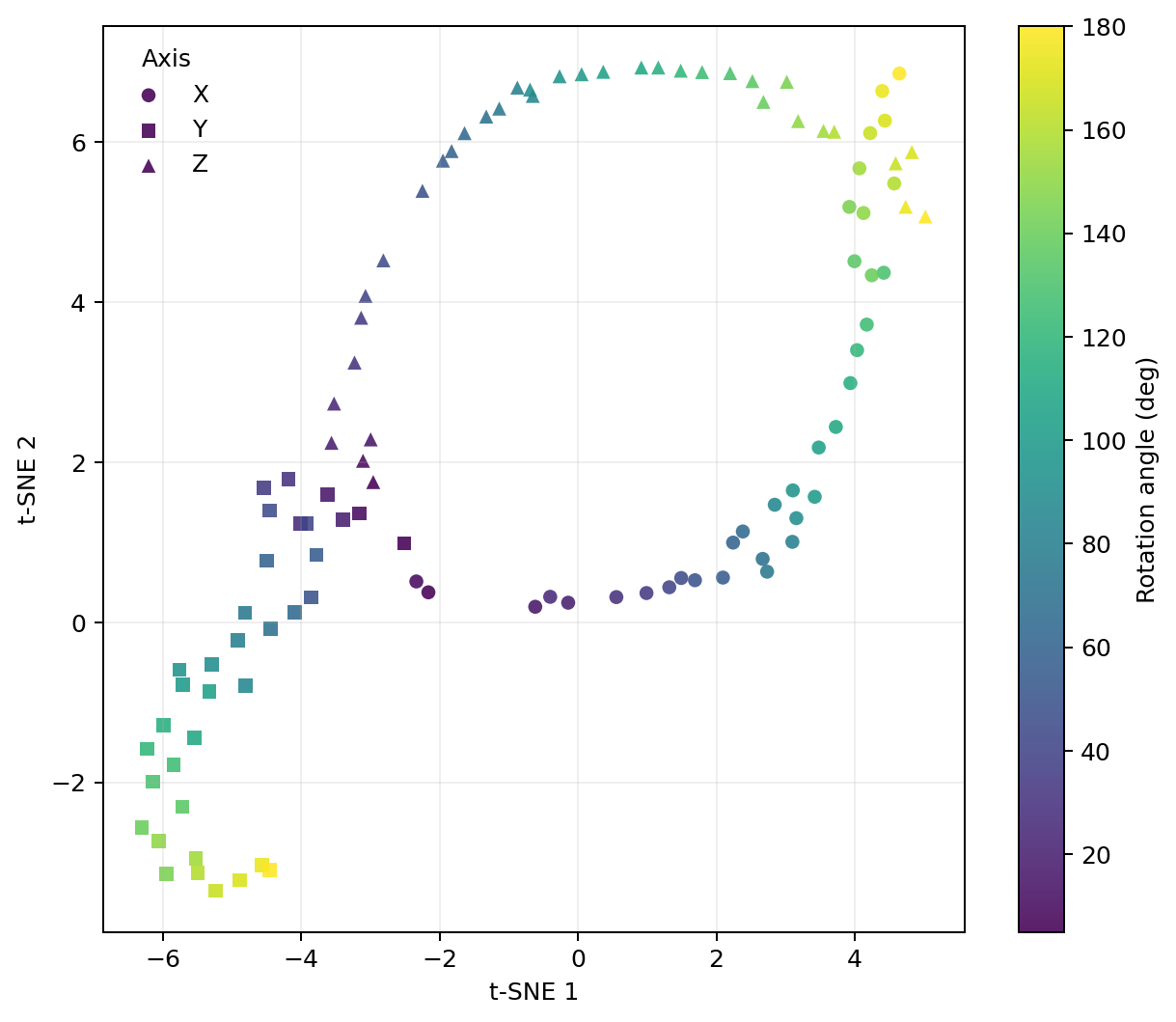}
  }
  \subfloat[Large marker, angle\label{fig:appendix_symmetry_tsne_large_marker}]{
    \includegraphics[width=0.31\linewidth]{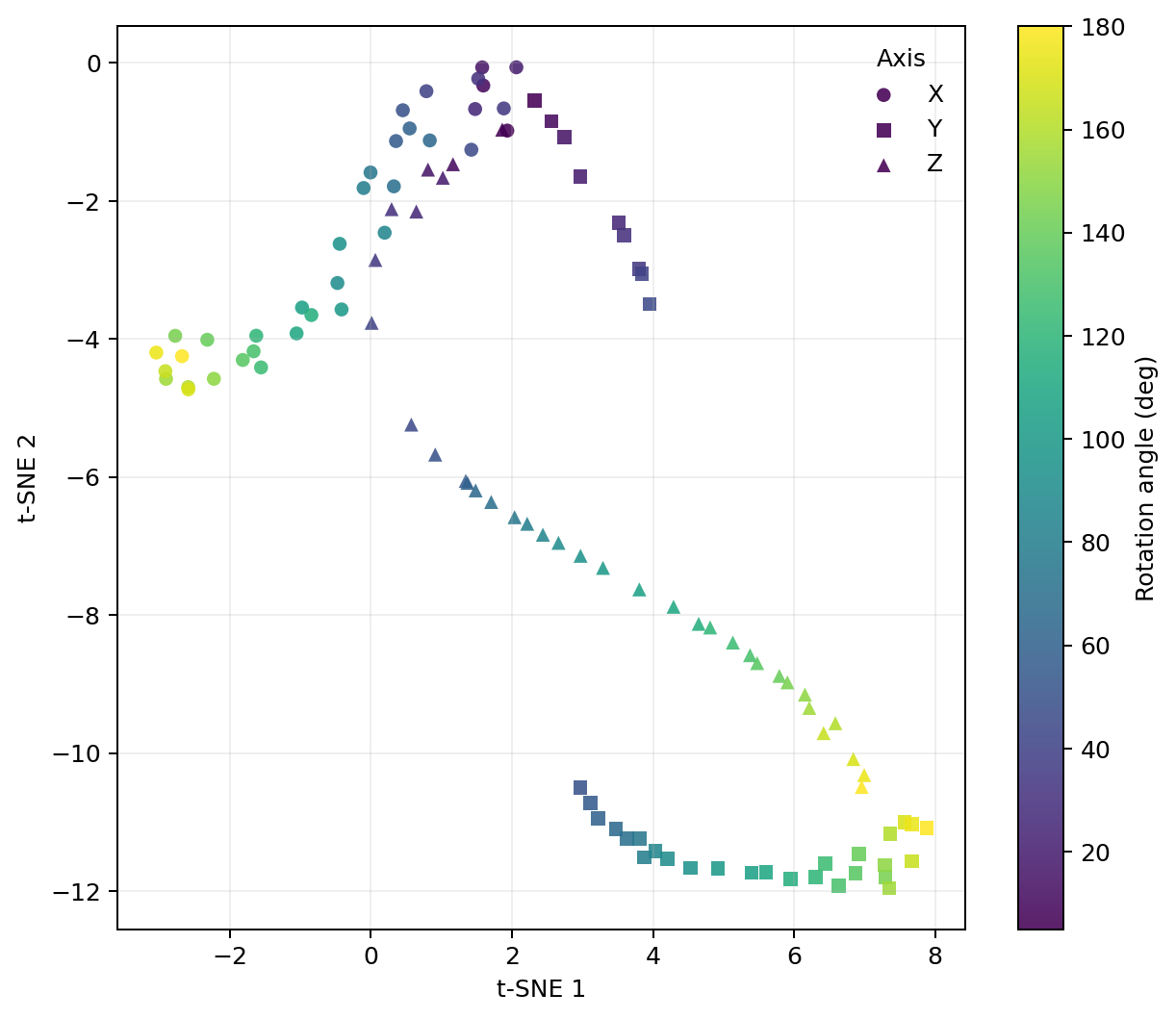}
  }
  \caption{Angle t-SNE visualizations for the medium clamp, scissors, and large marker. Marker style denotes rotation axis and color denotes rotation angle}
  \label{fig:appendix_symmetry_tsne}
\end{figure*}

\begin{figure*}[!htbp]
  \centering
  \subfloat[Medium clamp\label{fig:appendix_symmetry_axis_delta_medium_clamp}]{
    \includegraphics[width=0.82\linewidth]{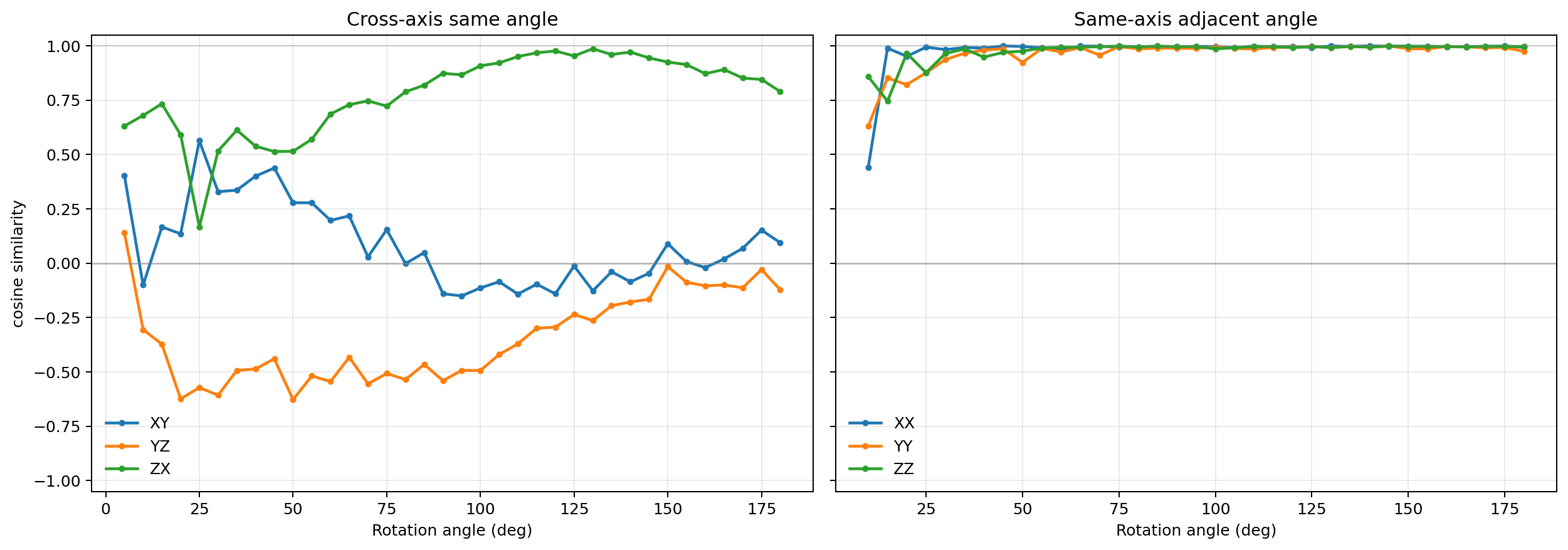}
  }\\
  \subfloat[Scissors\label{fig:appendix_symmetry_axis_delta_scissors}]{
    \includegraphics[width=0.82\linewidth]{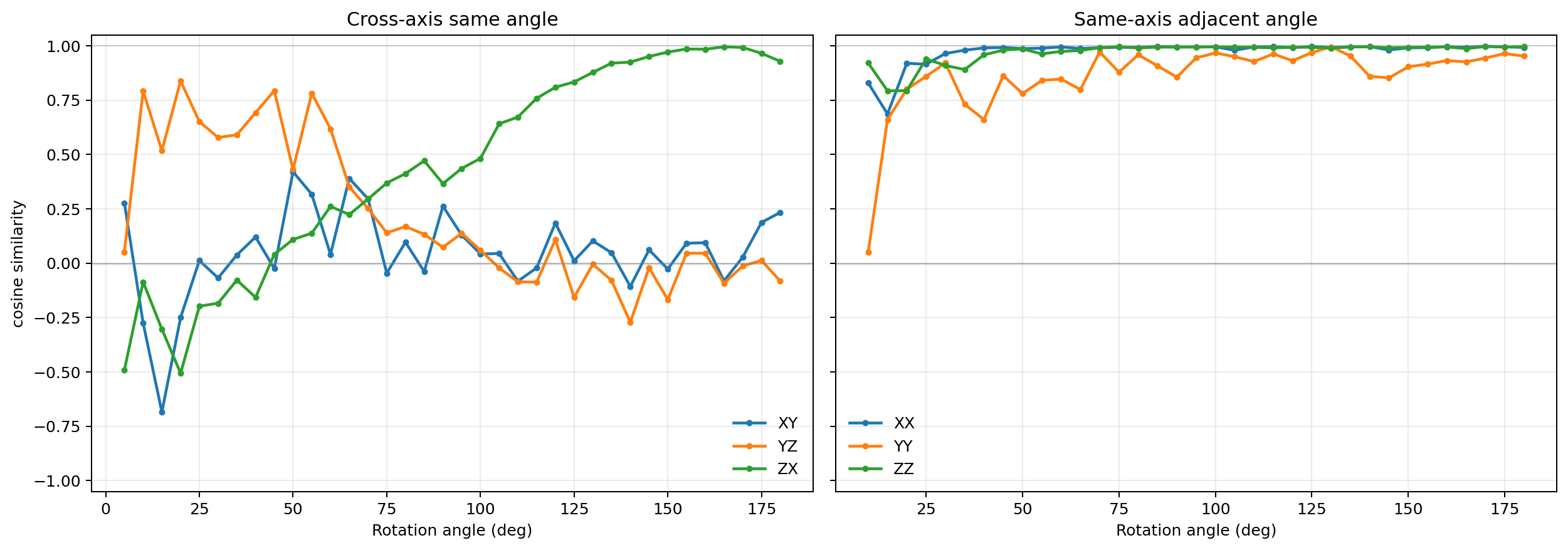}
  }\\
  \subfloat[Large marker\label{fig:appendix_symmetry_axis_delta_large_marker}]{
    \includegraphics[width=0.82\linewidth]{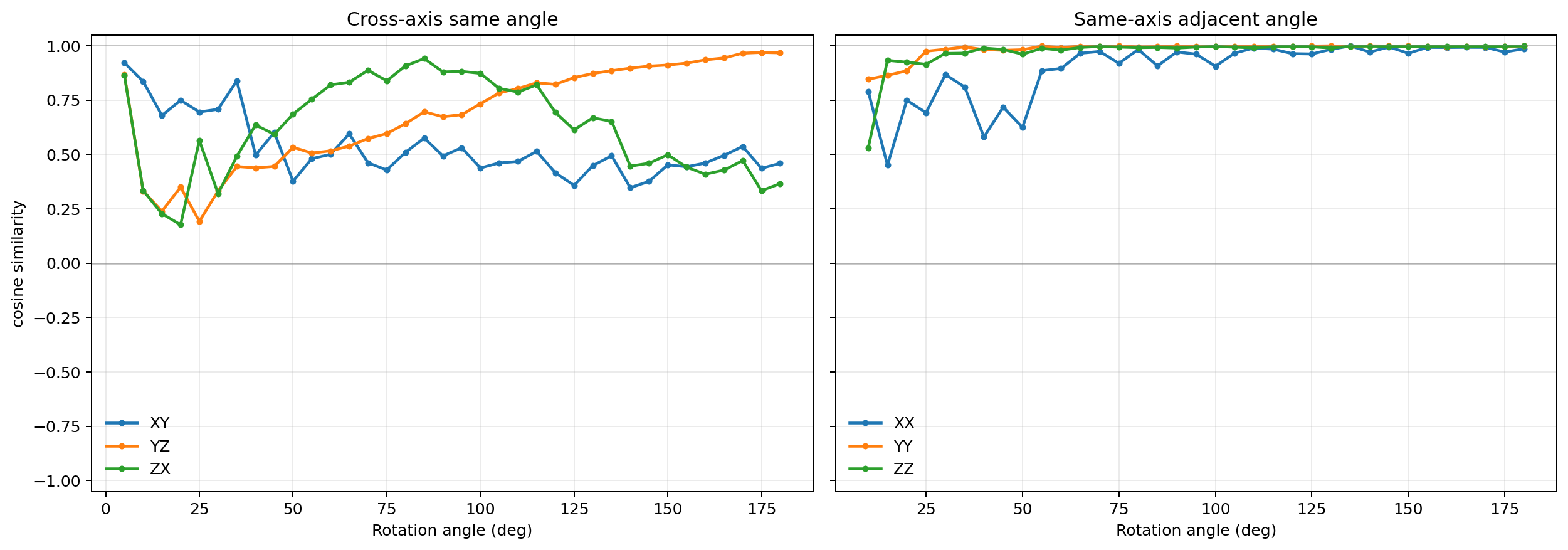}
  }
  \caption{Axis similarity diagnostics for the three symmetry stress cases. The high cross-axis curves identify rotations around different axes that induce nearly parallel embedding changes.}
  \label{fig:appendix_symmetry_axis_delta}
\end{figure*}

\begin{figure*}[!htbp]
  \centering
  \subfloat[Medium clamp: X110 vs. Z135\label{fig:appendix_symmetry_pointcloud_medium_clamp}]{
    \includegraphics[width=0.68\linewidth]{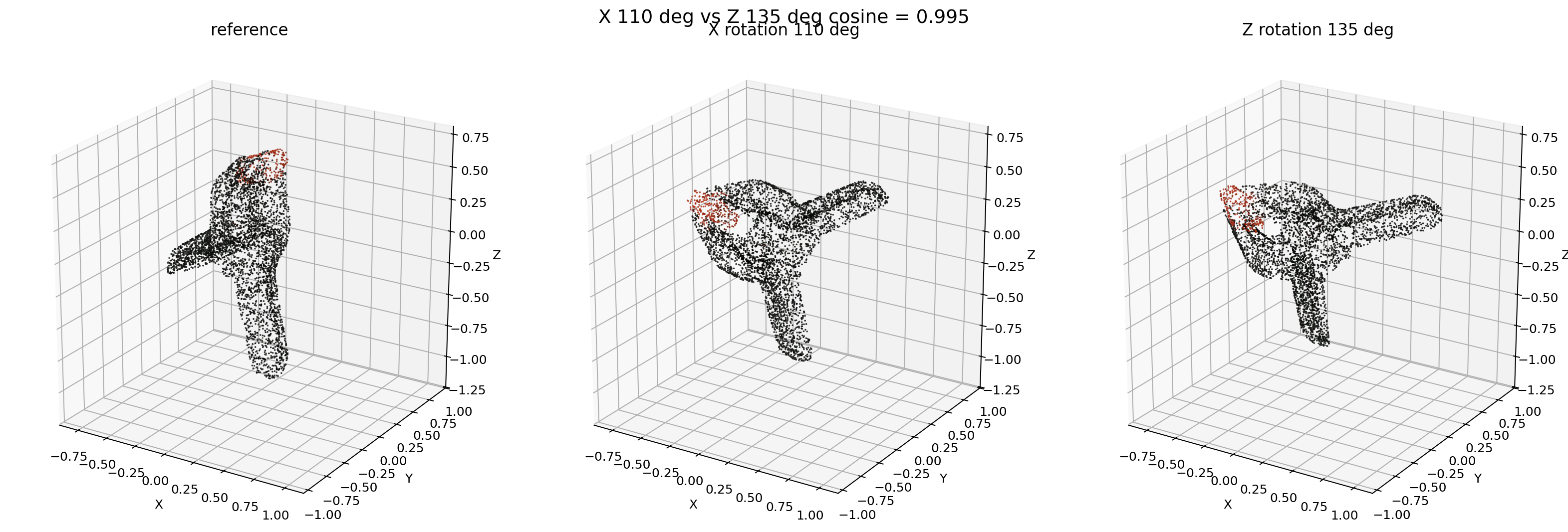}
    \hfill
    \includegraphics[width=0.28\linewidth]{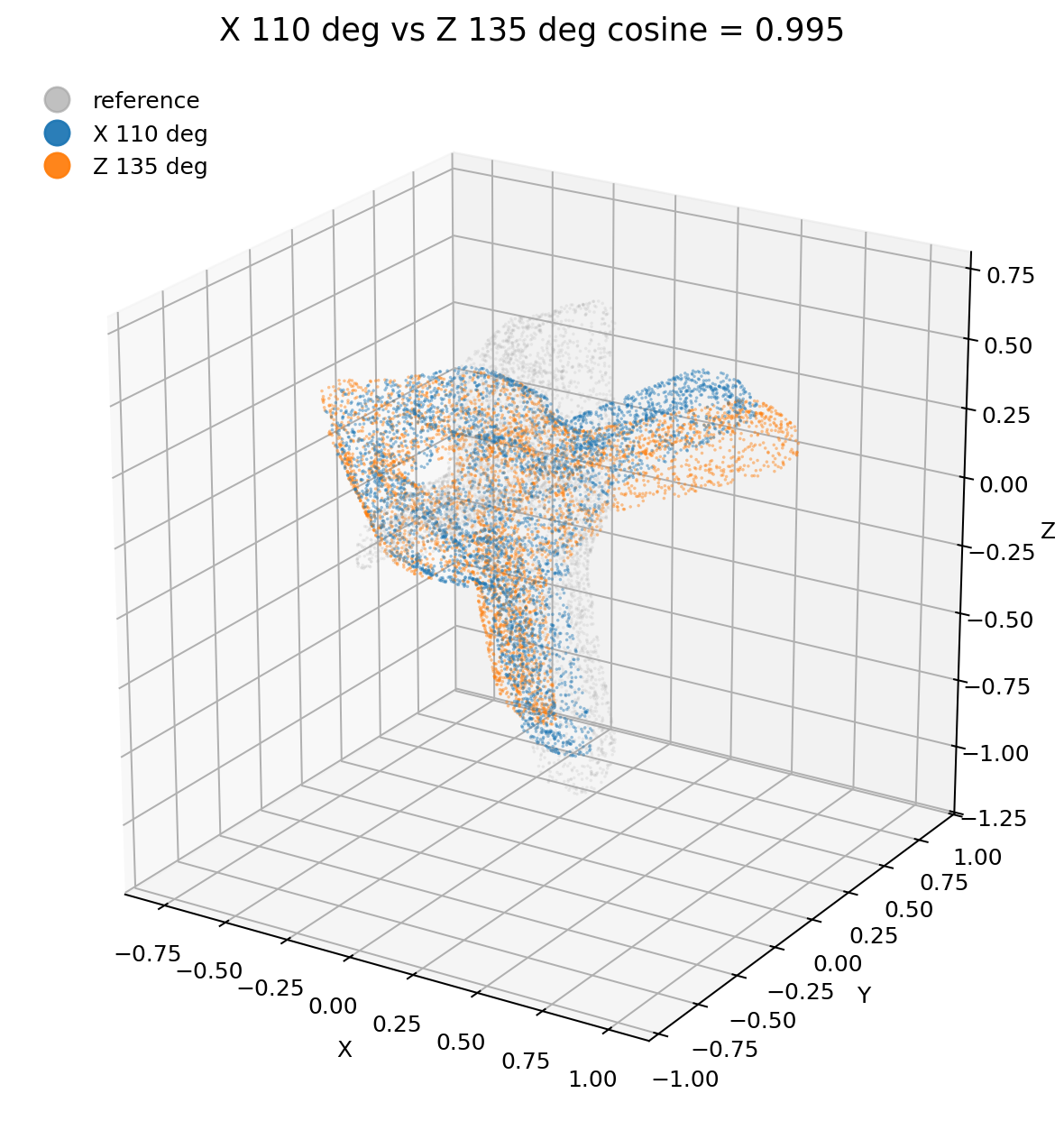}
  }\\
  \subfloat[Scissors: Z165 vs. X160\label{fig:appendix_symmetry_pointcloud_scissors}]{
    \includegraphics[width=0.68\linewidth]{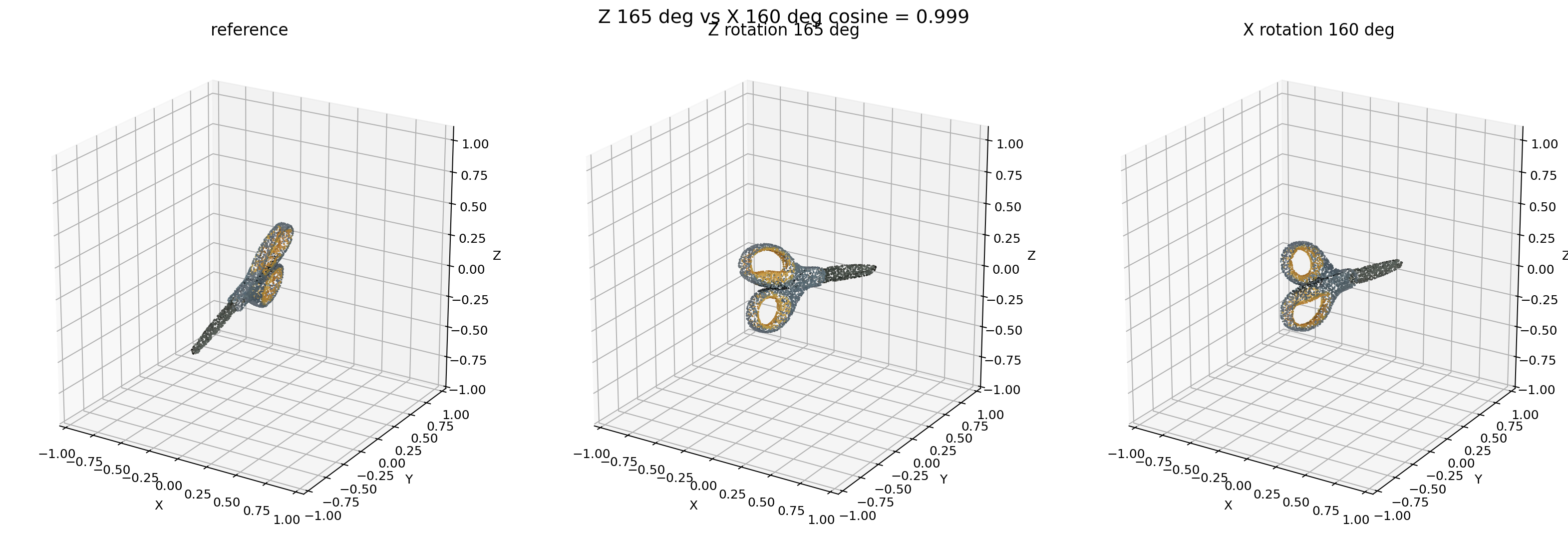}
    \hfill
    \includegraphics[width=0.28\linewidth]{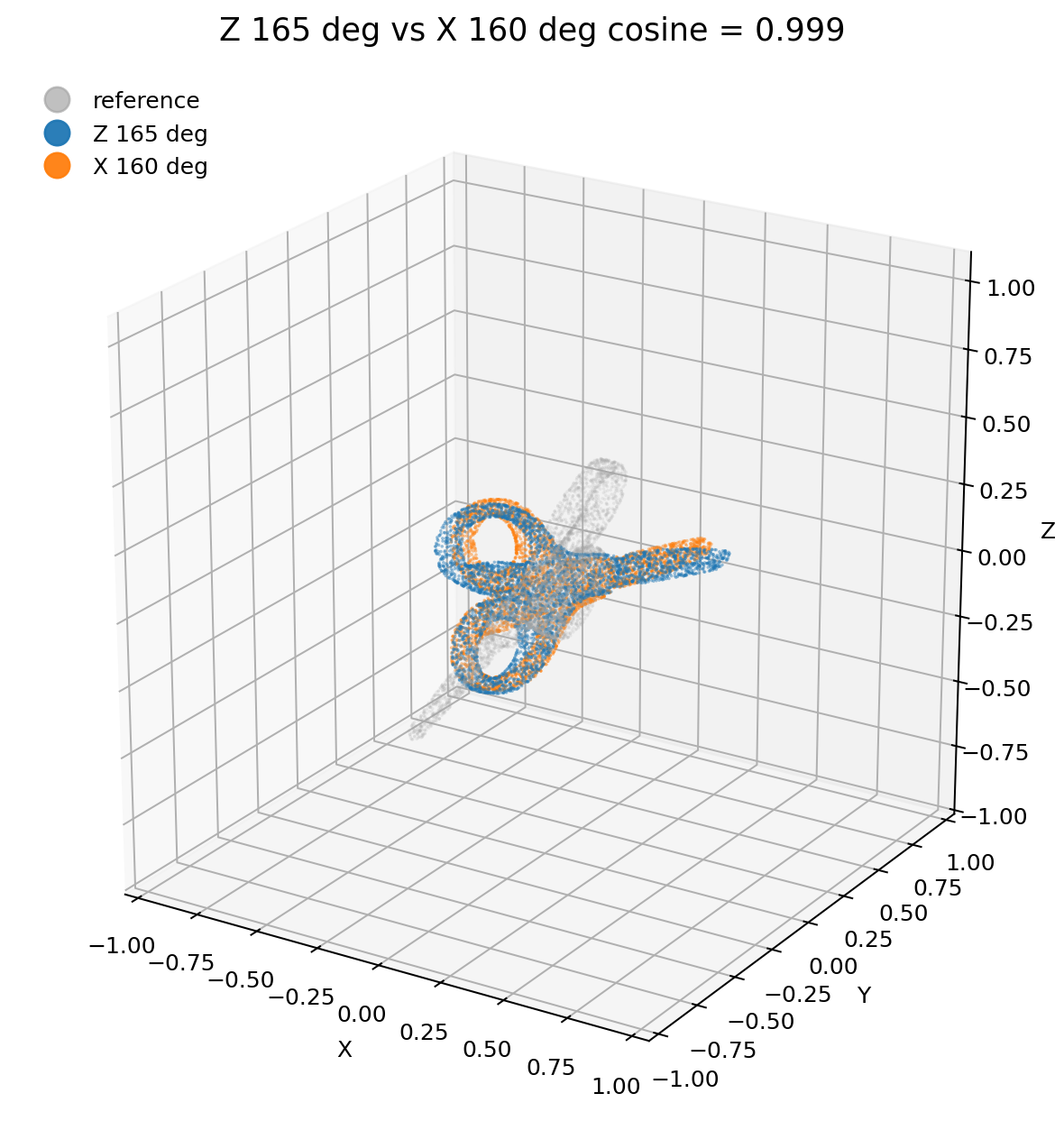}
  }\\
  \subfloat[Large marker: Y150 vs. Z180\label{fig:appendix_symmetry_pointcloud_large_marker}]{
    \includegraphics[width=0.68\linewidth]{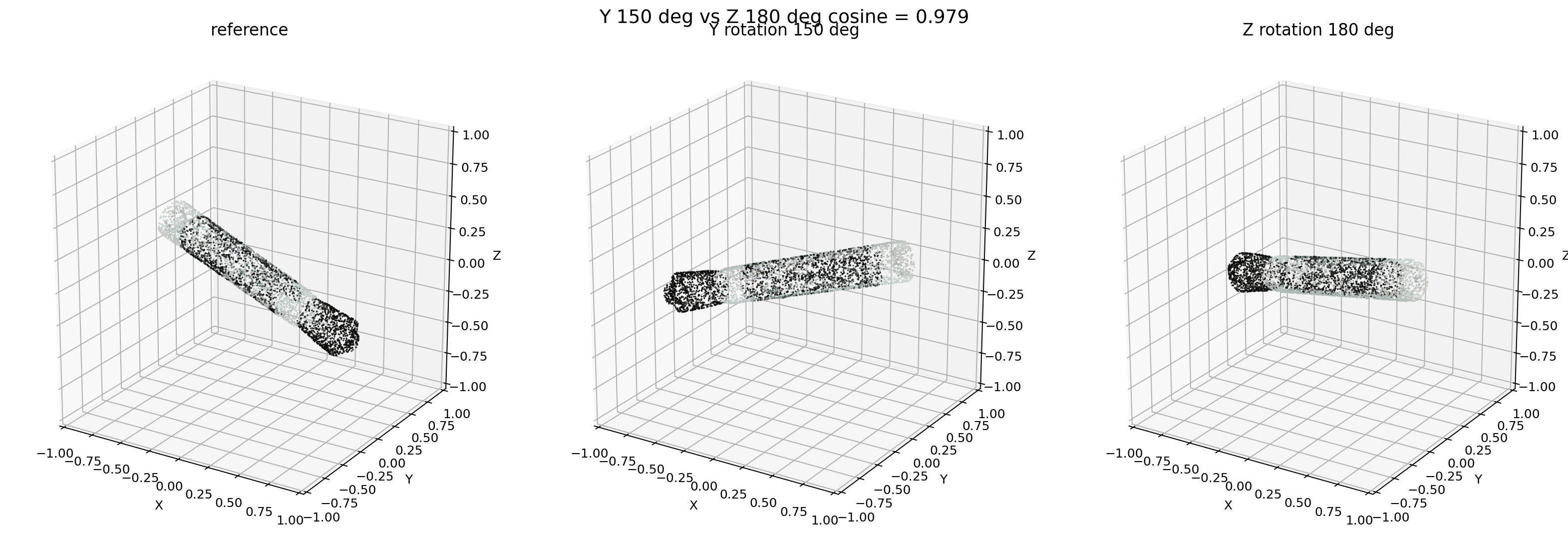}
    \hfill
    \includegraphics[width=0.28\linewidth]{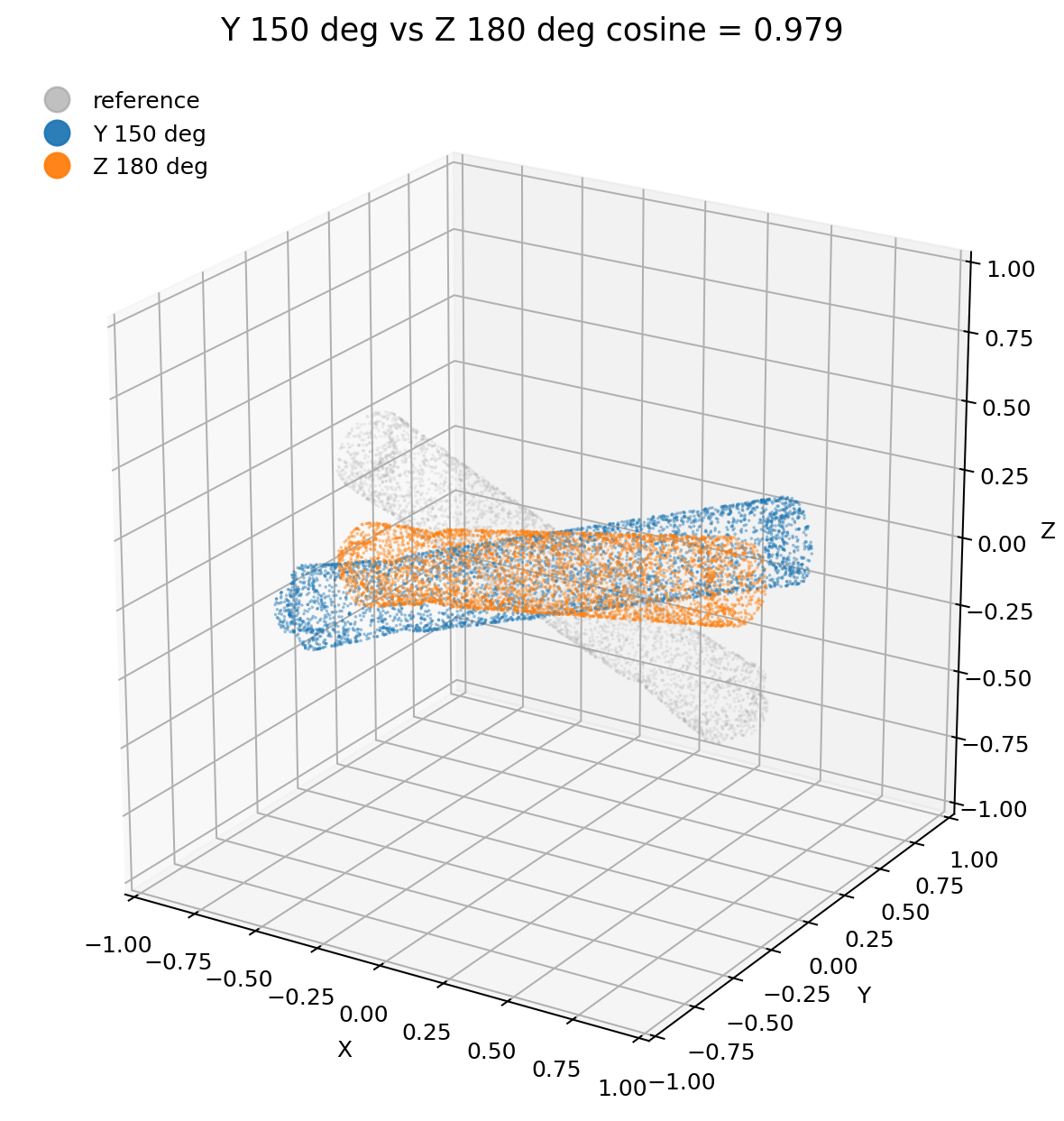}
  }
  \caption{Raw colored point-cloud renders and overlaid comparison clouds for selected high-cosine cross-axis rotations. Each row shows the reference and two RGB-colored rotated clouds on the left, followed by a single overlay with the compared rotations and reference on the right.}
  \label{fig:appendix_symmetry_raw_pointclouds}
\end{figure*}

The diagnostics above assume that rotation changes are visually identifiable from
object geometry. We include three stress cases where object symmetry makes
different axis-angle perturbations look similar from point-cloud geometry alone.
The angle t-SNE plots in Figure~\ref{fig:appendix_symmetry_tsne} identify these
neighborhoods using rotation-angle color and rotation-axis markers. The
axis-delta cosine plots in Figure~\ref{fig:appendix_symmetry_axis_delta} quantify the
same ambiguity, while Figure~\ref{fig:appendix_symmetry_raw_pointclouds} places the
raw colored point-cloud renders next to the overlaid comparison cloud that makes
the similarity visible.

\paragraph{Medium clamp.}
The medium clamp has mirrored jaws connected by a thicker central body. When a
rotation preserves the apparent C-shaped layout, distinct axes can produce very
similar colored point clouds. The selected $X=110^\circ$ and $Z=135^\circ$
rotations have cosine similarity $0.995$; the corresponding angle t-SNE,
axis-delta, and point-cloud panels are shown in
Figure~\ref{fig:appendix_symmetry_tsne_medium_clamp}, Figure~\ref{fig:appendix_symmetry_axis_delta_medium_clamp}, and Figure~\ref{fig:appendix_symmetry_pointcloud_medium_clamp}, respectively.

\paragraph{Scissors.}
The scissors are not globally rotationally symmetric, but the two handles and
long blades form a strong bilateral shape. Near half-turn rotations can align
the handle-and-blade outline, so the selected $Z=165^\circ$ and $X=160^\circ$
rotations nearly collapse in embedding direction with cosine similarity $0.999$;
the matching diagnostic panels are
Figure~\ref{fig:appendix_symmetry_tsne_scissors}, Figure~\ref{fig:appendix_symmetry_axis_delta_scissors}, and Figure~\ref{fig:appendix_symmetry_pointcloud_scissors}.

\paragraph{Large marker.}
The marker is dominated by a long cylindrical body with a near-circular cross
section.The selected $Y=150^\circ$ and $Z=180^\circ$ rotations have
cosine similarity $0.979$, with the angle t-SNE, axis-delta, and point-cloud
evidence shown in
Figure~\ref{fig:appendix_symmetry_tsne_large_marker}, Figure~\ref{fig:appendix_symmetry_axis_delta_large_marker}, and Figure~\ref{fig:appendix_symmetry_pointcloud_large_marker}, respectively.

\clearpage

\subsection{Rotation Surface Plots}
\label{app:rotation_surface_plots}

\begin{figure*}[!htbp]
  \centering
  \subfloat[Ours\label{fig:appendix_rotation_sweep_ours_peach}]{
    \includegraphics[width=0.9\linewidth]{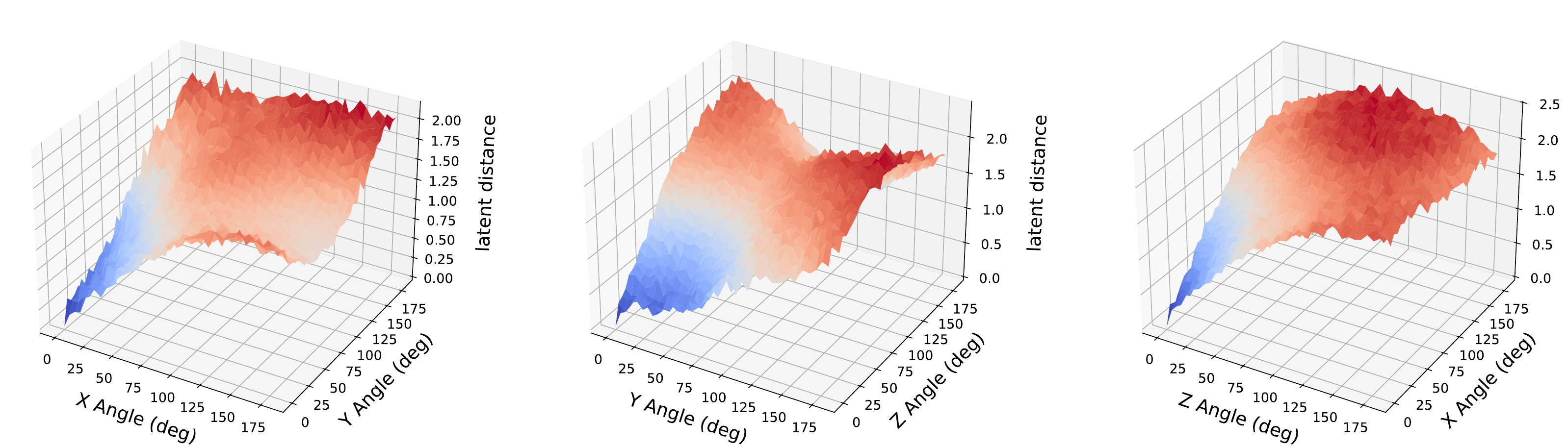}
  } \\
  \subfloat[PointMAE\label{fig:appendix_rotation_sweep_pointmae_peach}]{
    \includegraphics[width=0.9\linewidth]{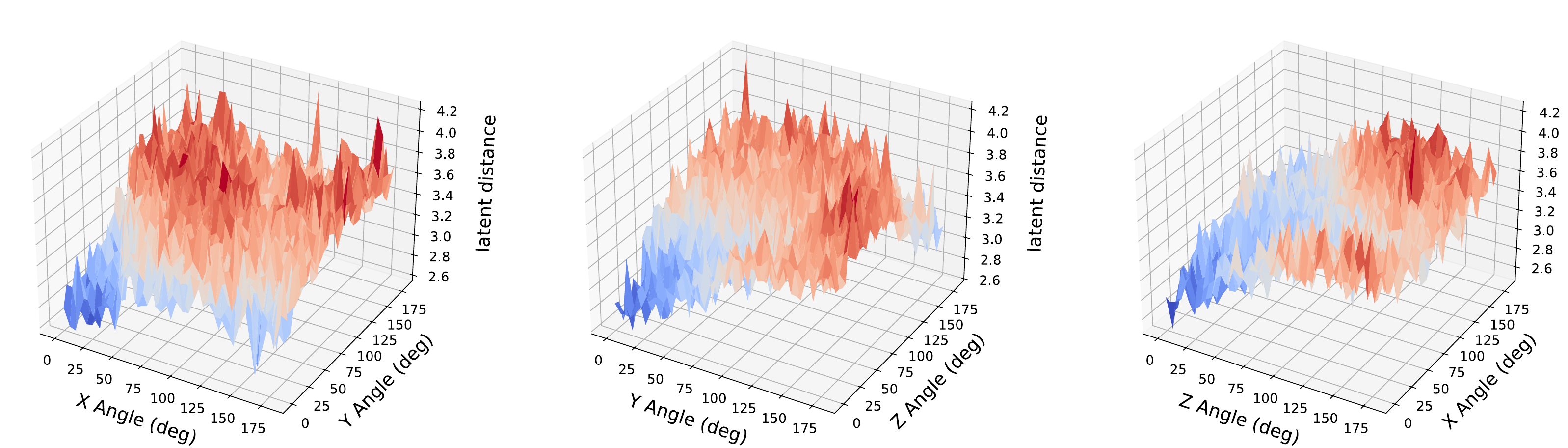}
  } \\
  \subfloat[Distilled Student\label{fig:appendix_rotation_sweep_student_peach}]{
    \includegraphics[width=0.9\linewidth]{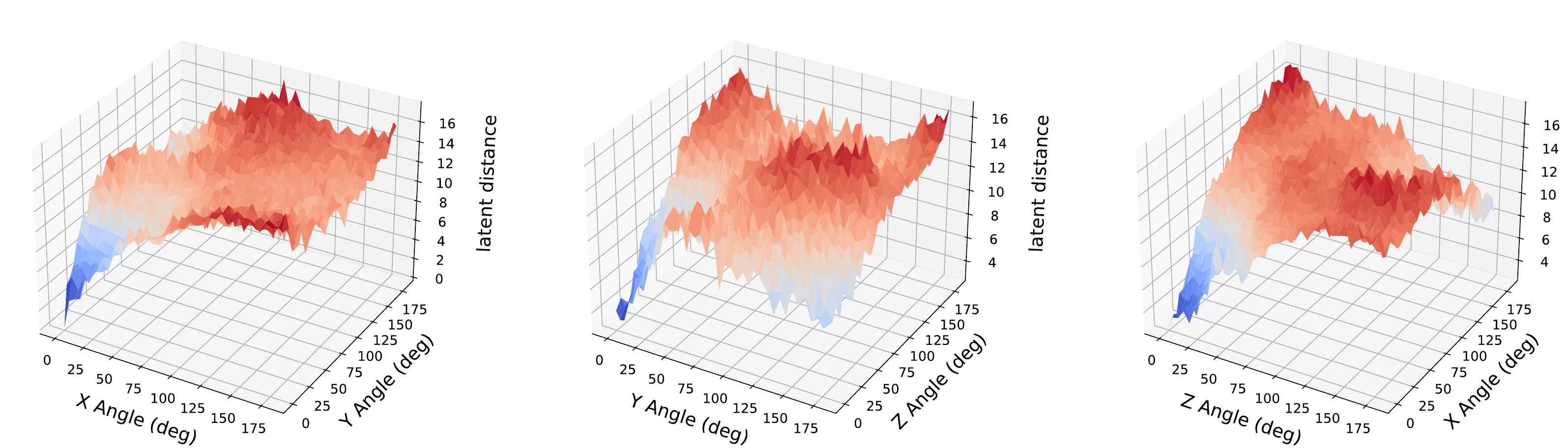}
  }
  \caption{Latent embedding distance as a function of simultaneous rotations around pairs of axes (X-Y, Y-Z, Z-X) for peach point clouds. Color scale: blue (low distance) to red (high distance).}
  \label{fig:appendix_rotation_sweep_peach}
\end{figure*}

\begin{figure*}[!htbp]
  \centering
  \subfloat[Ours\label{fig:appendix_rotation_sweep_ours_rubiks_cube}]{
    \includegraphics[width=0.9\linewidth]{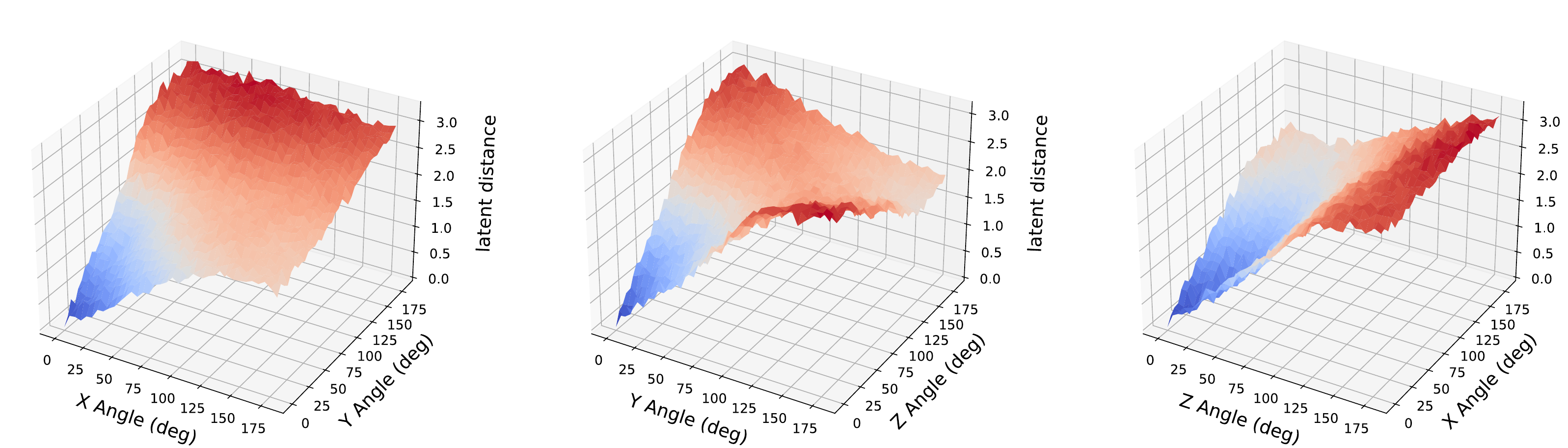}
  } \\
  \subfloat[PointMAE\label{fig:appendix_rotation_sweep_pointmae_rubiks_cube}]{
    \includegraphics[width=0.9\linewidth]{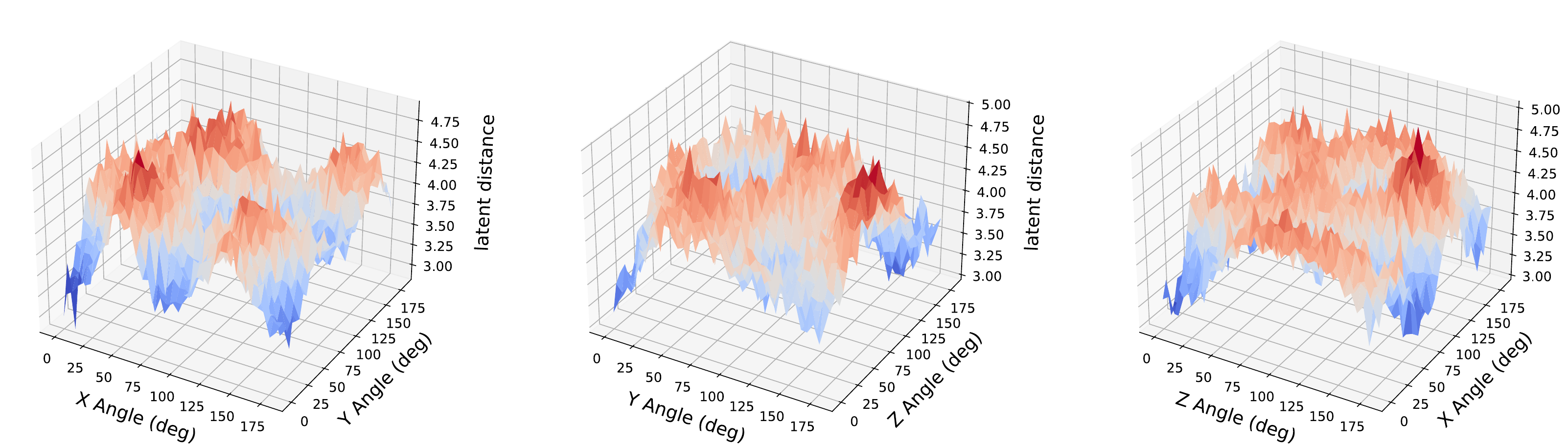}
  } \\
  \subfloat[Distilled Student\label{fig:appendix_rotation_sweep_student_rubiks_cube}]{
    \includegraphics[width=0.9\linewidth]{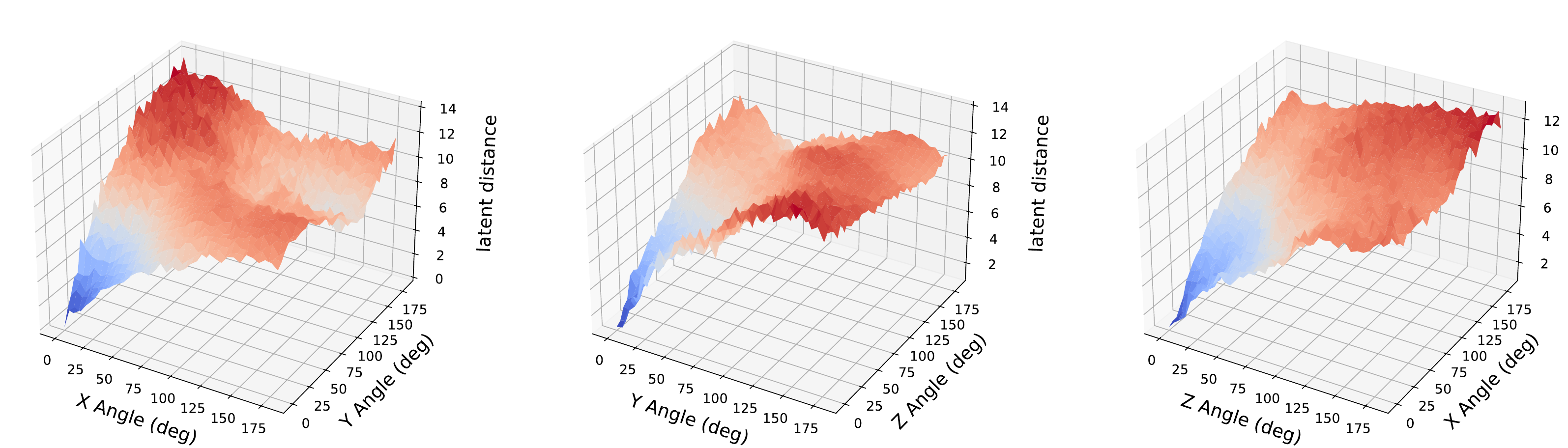}
  }
  \caption{Latent embedding distance as a function of simultaneous rotations around pairs of axes (X-Y, Y-Z, Z-X) for Rubik's cube point clouds. Color scale: blue (low distance) to red (high distance).}
  \label{fig:appendix_rotation_sweep_rubiks_cube}
\end{figure*}

\begin{figure*}[!htbp]
  \centering
  \subfloat[Ours\label{fig:appendix_rotation_sweep_ours_tuna_can}]{
    \includegraphics[width=0.9\linewidth]{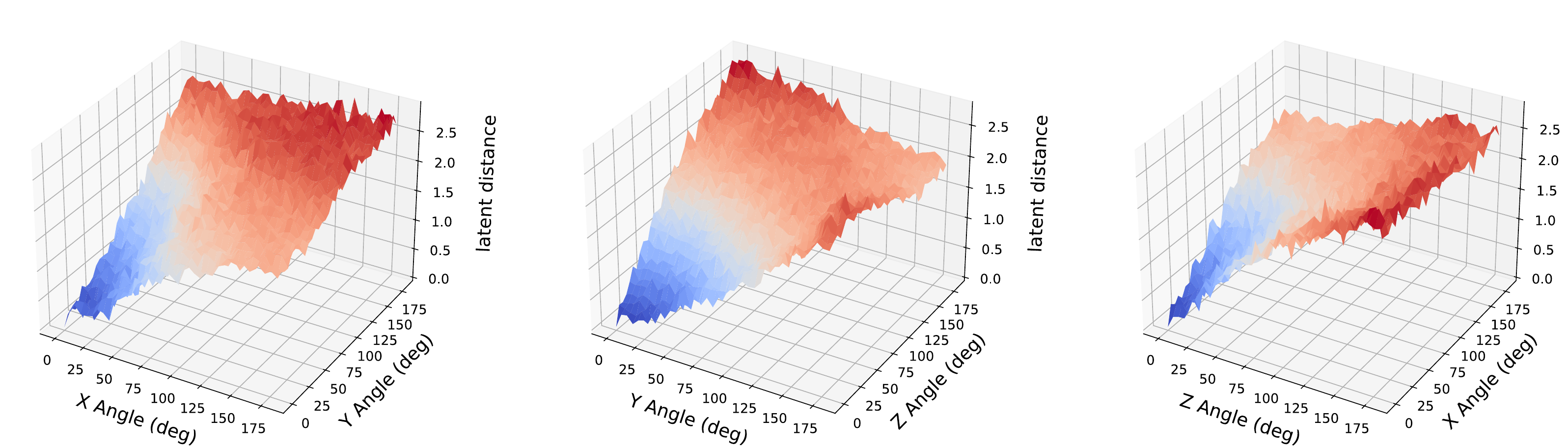}
  } \\
  \subfloat[PointMAE\label{fig:appendix_rotation_sweep_pointmae_tuna_can}]{
    \includegraphics[width=0.9\linewidth]{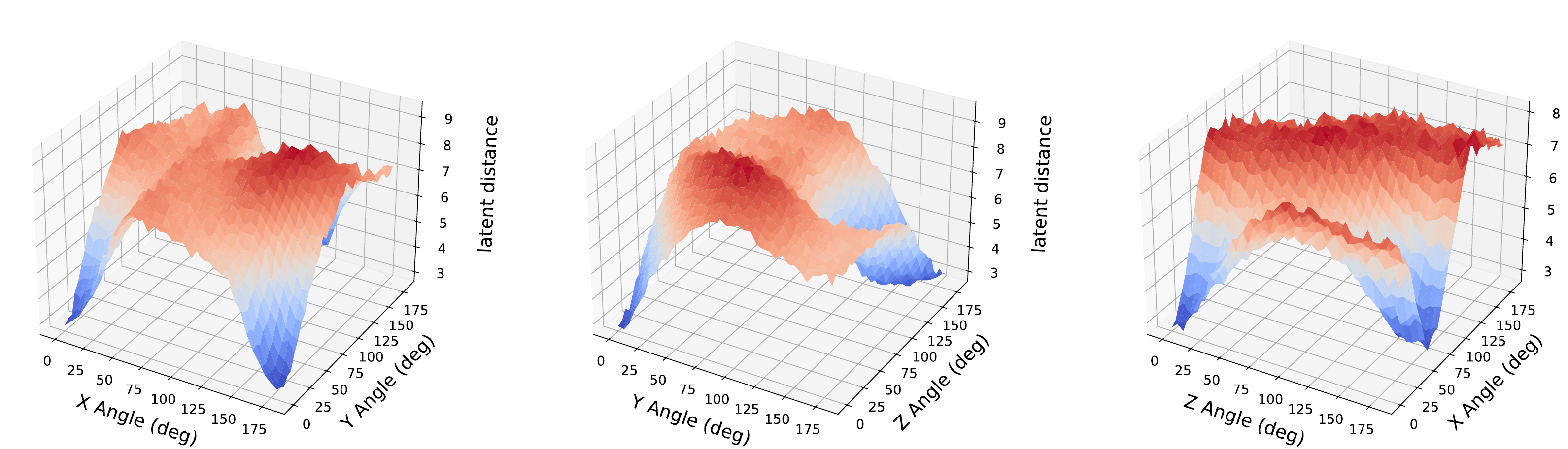}
  } \\
  \subfloat[Distilled Student\label{fig:appendix_rotation_sweep_student_tuna_can}]{
    \includegraphics[width=0.9\linewidth]{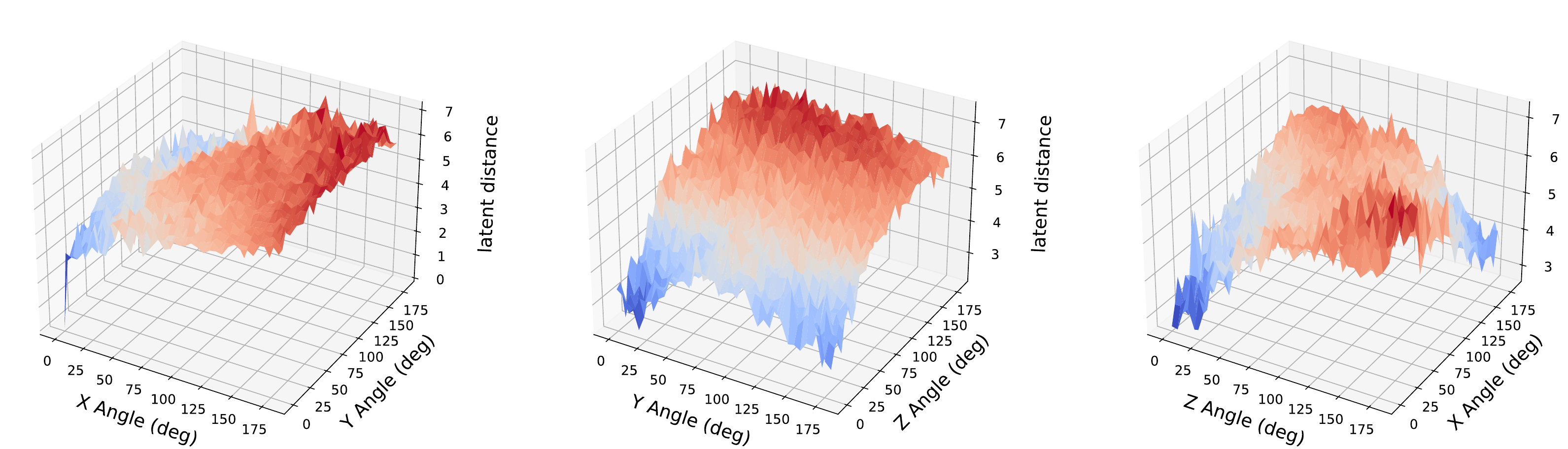}
  }
  \caption{Latent embedding distance as a function of simultaneous rotations around pairs of axes (X-Y, Y-Z, Z-X) for tuna can point clouds. Color scale: blue (low distance) to red (high distance).}
  \label{fig:appendix_rotation_sweep_tuna_can}
\end{figure*}

\end{document}